\documentclass[twocolumn]{article}
\usepackage{preprint}

\AtBeginDocument{%
  }

\usepackage{xcolor}
\usepackage{color}
\usepackage{wrapfig}
\newcommand{\revise}[1]{{\color{black} #1}}
\usepackage{geometry}
\usepackage{graphicx}
\usepackage{overpic}
\usepackage{siunitx}
\usepackage{pifont}
\usepackage{svg}
\usepackage{booktabs}
\usepackage{caption}
\usepackage{subcaption} 
\usepackage{amsmath}
\usepackage{amsthm}
\usepackage{amsfonts} 
\usepackage{makebox}
\usepackage{rotating}
\usepackage{makecell}
\usepackage{multirow}
\usepackage[table]{xcolor} 
\usepackage{array}    
\newcommand{\rednote}[1]{{\color{black} #1}}
\newcommand{\re}[1]{{\color{black} #1}}
\newtheorem{theorem}{Theorem}
\usepackage{hyperref}
\usepackage{natbib}
\usepackage{enumitem}
\theoremstyle{definition}
\newtheorem{definition}{Definition}
\usepackage{algorithm, algpseudocode}
\usepackage{newfloat}
\DeclareFloatingEnvironment[name={Supplementary Figure}]{suppfigure}
\usepackage{sidecap}
\sidecaptionvpos{figure}{c}

\usepackage{float}  %
\newfloat{teaserfigure}{t}{lop} %

\date{\phantom{\today}}
\usepackage{titlesec}
\titlespacing\section{0pt}{12pt plus 3pt minus 3pt}{1pt plus 1pt minus 1pt}
\titlespacing\subsection{0pt}{10pt plus 3pt minus 3pt}{1pt plus 1pt minus 1pt}
\titlespacing\subsubsection{0pt}{8pt plus 3pt minus 3pt}{1pt plus 1pt minus 1pt}

\title{Voronoi-Assisted Optimization for Diffusing Unsigned Distance Fields from Unoriented Points}

\author{Jiayi Kong$^*$\\S-Lab\\Nanyang Technological University\\Singapore\And
Chen Zong\thanks{J. Kong and C. Zong contribute equally to the project. }\\College of Mathematics\\Nanjing University of Aeronautics and Astronautics\\China
\And Junkai Deng\\College of Computing and Data Science\\Nanyang Technological University\\Singapore
\And Xuhui Chen\\Institute of Software\\Chinese Academy of Sciences\\China
\And Fei Hou\\Institute of Software\\Chinese Academy of Sciences\\China
\And Shiqing Xin\\School of Computer Science\\Shandong University\\China
\And Junhui Hou\\Department of Computer Science\\City University of Hong Kong\\China
\And Chen Qian\\SenseTime Research\\China
\And Ying He\thanks{Corresponding author: Y. He (yhe@ntu.edu.sg) }\\S-Lab\\Nanyang Technological University\\Singapore
}
\renewcommand{\shorttitle}{\textit{arXiv} Template}
\begin{document}
\maketitle

\begin{abstract}
Unsigned Distance Fields (UDFs) provide a flexible representation for 3D shapes with arbitrary topology, including open and closed surfaces, orientable and non-orientable geometries, and non-manifold structures. While recent neural approaches have shown promise in learning UDFs, they often suffer from numerical instability, high computational cost, and limited controllability. \rednote{We present a lightweight, network-free method, \emph{Voronoi-Assisted Optimization for Diffusing} (VAD), to compute UDFs directly from unoriented point clouds}. Our approach begins by assigning bi-directional normals to input points, guided by two Voronoi-based geometric criteria encoded in an energy function for optimal alignment. The aligned normals are then diffused to form an approximate UDF gradient field, which is subsequently integrated to recover the final UDF. Experiments demonstrate that VAD robustly handles watertight and open surfaces, as well as complex non-manifold and non-orientable geometries, while remaining computationally efficient and stable. 
\end{abstract}
\section{Introduction}
\label{sec:introduction}

Implicit functions are widely used in 3D reconstruction from unorganized point clouds. Several types have been explored, including distance fields~\cite{park2019deepsdf,chibane2020neural}, occupancy fields~\cite{Occupancy_Networks}, generalized winding numbers (GWN)~\cite{jacobson2013winding}, and Poisson indicator functions~\cite{Kazhdan2006PoissonSR}. Among these, distance fields are especially popular due to their smoothness, ease of learning, natural support for offset surfaces, and ability to encode rich geometric information.

Distance fields can be categorized into two types. Signed distance fields (SDFs) are well-suited for watertight surfaces, where interior and exterior regions are clearly distinguishable via signs. Unsigned distance fields (UDFs), in contrast, provide greater flexibility for modeling arbitrary topologies, including open and closed surfaces, manifold and non-manifold structures, and orientable or non-orientable geometries. \revise{Despite this flexibility, generating high-quality UDFs remains difficult due to the lack of oriented normals, the challenge of enforcing global consistency, and the need to robustly handle non-manifold configurations that frequently arise in real-world point clouds. }

Most existing methods compute UDFs from raw point clouds by training neural networks to fit the zero level set~\cite{Zhou2022CAP-UDF,DBLP:conf/iccv/RenHCHW23,Xu2024DEUDF, Zhou2023LearningAM, Fainstein2024DUDF,losf-udf-2024}. However, these approaches often suffer from numerical instability, leading to fragmented surfaces, and regions far from the surface may fail to exhibit the properties of a well-defined UDF. Moreover, the inherent characteristics of neural networks can introduce unpredictable variations in the results, making them difficult to precisely control. Additionally, the computational cost of these methods is non-negligible. 

\begin{figure}[tb]
    \centering
    \includegraphics[width=\linewidth]{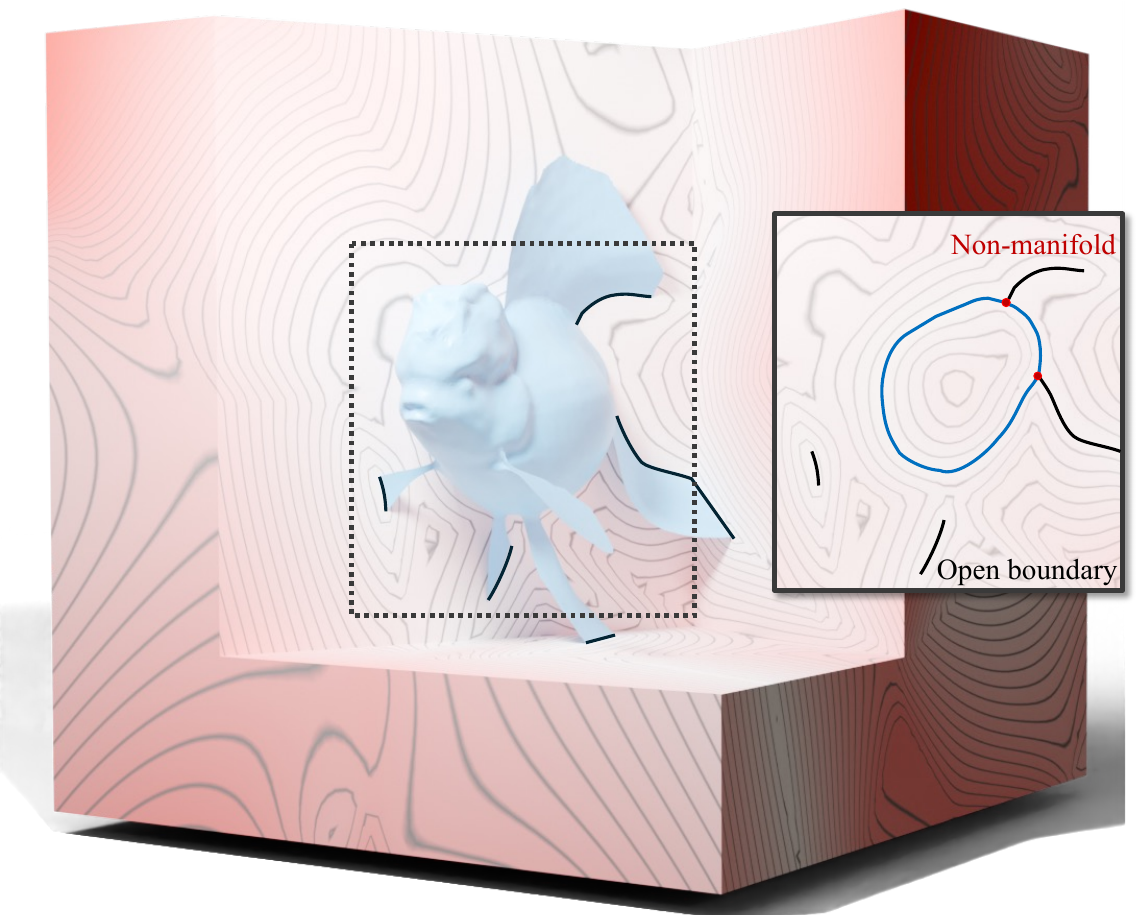}
    \caption{VAD is particularly effective in handling open surfaces, non-manifold structures, and non-orientable geometries—cases that are challenging for conventional SDF- or occupancy field-based methods. Shown here is the reconstructed golden fish model, which contains both open boundaries and non-manifold structures. We visualize the computed UDFs using cut views: black curves indicate open boundaries, red dots represent non-manifold features, and the zero level set is shown in blue.} 
    \label{fig:teaser}
\end{figure}

In this paper, we propose a lightweight, network-free approach for approximating unsigned distance fields from unoriented point clouds.
Inspired by the heat method~\cite{feng2024heat} for computing SDFs on oriented data, we develop a Voronoi-assisted framework that first aligns point normals and then reconstructs the UDF over the domain of interest. The key idea is that \rednote{the projection field---which measures the distance from a spatial query to its nearest point along the associated normal direction---}provides two geometric criteria for optimizing the bi-directional alignment of normals. Once aligned, 
the normals are diffused using a heat-based formulation
to approximate the UDF gradient field, 
and the final UDF is recovered by solving the Poisson equation.
Furthermore, we extend this method to handle noisy inputs. 

We demonstrate the effectiveness of our UDF approximation method on diverse reconstruction tasks, \rednote{particularly on open surfaces and geometries with ambiguous or inconsistent normal orientations, including non-manifold and non-orientable cases. In such scenarios, SDF- and GWN-based approaches are inherently limited due to their reliance on globally consistent orientation assumptions, while prior UDF methods often struggle to produce stable reconstructions. To further evaluate robustness, we also test on sparsely and non-uniformly sampled point clouds.}

\section{Related Work}
\label{sec:related}
Reconstructing 3D surfaces from point clouds has long been a central topic in computer graphics and geometry processing. Early computational geometry approaches, such as $\alpha$-shapes~\cite{edelsbrunner1994three, bernardini1997sampling}, Ball Pivoting~\cite{bernardini2002ball}, Power Crust~\cite{amenta2001power} and Tight CoCone~\cite{Dey2003}, reconstructed surfaces by exploiting local geometric and topological cues. Although these methods are efficient and come with theoretical guarantees, they typically require dense and uniform point samples, and their performance degrades significantly on noisy, sparse, or irregular inputs. 

To overcome these limitations, a widely adopted strategy is to define a non-degenerate scalar field whose level set represents the target surface~\cite{hoppe1992surface,kolluri2008imls, oztireli2009RIMLS, shen2004interpolating, schroers2014HessianIMLS, ohtake2003multi, SSD}. Implicit-function methods offer several advantages: they naturally produce smooth surfaces, fit well into optimization frameworks, and exhibit robustness against noise.

Implicit functions appear in several  forms, including signed distance fields~\cite{hoppe1992surface, carr2001reconstruction, ohtake2003multi, feng2024heat, kobbelt}, unsigned distance fields~\cite{chibane2020neural,Zhou2022CAP-UDF,mullen2010signing}, Poisson indicator functions~\cite{Kazhdan2006PoissonSR, kazhdan2013screened,Kazhdan2020PoissonSR, SPSR2022}, occupancy fields~\cite{Occupancy_Networks}, and generalized winding number fields~\cite{Barill2018FastWN,Lin2024fast,liu2025dwg,Huang2024Stochastic,lin2022surface,liu2024consistent,xu2023globally}. All of these implicit functions, except UDFs, are primarily intended for watertight objects, where interior and exterior regions are clearly distinguishable. In contrast, unsigned distance fields discard the sign in SDFs and can represent arbitrary topologies, including open surfaces and non-manifold structures.

Despite their flexibility, UDFs are difficult to learn with neural networks. Their inherent non-differentiability at the zero level set prevents straightforward integration into neural architectures, which are better suited for smooth functions. DEUDF~\cite{Xu2024DEUDF} addresses this challenge by relaxing the non-negativity constraint on network outputs and introducing an adaptive Eikonal regularizer that avoids penalizing points near the zero level set during training. However, network-based approaches still suffer from high computational costs, making them less practical for large-scale or time-sensitive applications. 

\begin{figure}
    \centering    \includegraphics[width=1.0\linewidth]{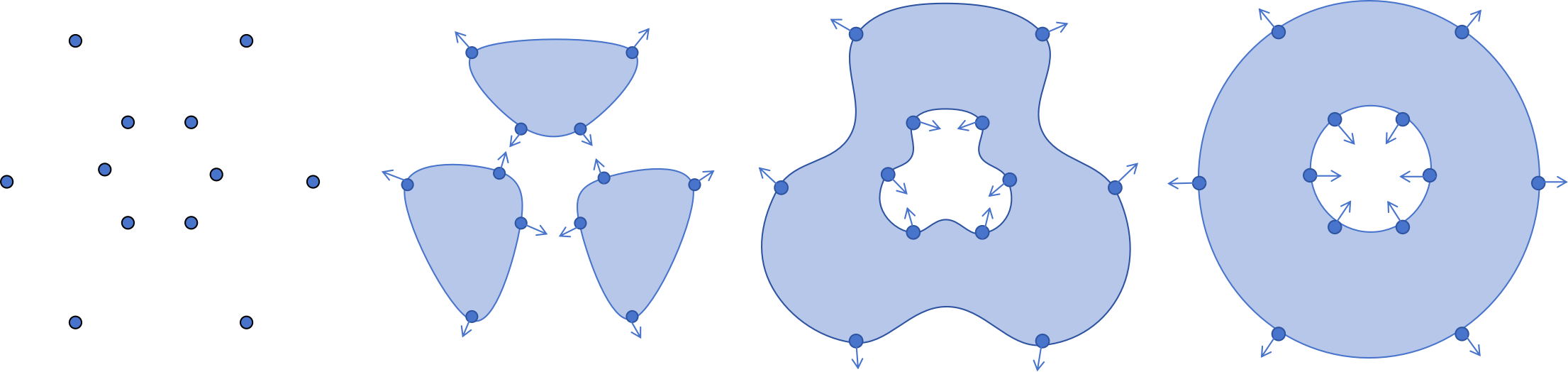}\\    \makebox[0.245\linewidth]{(a)}    \makebox[0.245\linewidth]{(b)}\makebox[0.245\linewidth]{(c)}\makebox[0.245\linewidth]{(d)}\\
    \caption{\revise{With identical input points at relatively low density (a), different normal orientations produce markedly different reconstructed surface topologies and geometries (b-d), illustrating that without reliable orientation the reconstruction problem is inherently ill-posed at low sampling rates. }}
    \label{fig:ambiguity-inset}
\end{figure}

\revise{Beyond network-related challenges, UDF computation is closely tied to the problem of point orientation. As illustrated in Figure~\ref{fig:ambiguity-inset}, without consistent normal directions, implicit field computation becomes ambiguous and unstable, particularly for non-watertight surfaces. To address this, recent work has focused on jointly estimating normals and constructing implicit fields.}

iPSR~\cite{hou2022iterative} employs an iterative scheme that refines normals within a Poisson surface reconstruction framework~\cite{kazhdan2005reconstruction,kazhdan2013screened}. GCNO~\cite{xu2023globally} formulates orientation as an optimization problem using generalized winding numbers. WNNC~\cite{Lin2024fast} and DWG~\cite{liu2025dwg} accelerate winding number computation with GPU implementations, enabling the processing of large-scale point clouds. \rednote{Dipole Propagation~\cite{dipolepropogation} utilizes a global electric dipole field for orientation propagation.} However, all of these approaches assume watertight models and cannot handle open surfaces or non-manifold structures, scenarios that our method is specifically designed to address.

\rednote{Our work tackles the challenges of UDF computation by leveraging the \emph{projection distance field}, a concept closely related to Voronoi diagrams.} Voronoi-based strategies have also been explored in prior work on surface reconstruction. For example, Alliez et al.~\citep{Alliez_variational} proposed a variational method that reconstructs surfaces from unoriented point clouds while allowing controllable smoothness. 
While this approach avoids the need for pre-oriented normals by exploiting Voronoi partitioning, it is restricted to manifold surfaces and cannot handle non-manifold configurations. 
Similarly, GCNO~\cite{xu2023globally} and BIM~\cite{liu2024consistent} employ Voronoi diagrams for spatial partitioning by associating each point with two auxiliary samples, one placed inside and the other outside the inferred surface, to guide the optimization of GWN fields. 

\re{In contrast, our method leverages Voronoi diagrams in a fundamentally different manner. Rather than using Voronoi cells solely for spatial partitioning or auxiliary sample generation, we treat point orientations as optimization variables and iteratively refine them through local projection-distance energy minimization, yielding a consistent unsigned distance representation. Consequently, our approach naturally accommodates open surfaces, non-orientable geometries, and non-manifold structures, which remain challenging for previous Voronoi-based methods~\cite{Alliez_variational,xu2023globally,liu2024consistent}.


While constructing a reliable UDF from raw point clouds is a fundamental prerequisite, practical reconstruction pipelines ultimately require explicit surface extraction. Unlike SDFs, UDFs do not encode inside--outside information, making classical Marching Cubes inapplicable. To address this challenge, a variety of UDF meshing methods have been proposed, including gradient-tracking approaches~\cite{guillard2022udf,stella2024nsdudf,Zhou2022CAP-UDF,DBLP:conf/iccv/RenHCHW23}, dual-contouring-based methods~\cite{zhang2023dualmeshudf}, Voronoi-based extraction schemes~\cite{DBLP:journals/corr/abs-2602-02907}, and topology-aware reconstruction techniques~\cite{hou2023dcudf,chen2025dcudf2}. These methods differ substantially in their underlying principles and exhibit different trade-offs. Gradient-tracking methods rely heavily on smooth and consistent gradients, whereas dual-contouring-based methods require accurate local distance estimation around the surface. Topology-aware methods explicitly address the ambiguity of unsigned fields and are generally more capable of handling non-manifold geometries.

Our work focuses on the upstream problem of UDF construction rather than mesh extraction. By producing accurate and smooth UDFs directly from unoriented point clouds, our method can be integrated with existing UDF meshing algorithms and provides a reliable foundation for high-quality surface reconstruction.

}

\section{Projection Distance Fields}
\label{sec:projection-field}
\subsection{Definitions}
\label{subsec:definition}

Let $\mathcal{S}$ be a curved surface embedded in $\mathbb{R}^3$. Its unsigned distance field, 
denoted by $u$, is a function 
$u:\mathbb{R}^3 \to \mathbb{R}_{\geq 0}$ defined as
\[
u(\mathbf{x}) = \inf_{\mathbf{y}\in\mathcal{S}} \|\mathbf{x}-\mathbf{y}\|,
\]
that is, the shortest distance from a query point $\mathbf{x}\in\mathbb{R}^3$ to the surface $\mathcal{S}$. The gradient at $\mathbf{x}$ is denoted by $\nabla u(\mathbf{x})$. 

In practice, however, the input is not a continuous surface but a discrete point cloud sampled from $\mathcal{S}$. In this setting, the computation of $u$ becomes ambiguous and non-trivial. \rednote{To address this, we adopt the concept of \emph{projection distance fields}, first introduced by~\citep{hoppe1992surface}, which approximate UDF while accounting for orientation uncertainty. }

For a 3D vector $\mathbf{v}$, let $\widetilde{\mathbf{v}}$ denote its \textbf{bi-directional} version, meaning $\mathbf{v}$ and $-\mathbf{v}$ are treated as equivalent. Using this, we define the  projection distance as follows:
\begin{definition}
Given a surface sample $\mathbf{p}\in\mathcal{S}$ associated with a bi-directional vector $\widetilde{\mathbf{v}}$, the \textbf{projection distance} $d_{\mathbf{x},\widetilde{\mathbf{v}}}$ of a query point $\mathbf{x} \in \mathbb{R}^3$ is 
\begin{equation}\label{eq:proj_dis}
    d_{\mathbf{p}, \widetilde{\mathbf{v}}}(\mathbf{x}) = \frac{|(\mathbf{x} - \mathbf{p}) \cdot \mathbf{v}|}{\|\mathbf{v}\|},
\end{equation} where $\mathbf{v}$ is any representative of $\widetilde{\mathbf{v}}$ with the same magnitude.
\end{definition}

Note that in this definition, $\widetilde{\mathbf{v}}$ is not necessarily a surface normal; it is simply a bi-directional vector associated with sample $\mathbf{p}$.
Now consider a point set $\mathcal{P}=\{\mathbf{p}_i\mid \mathbf{p}_i\in\mathcal{S},\ i=1,\ldots, n\}$ 
\begin{wrapfigure}{r}{0.12\textwidth}
   \centering 
   \vspace{-1.5em}
   \hspace{0.5em}
  \makebox[38pt][r]{\includegraphics[width=0.125\textwidth]{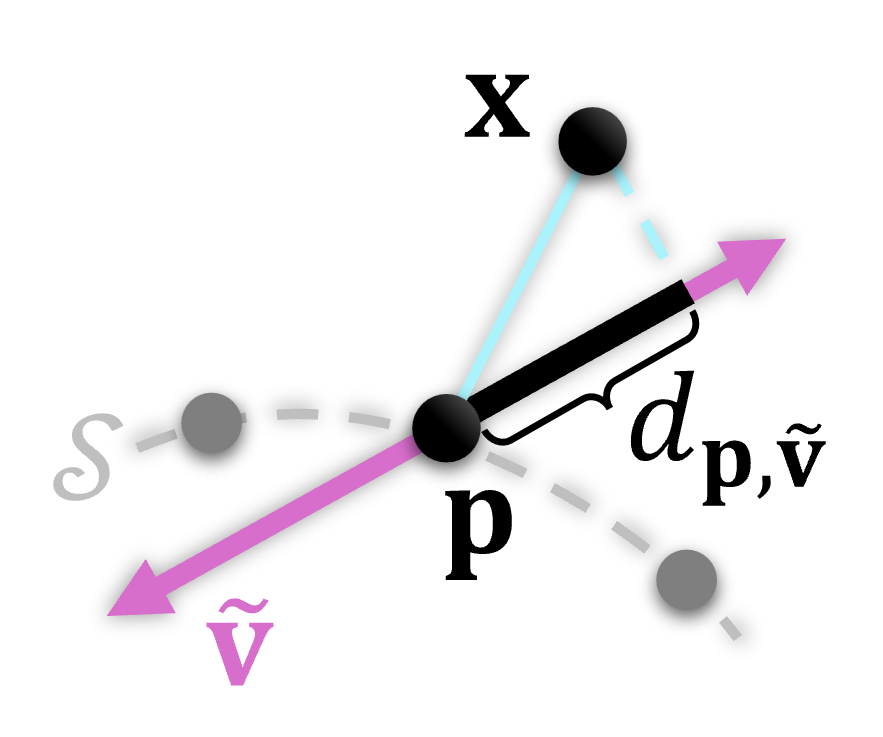}}
\end{wrapfigure}
sampled from $\mathcal{S}$, with each $\mathbf{p}_i$ associated with a bi-directional vector $\widetilde{\mathbf{v}}_i$. Let  $\mathcal{V}=\{\widetilde{\mathbf{v}}_i,\ i=1,\ldots, n\}$ be the set of such vectors. We define the projection distance field by projecting the standard unsigned distance onto the given direction $\mathbf{v}$:
\begin{definition}
Given surface samples $\mathcal{P}\in\mathbb{R}^3$ with associated bi-directional vector $\widetilde{\mathbf{v}}$, the \textbf{projection distance field} is the function $F:\mathbb{R}^3\rightarrow \mathbb{R}_{\geq0}$ defined as:
\[F_{\mathcal{P},\mathcal{V}}(\mathbf{x}) = d_{\mathbf{p}_j,\widetilde{\mathbf{v}}_j}(\mathbf{x}),
\]
where $\mathbf{p}_j$ is the closest point in $\mathcal{P}$ to $\mathbf{x}$.
\end{definition}

\subsection{Connections to Voronoi Tessellation}
\label{subsec:connectionstovoronoi}

The projection distance field is naturally linked to the Voronoi diagram of $\mathcal{P}$. Recall that the Voronoi cell  $\mathcal{C}_i$ of a point $\mathbf{p}_i$ is defined as
\[\mathcal{C}_i=\left\{\mathbf{x}\mid \|\mathbf{x}-\mathbf{p}_i\|^2\leq\|\mathbf{x}-\mathbf{p}_j\|^2,\ \forall j\neq i\right\}, 
\]
consisting of all points in $\mathbb{R}^3$ that are closest to $\mathbf{p}_i$. 

Within each Voronoi cell, the projection distance field is defined with respect to a single site $\mathbf{p}_i$ and vector $\widetilde{\mathbf{v}}_i$. Since the computation in Eq.~\eqref{eq:proj_dis} reduces to a linear function of $\mathbf{x}$ when $\mathbf{p}_i$ and $\widetilde{\mathbf{v}}_i$ are fixed, we obtain:
\begin{theorem}
The projection distance field $F_{\mathcal{P},\mathcal{V}}(\mathbf{x})$ is linear within each Voronoi cell $\mathcal{C}_i$.
\end{theorem}
This piecewise linearity implies that discontinuities, if any, must occur along Voronoi bisectors, where cells meet. Figure~\ref{fig:insight} illustrates typical discontinuities of $F_{\mathcal{P},\mathcal{V}}$ across bisectors under different orientation configurations.

\revise{\subsection{Discontinuity Analysis and Normal Alignment}

The \textbf{Voronoi bisector} $\mathcal{B}_{ij}$ between sites $\mathbf{p}_i$ and $\mathbf{p}_j$ is the locus of points equidistant to both sites. Discontinuities of $F_{\mathcal{P},\mathcal{V}}$ across Voronoi bisectors occur when the associated bi-directional vectors are not well aligned with the underlying surface normals, or when the surface has high curvature and the sampling density is insufficient. In our analysis, we assume that the surface has bounded principal curvature and that the point set $\mathcal{P}$ is sampled densely enough such that discontinuities arise mainly from misaligned bi-directional vectors.

Depending on the configuration, the projection distance field across $\mathcal{B}_{ij}$ may exhibit four characteristic cases:
\begin{enumerate}[leftmargin=1.8em]
    \item Both the values and gradients of $F_{\mathcal{P}, \mathcal{V}}$ match on the two sides of the bisector $\mathcal{B}_{ij}$ (see Fig.~\ref{fig:insight}~(a)).
    \item Only the gradients match, but the values are discontinuous (see  Fig.~\ref{fig:insight}~(b)).
    \item Only the values match, but the gradients are discontinuous (see  Fig.~\ref{fig:insight}~(c)).
    \item Neither the values nor the gradients match (see Fig.~\ref{fig:insight}~(d)).
\end{enumerate}

These cases motivate us to formulate orientation estimation as an optimization problem: the goal is to adjust bi-directional vectors to reduce discontinuities in both field values and gradients across bisectors. In the UDF setting, where inside–outside labels are undefined, bi-directional normals naturally replace signed normals. As illustrated in Figure~\ref{fig:insight}(a), when discontinuities vanish, the estimated bi-directional normals are nearly perpendicular to the underlying smooth surface.

\begin{figure}[htb]
    \centering
    \includegraphics[width=\linewidth]{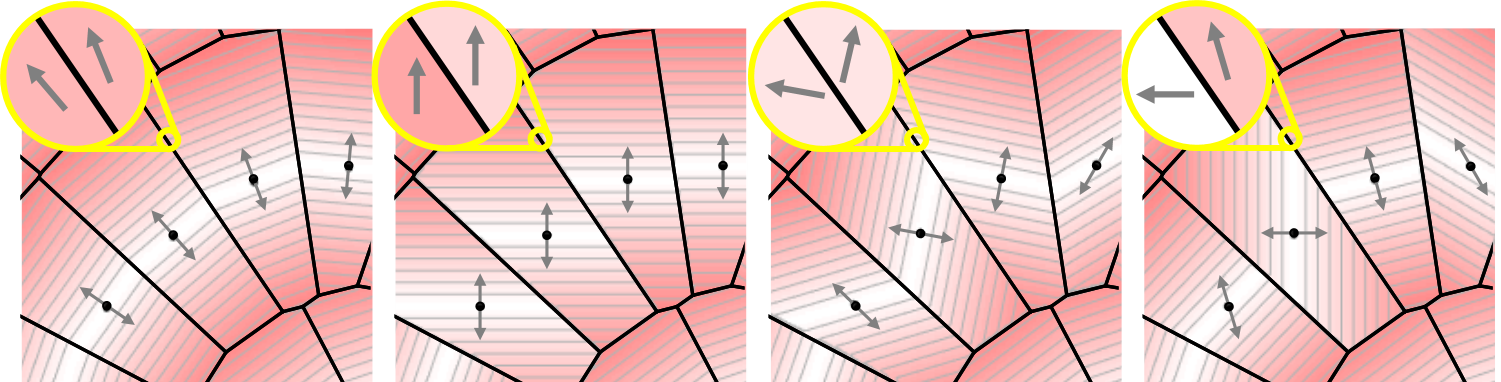}
    \makebox[0.12\textwidth]{(a)}%
    \makebox[0.12\textwidth]{(b)}%
    \makebox[0.12\textwidth]{(c)}%
    \makebox[0.12\textwidth]{(d)}
    \caption{2D illustrations of four typical behaviors of projection fields. All fields are computed from the same point set $\mathcal{P}$ but under different bi-directional vector configurations $\mathcal{V}$. Except in case (a), where the vectors align with the true surface normals, either the field value $F$ or its gradient $\nabla F$ exhibits discontinuities across bisectors.}

    \label{fig:insight}
\end{figure}

\begin{figure*}[htb]
    \centering
    \includegraphics[width=\linewidth]{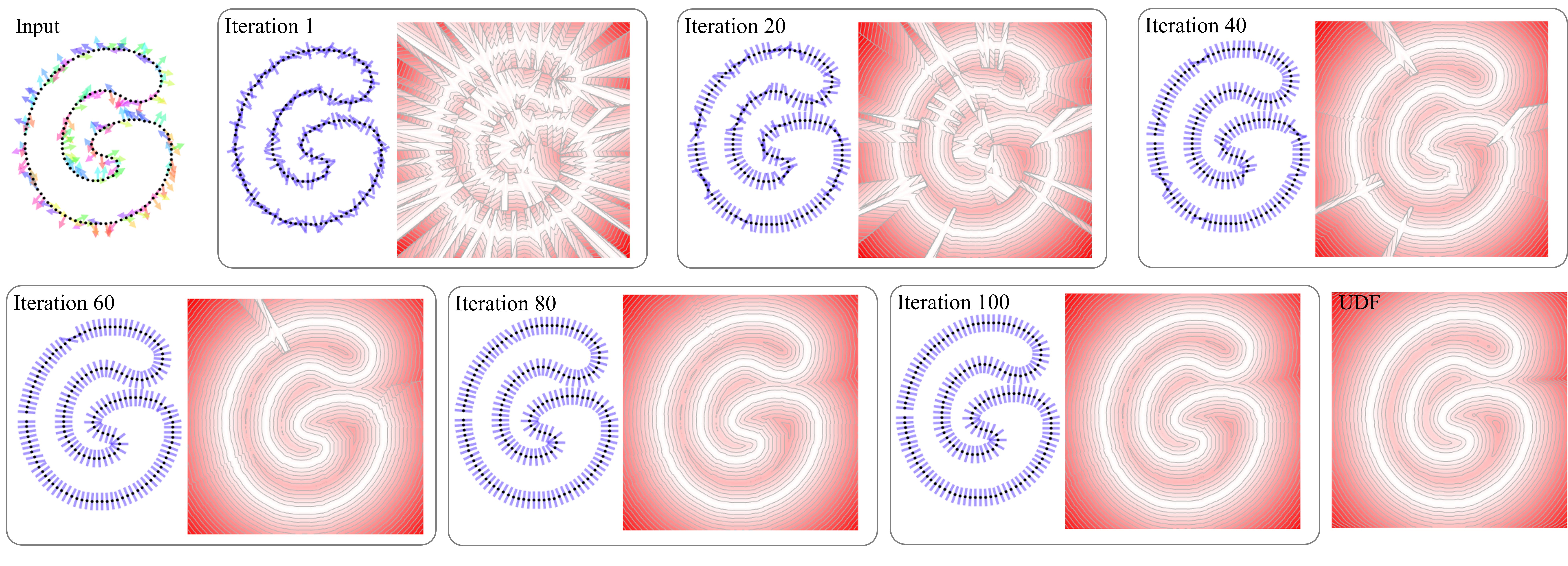}
    \caption{\revise{Illustration of VAD on a 2D toy example. At each iteration, we visualize the bi-directional vectors at input points and the resulting projection field. As optimization progresses, the vectors become better aligned and the field more consistent.
    }}
    \label{fig:energy}
\end{figure*}

We formalize the \emph{bi-directional normal alignment problem} by minimizing two energy terms:

\begin{eqnarray}\label{eq:twoterms1}
E_d &=& \int_{\mathcal{B}} \left|F^i_{\mathcal{B}_{ij}}(\mathbf{x}) - F^j_{\mathcal{B}_{ij}}(\mathbf{x})\right|\,d\mathcal{B}, \\
\label{eq:twoterms2}
E_g &=& \int_{\mathcal{B}} \left\|\nabla F^i_{\mathcal{B}_{ij}}(\mathbf{x}) - \nabla F^j_{\mathcal{B}_{ij}}(\mathbf{x})\right\|\,d\mathcal{B},
\end{eqnarray}
where $\mathbf{x}$ is a point on bisector $\mathcal{B}_{ij}$, $F^i_{\mathcal{B}_{ij}}(\mathbf{x})$ is the projection distance from $\mathbf{x}$ to site $\mathbf{p}_i$, and $F^j_{\mathcal{B}_{ij}}(\mathbf{x})$ is the corresponding value with respect to site $\mathbf{p}_j$. 

Minimizing $E_d$ enforces consistency of scalar values, while minimizing $E_g$ encourages gradient smoothness. 
The combined objective is
\begin{equation}\label{eq:energy}
E = E_d + \lambda E_g,
\end{equation}
with $\lambda$ balancing the two terms.}

\revise{\subsection{Discussions}
\label{discussions}
Our method builds on Voronoi partitioning and projection distance fields, which together address the challenges of orientation ambiguity and global consistency in UDF computation. The main advantages can be summarized as follows:  
\begin{itemize}[leftmargin=1.2em]
 \item \textbf{Smooth UDF field.}  
Minimizing discontinuities across bisectors yields consistently aligned bi-directional normals, even without signed information. The subsequent diffusion step propagates these corrected orientations across the entire domain, producing a coherent UDF gradient field. 
Solving the Poisson equation to produce a smooth and well-behaved UDF.
   
    \item \textbf{Efficient optimization.}  
    Since the projection distance field is linear within Voronoi cells, both energy terms vanish inside cells and are only evaluated on bisectors. This reduces the computational domain dramatically, making optimization efficient without requiring dense sampling.

 \item \textbf{Controllable handling of ambiguities.}  In point cloud reconstruction, the absence of normals leads to inherent orientation ambiguity, which cannot always be resolved automatically (e.g., deciding whether two nearby components should connect). Our method makes this process more controllable: once normals are specified, either through automatic initialization or interactive adjustment, the Voronoi-assisted optimization ensures consistency, and the subsequent diffusion yields a geometrically and topologically reasonable UDF. This controllability is a key advantage over other UDF- or GWN-based approaches, where such ambiguities are often hidden within the model.

    \item \textbf{Topology flexibility.}  
    Unlike many orientation techniques that rely on inside–outside classification, such as iPSR~\cite{hou2022iterative}, PGR~\cite{lin2022surface}, GCNO~\cite{xu2023globally}, BIM~\cite{liu2024consistent}, DWG~\cite{liu2025dwg}, and WNNC~\cite{Lin2024fast}, our method directly optimizes orientations through \rednote{Voronoi-assisted energy minimization}. As a result, it is robust to a wide range of scenarios, including watertight surfaces, open surfaces, non-manifold, and non-orientable geometries.



\end{itemize}

\rednote{\paragraph{Remark} The projection field was introduced by ~\citep{hoppe1992surface}, who proposed computing a signed distance field from the projection field.} Building on this insight, our method further exploits the intrinsic connection between projection distance fields and Voronoi diagrams, formulating a robust strategy to optimize bi-directional normals.
Unlike Hoppe’s local PCA-based normal estimation, our approach leverages global geometric information, making it more resilient to sparse sampling and non-manifold structures, as confirmed by our experimental results in Section~\ref{sec:results}. Moreover, while Hoppe et al. constructed an SDF directly from the distance field, we adopt a heat-based formulation that diffuses the normal field and then solves a Poisson equation, producing a smoother and more stable unsigned distance field. A detailed algorithmic discussion is provided in Sections~\ref{sec:diffuseUDF} and~\ref{sec:udf-computation}.}

\section{Algorithm}
\label{sec:algorithm}

\subsection{Overview}
\label{subsec:overview}

As discussed in Section~\ref{subsec:connectionstovoronoi}, it is sufficient to evaluate $E_d$ and $E_g$ only along Voronoi bisectors. We therefore compute the Voronoi diagram of the input points and uniformly sample along the bisectors. These samples serve as the loci where consistency conditions are enforced. \rednote{Each input point $\mathbf{p}_i$ is initialized with a random bi-directional vector $\widetilde{\mathbf{v}}_i$.  }

\revise{The core VAD algorithm then proceeds in three stages:  
(1) optimizing the bi-directional normals at input points to achieve consistent alignment,  
(2) diffusing the aligned normals over the domain to construct a smooth vector field that approximates the UDF gradient, and  
(3) solving a Poisson equation based on this vector field to reconstruct the UDF.

\begin{figure}[htb]
    \centering    \includegraphics[width=0.98\linewidth]{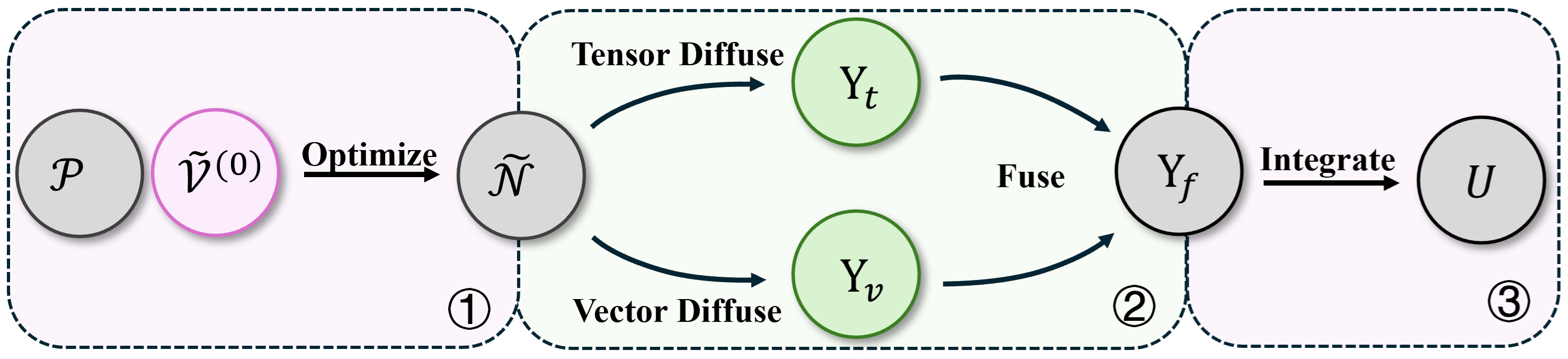}
    \caption{\revise{Conceptual illustration of the VAD pipeline. Starting from randomly initialized bi-directional vectors $\widetilde{\mathcal{V}}^{(0)}$, we optimize them by minimizing the energy $E$ to obtain well-aligned bi-directional normals $\widetilde{\mathcal{N}}$. The aligned normals are then diffused across the domain to construct a smooth vector field that approximates the UDF gradient. 
    The final UDF is recovered by integrating the approximated gradient field via solving a Poisson equation.
    }}
    \label{fig:pipeline}
\end{figure}

To further improve robustness, especially for noisy inputs, we introduce an alternating optimization strategy that jointly refines point positions and normals. This process enforces bi-directional vectors  to remain orthogonal to the underlying surface while simultaneously reducing input noise.}

Figure~\ref{fig:pipeline} illustrates the overall conceptual pipeline and data flow of our approach, while Figure~\ref{fig:energy} visualizes the iterative optimization on a 2D toy model, where bi-directional normals are progressively refined into well-aligned orientations as the energy converges. The corresponding pseudocode is given in Algorithm~\ref{alg:alg1}.

\revise{\subsection{Bi-directional Normal Optimization}\label{subsec:optimization}

\paragraph{Energy Discretization}
As discussed in Section~\ref{subsec:connectionstovoronoi}, the total energy $E$ is evaluated by integrating over Voronoi bisectors. We approximate this integral using discrete sampling. After computing the Voronoi diagram of the input points, we uniformly sample points $\mathcal{X}$ along the bisectors via Poisson sampling. Each bisector is a convex polygon, and the samples are distributed uniformly across its surface. For each sample $\mathbf{x}_k\in\mathcal{X}$, we assign a weight $w(\mathbf{x}_k)$ by dividing the polygonal face area by the number of samples on it. Using these samples, Equations~(\ref{eq:twoterms1}) and (\ref{eq:twoterms2}) can be approximated as:\rednote{
\begin{eqnarray}\label{eq:twoterms1new}
E_d &=&\frac{1}{h^2 } \sum_{k=1}^{|\mathcal{X}|} w(\mathbf{x}_k)\left|F_{\mathcal{B}_{ij}}^{i}(\mathbf{x}_k) - F_{\mathcal{B}_{ij}}^{j}(\mathbf{x}_k)\right|,\\
\label{eq:twoterms2new}
E_g &=& \sum_{k=1}^{|\mathcal{X}|} w(\mathbf{x}_k)\left\|\nabla F_{\mathcal{B}_{ij}}^{i}(\mathbf{x}_k) - \nabla F_{\mathcal{B}_{ij}}^{j}(\mathbf{x}_k)\right\|.
\end{eqnarray}}

\paragraph{Alignment Regularizer} To stabilize the optimization and accelerate its convergence, we introduce an alignment regularizer that 
\begin{wrapfigure}{r}{0.15\textwidth}
   \centering 
   \vspace{-1.5em}
   \hspace{2em}
  \makebox[40pt][r]{\includegraphics[width=0.15\textwidth]{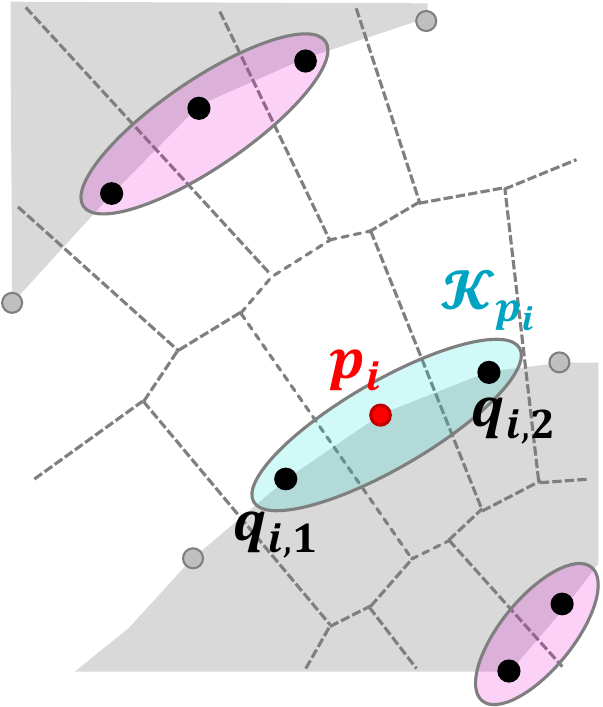}}
\end{wrapfigure}
encourages bi-directional normals to remain orthogonal to the locally inferred surface. For each input point $\mathbf{p}_i$, we identify its neighboring points \rednote{via Voronoi connectivity and partition them into two groups by applying $k$-means ($k=2$) clustering solely based on their Euclidean distances to the current point, effectively reducing the problem to a one-dimensional clustering along the distance axis}. The cluster whose centroid is closest to $\mathbf{p}_i$ is selected as the local neighbor set, denoted by $\mathcal{K}_{\mathbf{p}_i}$  (the cyan region in the inset). The alignment regularizer is then defined as:\rednote{
\begin{equation}
    E_{\text{align}} = \sum_{i=1}^{n} \sum_{\mathbf{q}_j\in\mathcal{K}_{\mathbf{p}_i}} \left(1-\mathbf{v}_i\cdot \frac{\mathbf{q}_j - \mathbf{p}_i}{\|\mathbf{q}_j - \mathbf{p}_i\|}\right)^2,
\end{equation}} where $\mathbf{q}_j\in\mathcal{K}_{\mathbf{p}_i}$ denotes a neighboring point of $\mathbf{p}_i$.

\paragraph{Optimization}
The overall objective for normal optimization is :
\begin{equation}
E_{\text{normal}} \;=\; 
\lambda_d E_d \;+\; \lambda_g E_g \;+\;\lambda_a E_{\text{align}}.
\label{eqn:discrete-energy-1}
\end{equation} We solve this optimization problem using gradient descent with the Adam optimizer~\cite{kinga2015method} in PyTorch. 

\subsection{Point Position Optimization (Optional)}\label{sec:denoise}

When the input points are noisy and deviate from the true surface, optimizing only the bi-directional normals is insufficient to compute a reliable UDF. In stage one (bi-directional normal optimization on Voronoi bisectors), the method is relatively robust to small or moderate noise, as Voronoi cells still encourage normals to align roughly with the underlying surface. In stage two (bi-directional normal diffusion), however, splitting and propagating normals amplify local inconsistencies, leading to an unreliable gradient field and consequently degrading the UDF quality. 
\begin{figure}[htb]
    \centering
    \includegraphics[width=0.99\linewidth]{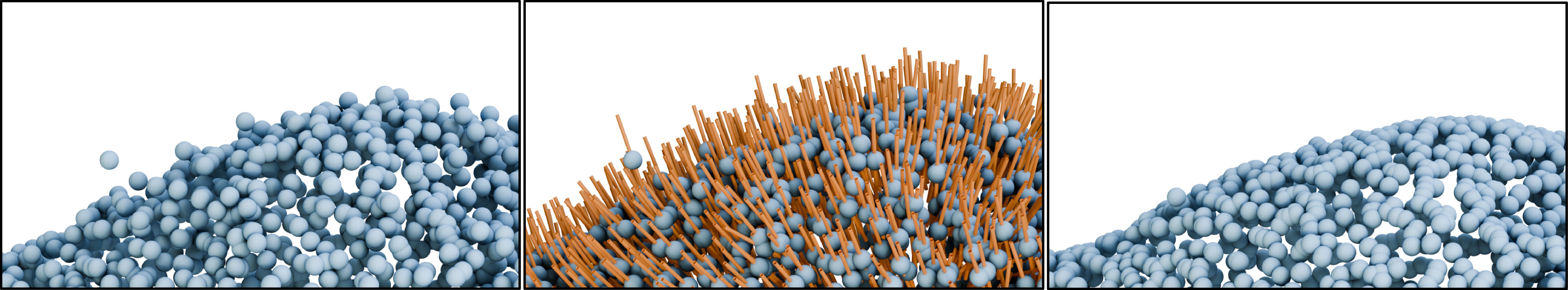} 
    \caption{Handling noisy input. From left to right: Input points corrupted with 0.5\% Gaussian noise, bi-directional normals estimated from the noisy points, and points after displacement optimization.} 
    \label{fig:denoise-3d}
\end{figure}

To address this issue, we introduce an optional \emph{position rectification step}. After normal optimization, we keep orientations fixed and displace points along their normals by offsets $\delta$, chosen to minimize the projection field energy. This adjustment flattens and smooths the local geometry of the point cloud. The energy function is defined as:
\begin{equation}
E_{\text{offset}}(\delta)
= \lambda_d E_d
+ \lambda_a E_{\text{align}}
+ \lambda_p E_{\text{reg}}(\delta),
\label{eqn:discrete-energy-2}
\end{equation}
where $E_d$ and $E_{\text{align}}$ take the same form as in the normal optimization stage but are evaluated at updated positions. 
\rednote{Since the offsets $\delta$ are reset in each outer alternating loop, relying solely on a local magnitude penalty cannot prevent progressive spatial drift over multiple iterations. Therefore, we design $E_{\text{reg}}$ to incorporate two distinct safeguards: a global anchor and a local damping term. The regularization energy is formulated as:
\begin{equation}
\label{eqn:displacementregularizer}
E_{\text{reg}}(\delta) = \frac{1}{n} \sum_{i=1}^n \left\| (p_i + \delta_i v_i) - p_i^{(0)} \right\|^2+\; \frac{1}{n} \sum_{i=1}^n \delta_i^2.
\end{equation}

The first term acts as a global spring constraint. It measures the deviation of the currently updated position from the raw, unoptimized noisy input $p_i^{(0)}$, effectively tying the geometry to the original data and preventing long-term drift. The second term restricts the magnitude of the local displacement $\delta_i$, which prevents excessively large jumps and ensures numerical stability within the current inner optimization step.}

\rednote{
Our method adopts a nested optimization framework: an outer loop alternating between offset updates and normal refinement (which necessitates recomputing the Voronoi diagram), and inner iterative solvers ensuring local convergence for each sub-problem. This alternating process terminates when the normal field stabilizes—specifically, when the average of the top $10\%$ largest normal variations between consecutive outer iterations falls below a threshold (default: $0.3$).To demonstrate the effectiveness of our positional rectification, Figures~\ref{fig:denoise} and~\ref{fig:denoise-3d} visualize the inner offset optimization across 2D and 3D domains. The progressive geometric refinement within a single outer iteration shows that localized normal-driven displacements successfully suppress noise and improve projection consistency. Additionally, Figure~\ref{fig:pos-change} details the spatial evolution of point coordinates and the corresponding convergence curve. This highlights the rapid stabilization of the offset updates, validating their ability to produce smooth and consistent geometry from noisy input data.}

\begin{figure}[htb]
    \centering
    \includegraphics[width=0.99\linewidth]{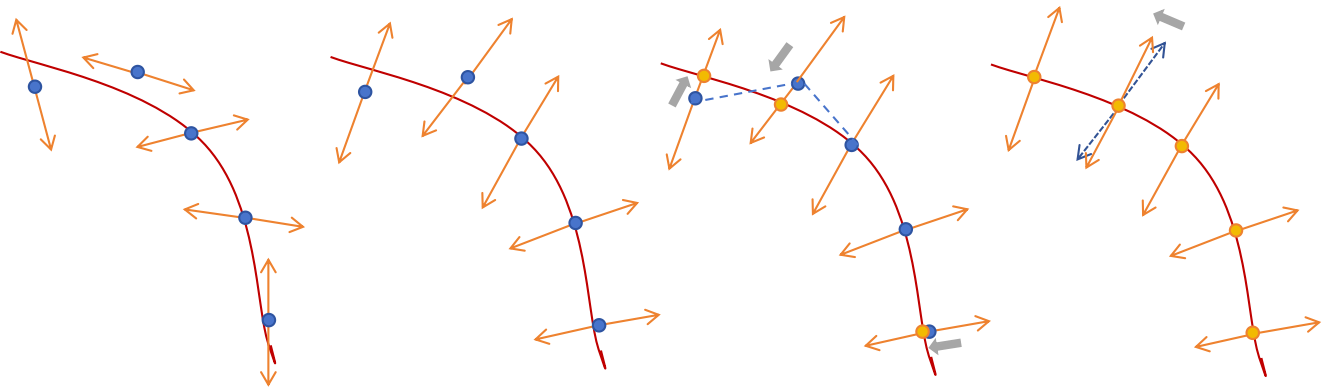}
    \makebox[0.24\linewidth]{(a) }\makebox[0.24\linewidth]{(b) }\makebox[0.24\linewidth]{(c) }\makebox[0.24\linewidth]{(d) }
    \caption{\revise{A 2D illustrative example of handling noisy input. The ground-truth smooth surface is drawn in red and the noisy samples in blue. (a) Noisy points with randomly initialized normals. (b) Result after the bi-directional normal optimization. (c) With normals fixed, point positions are optimized by allowing displacement along their normal directions. (d) With positions fixed, bi-directional normals are further refined for more accurate orientations. The overall procedure demonstrates how noisy points are iteratively adjusted through normal-guided offsets to yield rectified positions.}}
    \label{fig:denoise}
\end{figure}
\begin{figure}[htb]
    \centering
    \includegraphics[width=\linewidth]{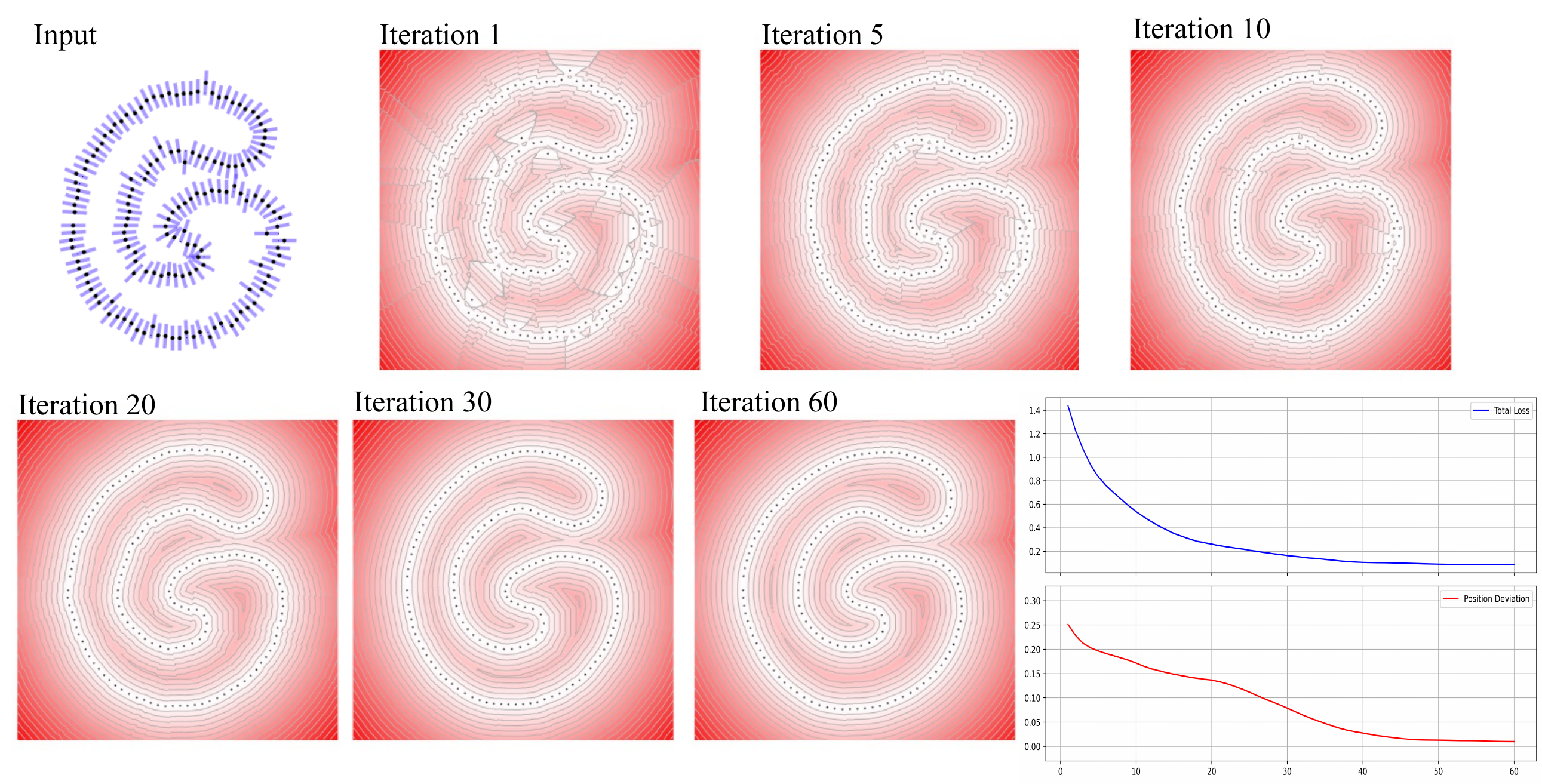}
    \includegraphics[width=\linewidth]{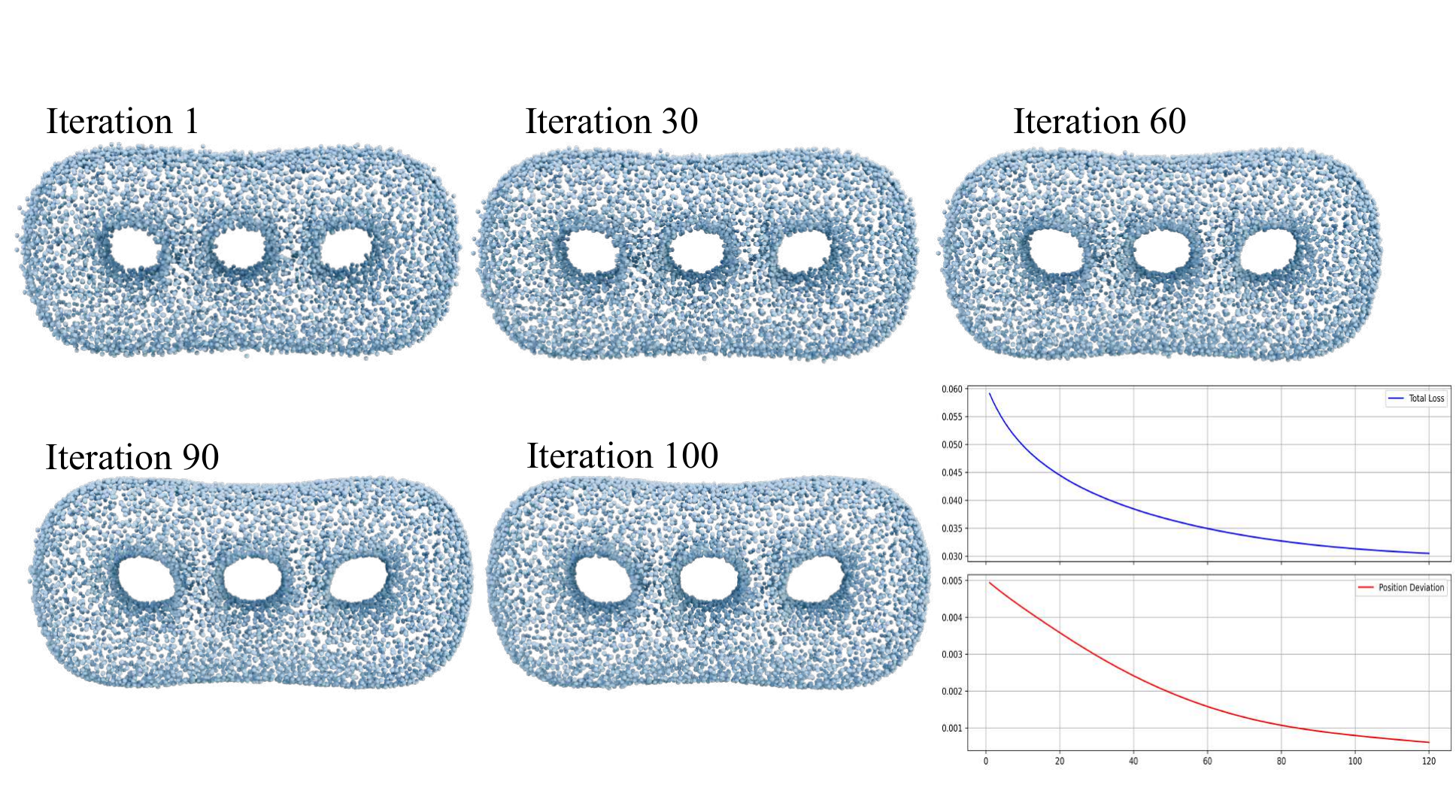}
    \caption{\rednote{Illustration of VAD on both 2D/3D toy examples. For the 2D example, we visualize the displacement of input points and the corresponding projection field at different iteration, while keeping the input normals fixed. For the 3D example, we visualize the positions of 3D point at different iterations. As optimization proceeds, noisy point positions gradually move toward a more aligned configuration, resulting in a cleaner and more consistent projection field.}}
    \label{fig:pos-change}
\end{figure}
\subsection{Normal Diffusion}\label{sec:diffuseUDF}

\paragraph{Motivations} Directly using the projection distance field works well for dense, uniform, and clean point sets, where the field is smooth and suitable for downstream applications. However, for sparse or noisy inputs, the projection distance field often exhibits discontinuities or non-smooth transitions across Voronoi cell boundaries (see Figure~\ref{fig:fields_discuss}(a)). To overcome these limitations, we adopt a more robust strategy: diffusing the computed bi-directional normals over the entire domain. \rednote{Specifically, we discretize the continuous computational domain bounding the point cloud into a uniform Cartesian grid. A default grid resolution of $128^3$ is used throughout the paper unless stated otherwise.} The UDF is subsequently recovered from this discretized diffused vector field via numerical integration (see Figure~\ref{fig:fields_discuss}(b)).

\begin{figure}[!htbp]
    \centering
    \includegraphics[width=0.475\linewidth]{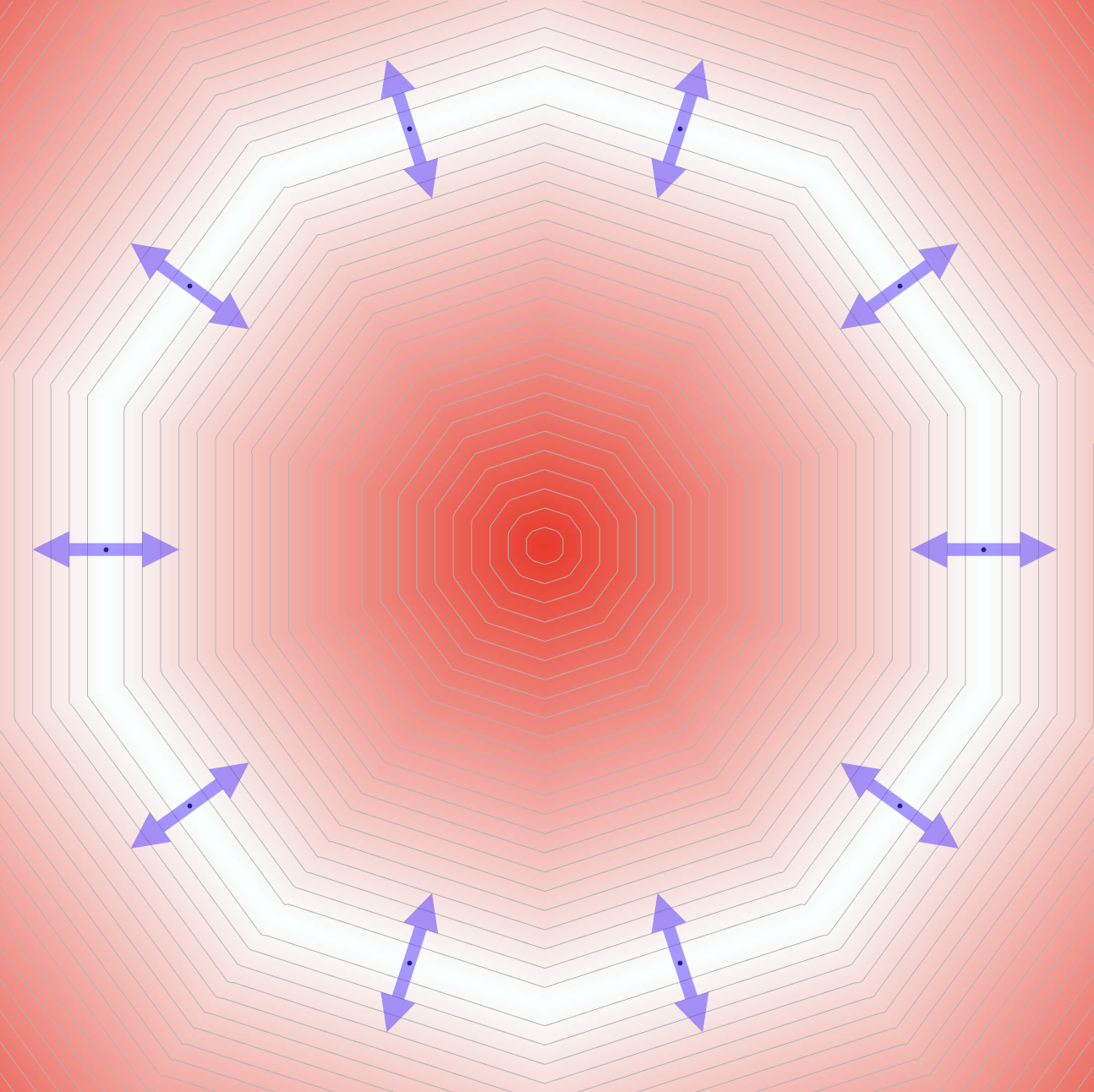}
\includegraphics[width=0.475\linewidth]{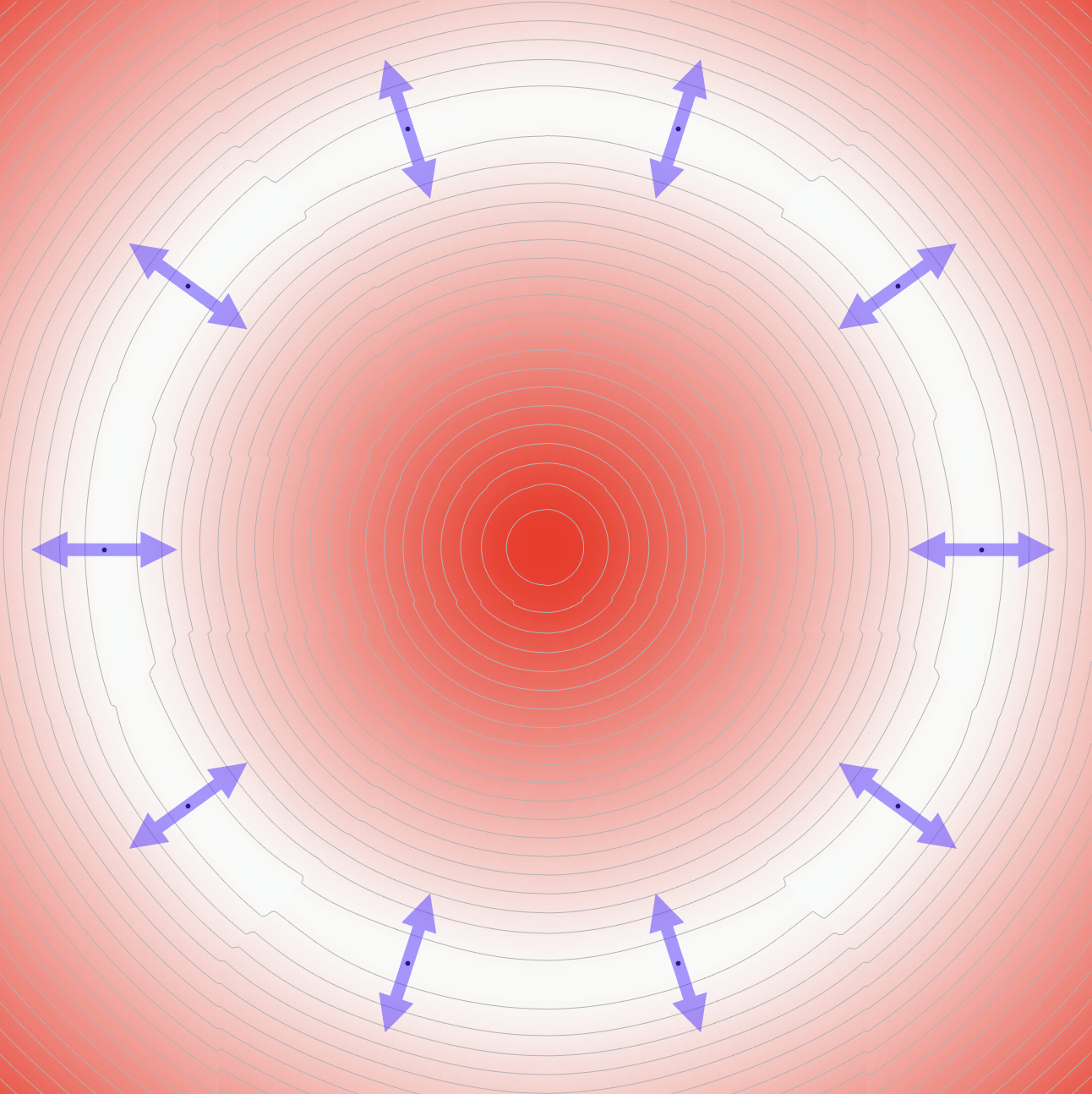}
    \makebox[0.475\linewidth]{(a)}\makebox[0.475\linewidth]{(b)}
    \caption{\revise{Motivation for using diffusion-based method in recovering UDFs. Consider 10 points uniformly distributed on a circle. (a) Due to the low sampling density, the projection distance field computed from well-aligned bi-directional normals exhibits non-smooth variations near the decagon vertices. (b) By diffusing the bi-directional normals and integration, we obtain a UDF with improved smoothness.}}
    \label{fig:fields_discuss}
\end{figure}

\begin{figure}[!htbp]
    \centering
    \includegraphics[width=0.99\linewidth]{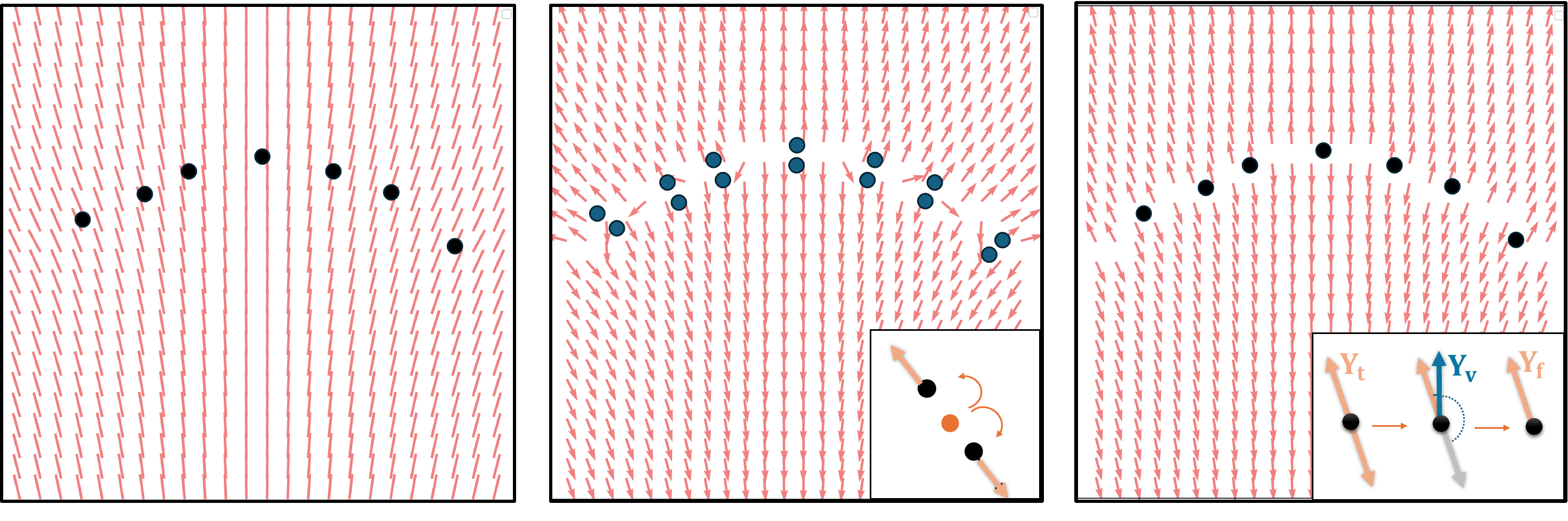}
    \makebox[0.33\linewidth]{(a) Tensor diffusion}\makebox[0.33\linewidth]{(b) Vector diffusion}\makebox[0.33\linewidth]{(c) Field fusion }
    \caption{\revise{A 2D illustrative example of constructing a smooth vector field  $\mathbf{Y}_f$ as an approximation of the UDF gradient. (a) Each bi-directional normal at an input point is represented as a rank-one tensor and diffused over the domain to obtain a smooth tensor field $\mathbf{Y}_t$. (b) Each bi-directional normal is split into two opposite directions, which are diffused to obtain a smooth vector field $\mathbf{Y}_v$. (c) At each point $\bf x$, the principal axis extracted from $\mathbf{Y}_t(\mathbf{x})$ is oriented by aligning it with  $\mathbf{Y}_v(\mathbf{x})$, yielding a consistent vector field. The fused result provides $\mathbf{Y}_f$, which serves as an approximation of the UDF gradient.}}
    \label{fig:correction}
\end{figure}

\paragraph{Tensor Diffusion} Inspired by the heat method~\cite{sharp2019vector, feng2024heat}, which employs diffusion to robustly extrapolate orientation information, we diffuse a \textit{tensor} representation of bi-directional normals to obtain a smooth and globally consistent direction field. Each bi-directional vector $\widetilde{\mathbf{n}}(\mathbf{x})$ is represented as a rank-one symmetric tensor:
\[
\mathbf{T}(\mathbf{x}) = \mathbf{n}(\mathbf{x}) \otimes \mathbf{n}(\mathbf{x}).
\]
This formulation naturally encodes bi-directional orientation, since $\mathbf{n}$ and $-\mathbf{n}$ yield the same tensor. We apply heat diffusion~\cite{sharp2019vector, feng2024heat} to each tensor entry, solving a heat equation of the form
\[
\left(\Delta - \frac{1}{t}\right) \mathbf{Y}_t = -\frac{1}{t} \mathbf{T},
\]
where $\Delta$ is the Laplace–Beltrami operator and $t$ is the diffusion time. Intuitively, $1/t$ controls the scale of diffusion: small $t$ yields more local propagation, while large $t$ produces more global smoothing This process smoothly propagates local orientation information across the domain while suppressing noise. The resulting tensor field $\mathbf{Y}_t$ is continuous and smooth. The principal axis directions are then extracted via eigen-decomposition of $\mathbf{Y}_t$. 

\paragraph{Vector Diffusion}  
Although the principal axis can be recovered from the diffused tensor field, its orientation remains ambiguous. To resolve this, we diffuse duplicated bi-directional normals to propagate consistent directional information. Specifically, each input point is split into two perturbed copies shifted by $\pm \epsilon$ along its bi-directional normal, giving opposite orientations. These perturbed samples serve as directional constraints. Similar to the tensor diffusion step, here we apply heat diffusion~\cite{sharp2019vector, feng2024heat} to the vector field, solving a screened Poisson equation of the form
\[
\left(\Delta - \frac{1}{t}\right)\mathbf{Y}_v = -\frac{1}{t}\mathbf{N},
\] \rednote{where $\mathbf{N}$ is the vector field formed by the perturbed normals.} This process robustly propagates directional information across the domain while suppressing noise. The result is a smooth vector field $\mathbf{Y}_v$ whose orientations are globally consistent but not guaranteed to remain strictly orthogonal near the surface.  

\paragraph{Field Fusion} The diffused vector field $\mathbf{Y}_v$ preserves directional information but often fails to maintain orthogonality near the surface, especially for sparse inputs. In contrast, the tensor field  $\mathbf{Y}_t$ better preserves orthogonality but loses directional consistency, since both $\bf n$ and $-\bf n$ produce the same tensor. It is therefore natural to combine these two fields to approximate the gradient vector field. Specifically, at each point $\mathbf{x}$, we extract the principal axis from $\mathbf{Y}_t$ and compare it with the vector $\mathbf{Y}_v(\mathbf{x})$. If the two vectors form an angle less than $90^\circ$, we keep the principal axis as is; otherwise, we flip it. The aligned principal axes form a smooth vector field $\mathbf{Y}_f$, which is perpendicular to the surface and consistently oriented across both interior and exterior regions. The complete diffusion and fusion procedure is illustrated in Figure~\ref{fig:correction}.

\subsection{UDF Computation}
\label{sec:udf-computation}
After the fusion step, the vector field $\mathbf{Y}_f$ can be regarded as an approximation of the gradient of the UDF. To ensure a well-posed solution and maintain consistency with the input, we impose Dirichlet boundary conditions $u(\mathbf{p}_i)=0$ at all surface samples $\mathbf{p}_i$. 

We then reconstruct the scalar UDF $u$ by solving a Poisson equation that enforces $\nabla u \approx \mathbf{Y}_f$. Concretely, we minimize the quadratic energy
\[
E(u) = \int_{\Omega} \left\|\nabla u - \mathbf{Y}_f\right\|^2 \, d\Omega ,
\]
whose Euler–Lagrange equation yields
\[
\Delta u = \nabla \cdot \mathbf{Y}_f \quad \text{in } \Omega,
\]
subject to Dirichlet boundary conditions $u(\mathbf{p}_i)=0$. 

This formulation is identical to that of~\cite{feng2024heat}, which reconstructs a signed distance field by diffusing oriented normals. In our case, the same principle applies to unsigned distances: the fused vector field $\mathbf{Y}_f$ provides a globally consistent gradient approximation, and solving the Poisson equation yields a smooth UDF defined over the domain.}

\begin{algorithm}
  \caption{Voronoi-assisted optimization for diffusing UDFs from unoriented points}\label{alg:alg1}
  \begin{algorithmic}[1]        
    \Require Unoriented points $\mathcal{P}=\{\mathbf{p}_i\}_{i=1}^{n}$; denoising option (boolean)
    \Ensure Aligned bi-directional normal 
     $\widetilde{\mathcal{N}}$; unsigned distance field $u$
        \Statex \textbf{Initialization}
     \State Construct the 3D Voronoi diagram of $\mathcal{P}$.
    \State Initialize bi-directional normals $\widetilde{\mathcal{V}}^{(0)}$ randomly.
    \State Uniformly sample Voronoi bisectors $\mathcal{B}_{ij}$.
    \Statex \textbf{Bi-directional Normal Optimization}        
    \State Minimize the energy $E_{\text{normal}}(\widetilde{\mathcal{V}})$ to obtain bi-directional normal $\widetilde{\mathcal{N}}$.
     
     \rednote{\Statex \textbf{(Optional) Alternating Position-Normal Optimization}
    \If {denoising == true}
        \While {not converged}
            \State Fix $\widetilde{\mathcal{N}}$ and optimize offsets $\delta$ to minimize $E_{\text{offset}}$.
            \State Update point positions using $\delta$ and recompute Voronoi diagram.
            \State Fix updated positions and re-optimize $\widetilde{\mathcal{N}}$ to minimize $E_{\text{normal}}$.
        \EndWhile
    \EndIf}
 \Statex \textbf{Bi-directional Normal Diffusion}
\State Diffuse the tensor field $\mathbf{T}$ to obtain $\mathbf{Y}_t$.
\State Diffuse the perturbed normals to obtain $\mathbf{Y}_v$.
\State Extract principal axes from $\mathbf{Y}_t$ and orient them using $\mathbf{Y}_v$.
\State Fuse the results to obtain the consistent vector field $\mathbf{Y}_f$.
 \Statex \textbf{UDF Computation}
\State Solve the Poisson equation $\Delta u = \nabla \cdot \mathbf{Y}_f$ with Dirichlet boundary conditions $u(\mathbf{p}_i)=0$, $1\leq i\leq n$, to recover $u$.   
  \end{algorithmic}
\end{algorithm}

\section{Experimental Results}
\label{sec:results}

\subsection{Setup}\label{subsec:setup}

\paragraph{Implementation}
We implement VAD in C++ and Python with PyTorch and evaluate it on a workstation with an Intel Core i9-10900 CPU, 32 GB of RAM, and an NVIDIA RTX 4090 GPU with 24 GB of VRAM (CUDA 11.8). For bi-directional normal optimization, we use the Adam solver~\cite{kinga2015method} with a maximum of 1000 iterations; although convergence is typically achieved well before this limit. 
\rednote{To enforce unit-length normals, we optimize unconstrained vectors $\mathbf{v}_i$ and obtain the normals used in the objective via explicit normalization at each iteration:
$
\mathbf{v}_i=\frac{\mathbf{v}_i}{\|\mathbf{v}_i\|_2}.
$} All point cloud models are uniformly scaled to the range $[-0.4, 0.4]^3$ to ensure that parameter values are independent of the model scale. The weighting parameters are empirically \rednote{fixed to $\lambda_d = 1$, $\lambda_g = 10^{-3}$, $\lambda_a = 10^2$ for clean inputs and $\lambda_d = 1$, $\lambda_g = 10^{-2}$, $\lambda_a = 10^2$, $\lambda_p = 10^2$ for noisy inputs. }
Parameters $t$ and $\epsilon$ are also fixed. Let $h$ be the mean nearest-neighbor distance in the input point cloud, $t=h^2$ and $\epsilon = 10^{-4}h$. To extract zero-level sets from the computed UDFs, particularly for complex geometries and non-manifold topologies, we adopt DCUDF~\cite{hou2023dcudf}, which generates a double-covered mesh that tightly wraps the target surface. \rednote{The extraction offset for the double-covered surface is set to $0.005$ when using grid resolutions of $128^3$ or $256^3$. We note that the proposed VAD framework is independent of the downstream meshing stage.
We further evaluate the compatibility of VAD with different mesh extraction methods in the Appendix.
}
\begin{table}[t]
\centering
\resizebox{0.48\textwidth}{!}{
\begin{tabular}{c|ccc}
\hline
\makecell[c]{\textbf{Method}}  & \textbf{Field Type}  & \textbf{Topology}   & \textbf{Point Orientation} \\
\hline
GCNO, WNNC, DWG & GWN &  Watertight & \makecell[c]{Globally consistent} \\
\hline
SNO & Poisson &  Watertight & \makecell[c]{Globally consistent} \\
\hline
Hoppe et al. & SDF*  &  \makecell[c]{Watertight/Open} & \makecell[c]{Globally consistent} \\
\hline
 DACPO & Poisson &  \makecell[c]{Watertight/Open} & \makecell[c]{Globally consistent} \\
\hline
\makecell[c]{CAPUDF, GeoUDF,\\DEUDF, DUDF, LoSF} & UDF & \makecell[c]{Watertight/Open/\\Non-manifold/Non-orientable} & Not reliable \\
\hline
VAD (Ours) & UDF & \makecell[c]{Watertight/Open/\\ Non-manifold/Non-orientable} & \makecell[c]{Bi-directional}  \\
\hline
\end{tabular}}
\caption{\revise{Summary of baselines used in our experiments. Existing deep learning-based UDF methods focus on learning UDFs directly from raw points. However, due to non-differentiability at zero level sets, they cannot reliably compute point orientations from UDF gradients. Our method retains the flexibility of UDFs for handling arbitrary geometries while also enabling normal computation, similar to SDF-based approaches. (SDF*: tangent-plane approximation of SDF near the surface.)}}
    \label{tab:summary-total}
\end{table}
\begin{figure*}[t]
    \centering
    \includegraphics[width=0.95\linewidth]{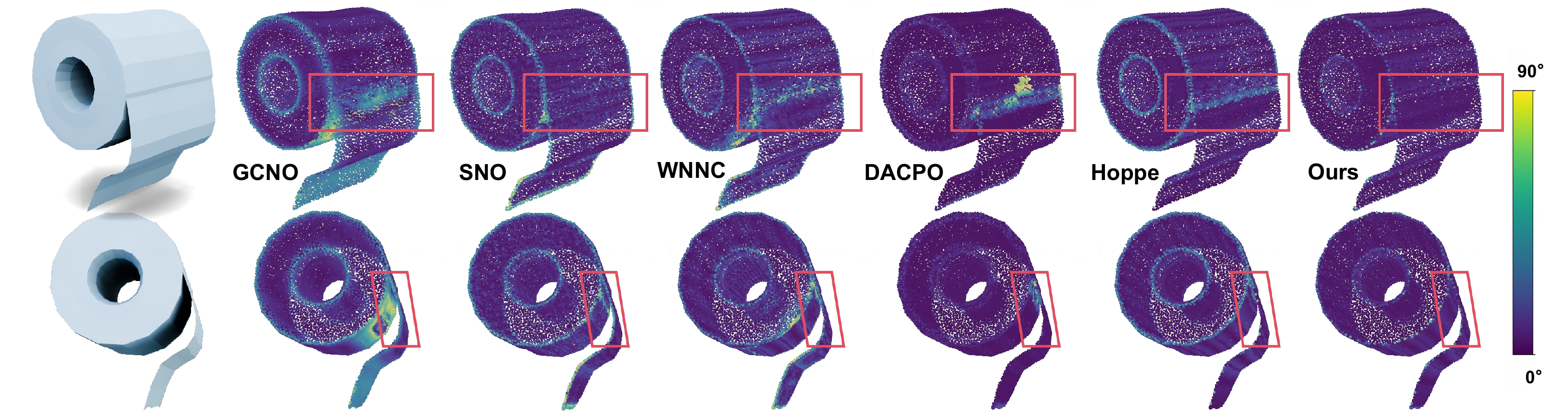}
\footnotesize
        \makebox[0.12\linewidth]{NC (all points):}
    \makebox[0.12\linewidth]{0.9465}
    \makebox[0.12\linewidth]{0.9788}
    \makebox[0.12\linewidth]{0.9739}
    \makebox[0.12\linewidth]{ 0.9835}
    \makebox[0.12\linewidth]{0.9887}
\makebox[0.12\linewidth]{0.9922}\\
\makebox[0.12\linewidth]{local NC (non-manifold):}
    \makebox[0.12\linewidth]{0.8696}
    \makebox[0.12\linewidth]{0.9638}
    \makebox[0.12\linewidth]{0.9313}
    \makebox[0.12\linewidth]{ 0.5412}
    \makebox[0.12\linewidth]{0.9755}
\makebox[0.12\linewidth]{0.9911}\\
\includegraphics[width=0.95\linewidth]{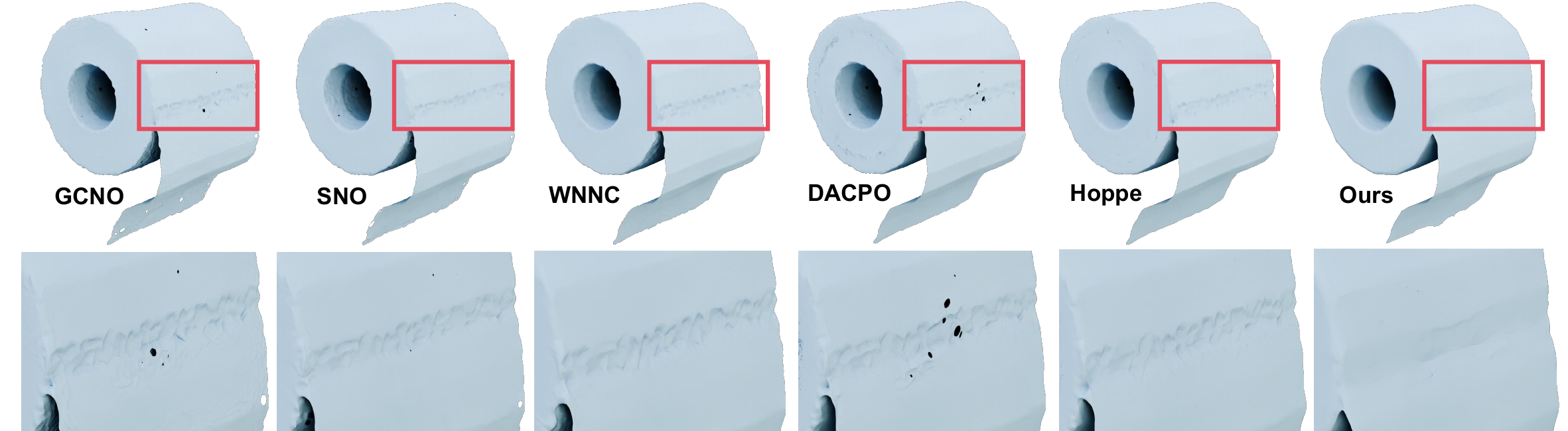}
    \caption{\revise{Normal alignment. The first and second rows visualize the angular error between predicted and ground-truth normals on the Toilet-Paper model containing non-manifold structures. Our method achieves higher normal consistency (closer to 1), indicating better alignment. Notably, the improvement is more pronounced in the detected non-manifold regions, where our method produces significantly more accurate and consistent normal predictions. The third and fourth rows show surfaces reconstructed from these normals using the same UDF construction (heat diffusion and integration) and zero-level set extraction (DCUDF). Inaccurate normals lead to artifacts in the reconstructed surfaces, whereas our method produces cleaner and more accurate geometry.}}
    \label{fig:normal-accuracy}\vspace{-0.15in}
\end{figure*}

\paragraph{Baselines}
\revise{For comparison, we consider two representative categories of existing approaches (summarized in Table~\ref{tab:summary-total}). \emph{Techniques for watertight surfaces}, including the classic method of Hoppe et al.~\citep{hoppe1992surface} and more recent methods such as GCNO~\cite{xu2023globally}, SNO~\cite{Huang2024Stochastic}, WNNC~\cite{Lin2024fast}, and DWG~\cite{liu2025dwg}, enforce an interior–exterior distinction to produce globally consistent point orientations. While these methods are effective for watertight surfaces, they struggle with open, non-manifold, or non-orientable geometries, where the lack of a consistent inside–outside distinction leads to orientation ambiguities. Although Hoppe et al.’s method can be extended to open surfaces, it remains limited when applied to non-manifold and non-orientable geometries. 
\emph{Techniques for open surfaces}, including CAPUDF~\cite{Zhou2022CAP-UDF}, GeoUDF~\cite{DBLP:conf/iccv/RenHCHW23}, DUDF~\cite{Fainstein2024DUDF}, DEUDF~\cite{Xu2024DEUDF}, LoSF~\cite{losf-udf-2024}, and DACPO~\cite{li2025divideandconquerapproachglobalorientation}, relax the requirement for a clear interior-exterior distinction. With the exception of DACPO, these are deep learning-based methods that directly fit UDFs from input points and can naturally handle more general geometries. However, they often incur high computational costs and offer limited controllability. In practice, they are typically constrained to point clouds with fewer than 100k points due to large GPU memory demands. DACPO, by contrast, adopts a divide-and-conquer strategy using 0-1 integer programming. By limiting the number of partitioned blocks to a few hundred, it remains efficient for large-scale inputs but still does not support non-manifold or non-orientable geometries. In contrast, our method combines the flexibility of unsigned distance representations with improved controllability and efficiency by leveraging bi-directional normal 
alignment optimization and heat diffusion. 
}

\paragraph{Test Models}
We evaluate our method on a diverse set of 3D models, including cloth models from Deep Fashion3D V2~\cite{zhu2020deep}, which feature open surfaces as well as objects with non-manifold configurations and non-orientable geometries. To ensure a fair comparison with deep learning-based methods, which typically require low-resolution inputs with uniform sampling, we generate point clouds by uniformly sampling 10k to 20k points from the ground-truth meshes. Our method is not restricted to this setting and remains effective for larger models and varied sampling distributions. In the following subsections, we further evaluate its robustness under more challenging scenarios.

\subsection{Normal Alignment}
\label{subsec:normalalignment}
\revise{
To evaluate the robustness of our normal alignment strategy, we conduct experiments on manifold, non-manifold, and non-orientable models.
Figure~\ref{fig:normal-accuracy} (first and second rows) compares our method with existing approaches on the Toilet-Paper model, featuring both boundaries and non-manifold edges.
Accuracy is measured using the \rednote{Normal Collinearity (NC) defined as $ \frac{1}{N} \sum_{i=1}^{N} \left| \frac{\mathbf{n}_i \cdot \mathbf{n}_i^*}{\|\mathbf{n}_i\| \|\mathbf{n}_i^*\|} \right|$, where $\mathbf{n}^{*}$ denotes the ground-truth unit normal; values closer to one indicate better alignment.} As shown in the figure, our method outperforms existing methods, particularly in non-manifold regions where orientation ambiguities occur, and it also better preserves sharp features compared with the baselines. 
Further statistical results are provided in Table~\ref{normal-consistency}, with quantitative comparisons across both watertight and non-manifold models.}

\begin{table}[!htbp]
\centering
\resizebox{0.45\textwidth}{!}{
\begin{tabular}{l|cc|cc}
\hline
Method & 4-Children                      & Bears                  & Toilet-Paper                   & Strawberry                                           \\ \hline
Hoppe et al. & 0.9780                          & 0.9821                         & \cellcolor[HTML]{EFEFEF}0.9887                         & \cellcolor[HTML]{EFEFEF} 0.9710 \\
GCNO   & 0.9624                         & 0.9682                         & 0.9465                         & 0.9479                                               \\
SNO    & 0.9586                         & 0.9661                         & 0.9788                         & 0.9474                                               \\
WNNC   & \cellcolor[HTML]{FFFC9E}0.9902 & \cellcolor[HTML]{FFFC9E}0.9909 & 0.9739                         & 0.9001                                               \\
DWG    & 0.9321                         & 0.9564                         &         0.8740                        &         0.9419                                             \\
DACPO  &        0.9703                        &                       0.9763         & 0.9835 & 0.9581                                               \\ \hline
Ours  & \cellcolor[HTML]{EFEFEF}0.9860  & \cellcolor[HTML]{EFEFEF}0.9873 & \cellcolor[HTML]{FFFC9E}0.9922 & \cellcolor[HTML]{FFFC9E}0.9799      \\\hline                  
\end{tabular}
}
\caption{\revise{Comparison of NC on four representative models. 4-Children and Bears are watertight surfaces, while Toilet-Paper and Strawberry are open surfaces with non-manifold structures. The best results are highlighted in yellow, and the second best in gray. Although primarily designed for non-watertight surfaces, our method also achieves strong performance on watertight models.}}
\label{normal-consistency}
\end{table}

\begin{figure}[!htbp]
    \centering
    \includegraphics[width=\linewidth]{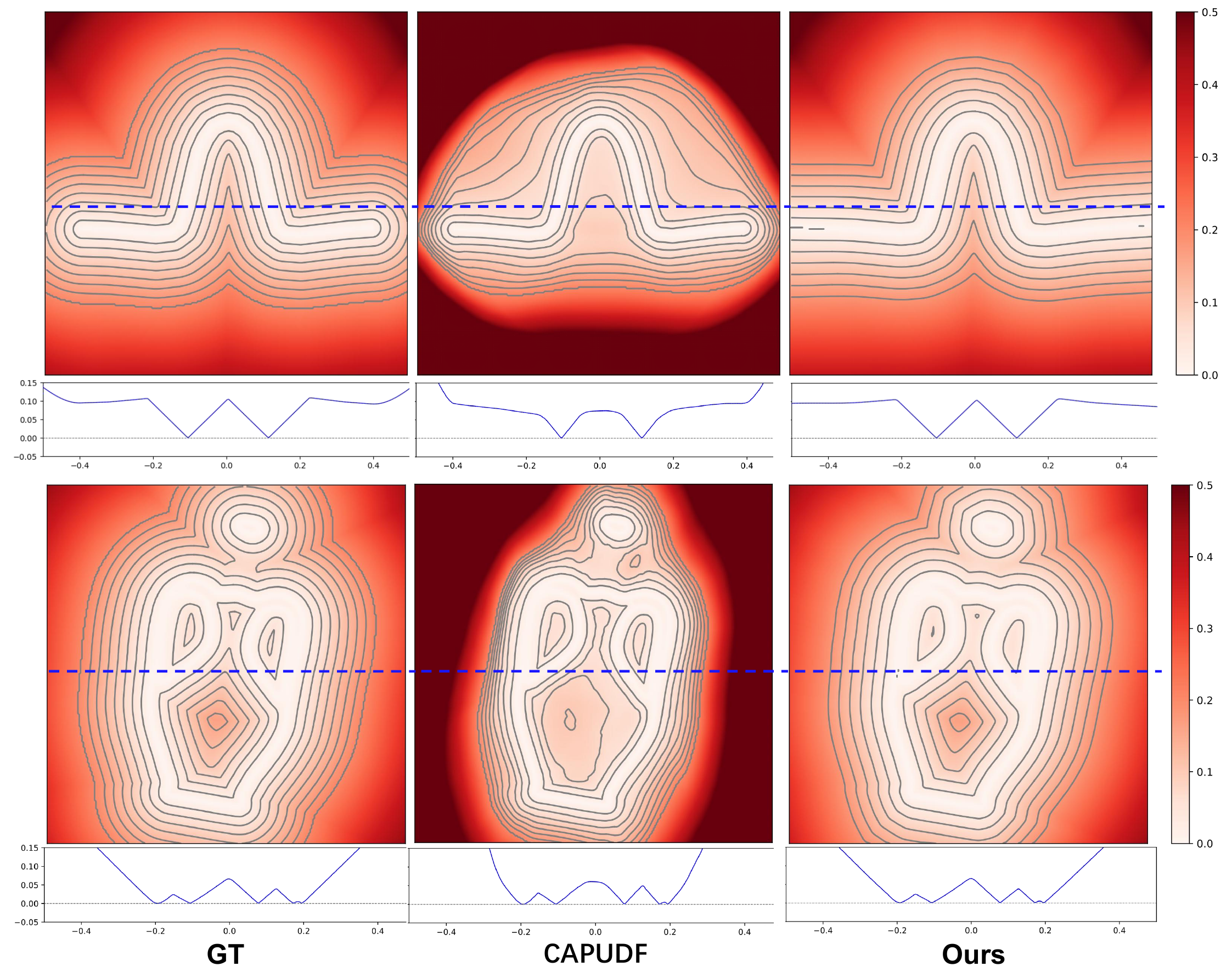}
\caption{Accuracy comparison with CAPUDF~\cite{Zhou2022CAP-UDF}. We show cut views of the computed UDFs and plot distance values along a horizontal line within the cutting plane. } 
    \label{fig:accuracy_color}
\end{figure}

\subsection{Unsigned Distance Fields}
\label{subsec:udfevaluation}

\paragraph{Accuracy}
By adopting the heat method~\cite{feng2024heat} to compute the UDFs through the diffusion of the optimized bi-directional normals, our approach produces UDFs that maintain smoothness and isotropy (i.e., unit-gradient magnitude) throughout the domain.
Figure~\ref{fig:accuracy_color} shows slices of the UDF computed from the Hat model (an open surface) and the Bottle model (a watertight surface) with 5,000 sampled points each.
The isosurface generated by our method closely matches the ground truth, whereas CAPUDF~\cite{Zhou2022CAP-UDF}, a representative deep learning-based method, yields a less accurate approximation.
By examining UDF values along a line traversing the domain, we observe that our method preserves sharp transitions when crossing the surface, while CAPUDF produces an overly smoothed result. Furthermore, the gradient field produced by our method remains consistent across the domain, whereas CAPUDF exhibits greater variation, especially in the region far from the zero level set. 

\begin{figure}[!htb]
    \centering
    \includegraphics[width=0.975\linewidth]{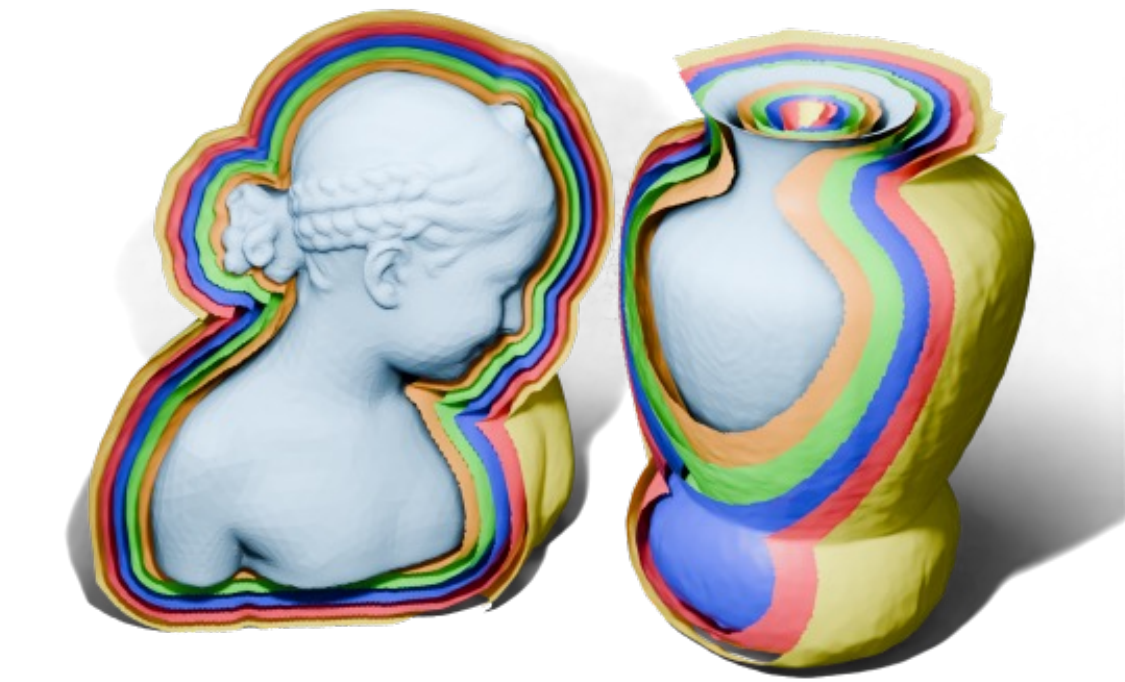}
\caption{Offset surfaces computed at an interval of 0.02 on shapes scaled to the range $[-0.4, 0.4]^3$.}    \label{fig:accuracy_offset}
\end{figure}

\paragraph{Offsets} To further validate the stability of our UDFs, we compute offset surfaces. As shown in Figure~\ref{fig:accuracy_offset}, for both a watertight model and an open surface, the level sets extracted at 0.02 intervals from 0 to 0.1 faithfully preserve the underlying geometry, demonstrating that our method produces well-behaved UDFs. 

\revise{\paragraph{Sensitivity to Normal Accuracy}
During bi-directional normal diffusion followed by the integration for UDF recovery, inaccuracies in the input normals may propagate and lead to erroneous UDF values. In Figure~\ref{fig:normal-accuracy}, we use normals estimated by different methods as input for the diffusion step and subsequent UDF computation, followed by zero-level set extraction using DCUDF. The results show that when point normals are unreliable, the resulting UDF becomes less smooth and less accurate near the zero-level set, which in turn introduces artifacts in the reconstructed surfaces.}

\subsection{Surface Reconstruction}
\label{subsec:reconstructionresults}
We evaluate reconstruction performance on both watertight and non-watertight surfaces, with quantitative results summarized in Table \ref{tab:non-watertight}. \rednote{To evaluate robustness across scales, the dataset features a diverse range of input point densities, spanning from 1K to 100K points.} For watertight models, we compare VAD with SDF, GWN, and Poisson-based methods. Since these methods cannot handle non-watertight surfaces, we instead compare with UDF-based methods in the non-watertight cases.

\begin{figure}[!htbp]
    \centering
\includegraphics[width=\linewidth]{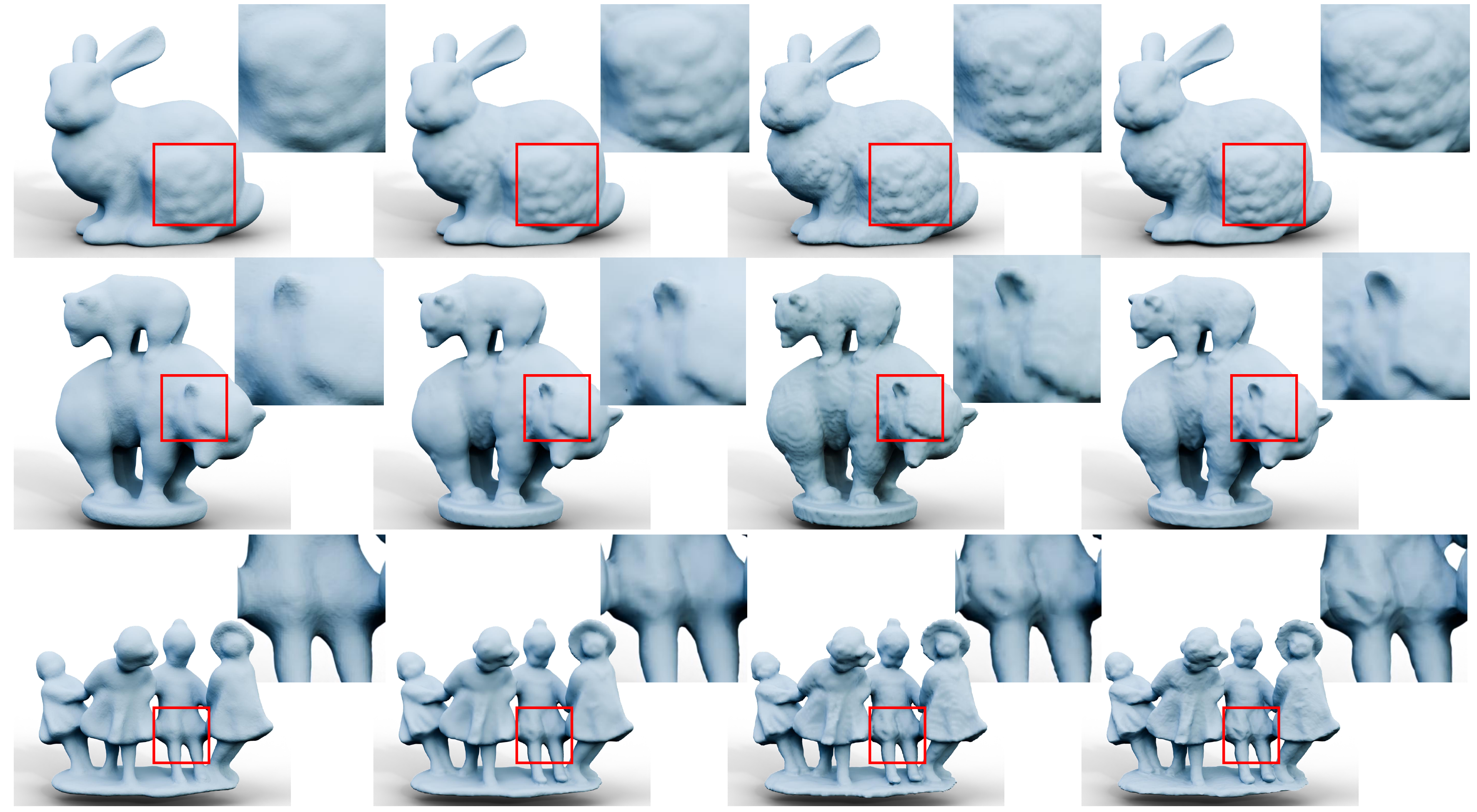}


 \makebox[0.24\linewidth]{DWG}
    \makebox[0.24\linewidth]{SNO+sPSR}
    \makebox[0.24\linewidth]{WNNC}
    \makebox[0.24\linewidth]{Ours}
\caption{\rednote{Qualitative comparison on watertight surfaces with state-of-the-art methods: SNO~\cite{Huang2024Stochastic}, WNNC~\cite{Lin2024fast}, and DWG~\cite{liu2025dwg}. SNO, being a pure point-orientation method, requires an additional reconstruction algorithm such as sPSR to generate the final surface, whereas both DWG and WNNC support direct iso-surface extraction from their GWN fields. VAD achieves comparable or slightly better results, particularly in regions with fine geometric details, where WNNC and DWG tend to over-smooth the surface geometry.}}
    \label{fig:watertightcomparison}
\end{figure}

\begin{figure}[!htbp]
    \centering
    \includegraphics[width=1.0\linewidth]{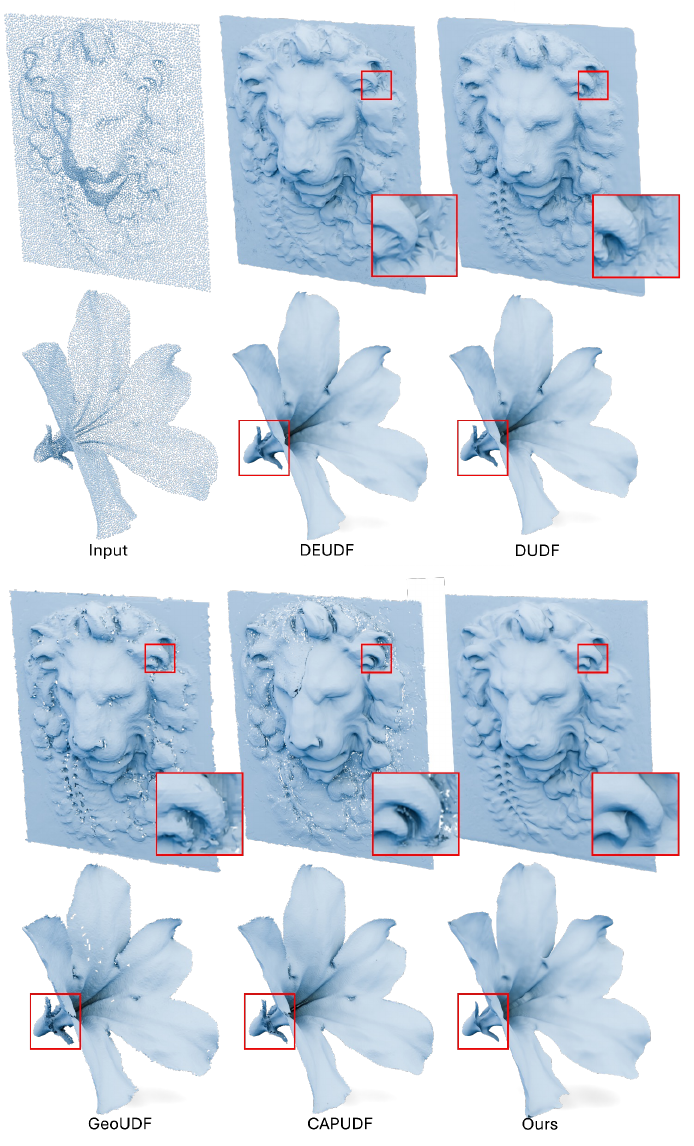}
\caption{Our method reconstructs the relief model with fine geometric detail while preserving smoothness and overall completeness. For the floral model, which features open boundaries and thin structures, our method also produces high-quality reconstruction.}    
\label{fig:Openbound_hard}
\end{figure}
\paragraph{Watertight Surfaces} Existing GWN-based methods (e.g., GCNO~\cite{xu2023globally}, WNNC~\cite{Lin2024fast}, and DWG~\cite{liu2025dwg}), SDF-based methods (e.g., \citep{hoppe1992surface}), and Poisson-based methods (e.g., iPSR~\cite{hou2022iterative} and SNO~\cite{Huang2024Stochastic}) are primarily designed for watertight surfaces, leveraging the clear interior-exterior distinction to compute the implicit function. Figure~\ref{fig:watertightcomparison} shows reconstructions from 10k uniformly sampled points for each test model. DWG tends to over-smooth geometric details, while WNNC introduces noticeable artifacts. In contrast, our method preserves fine structures and produces visually cleaner and more accurate surfaces.

\begin{figure}[htbp]
    \centering    \includegraphics[width=0.95\linewidth]{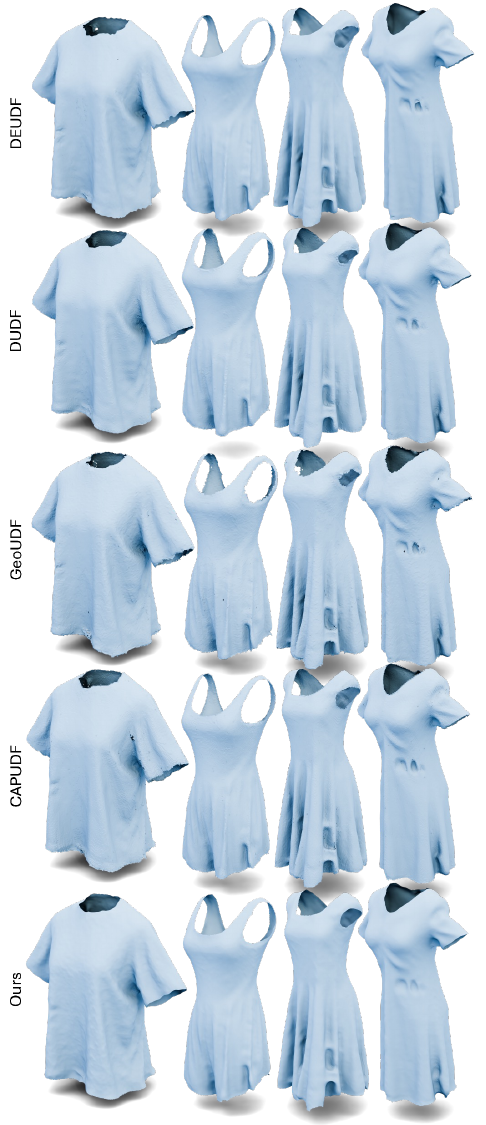}
\caption{Reconstruction results on cloth models from the DeepFashion3D dataset. Compared with existing deep learning-based methods, our approach generates meshes with smoother surfaces, fewer artifacts, and more accurate recovery of fine geometric details such as wrinkles.}
    \label{fig:Openbound_cloth}
\end{figure}

\paragraph{Open Surfaces}
For open surfaces such as the Relief model, the Floral model (Figure~\ref{fig:Openbound_hard}), and the cloth models (Figure~\ref{fig:Openbound_cloth}), our method demonstrates superior reconstruction quality. The existing deep learning methods  CAPUDF~\cite{Zhou2022CAP-UDF}, GeoUDF~\cite{DBLP:conf/iccv/RenHCHW23}, DUDF~\cite{Fainstein2024DUDF}, and DEUDF~\cite{Xu2024DEUDF} often produce holes or artifacts in open surfaces. This is primarily due to the difficulty of neural networks in fitting UDFs near the zero level set, which is theoretically non-differentiable. In contrast, our method provides a more stable and controllable reconstruction by better fitting both the UDF values and their gradients around the zero level set, leading to more reliable surface extraction and reconstruction results.

\begin{figure*}[!htbp]
    \centering    \includegraphics[width=0.95\textwidth]{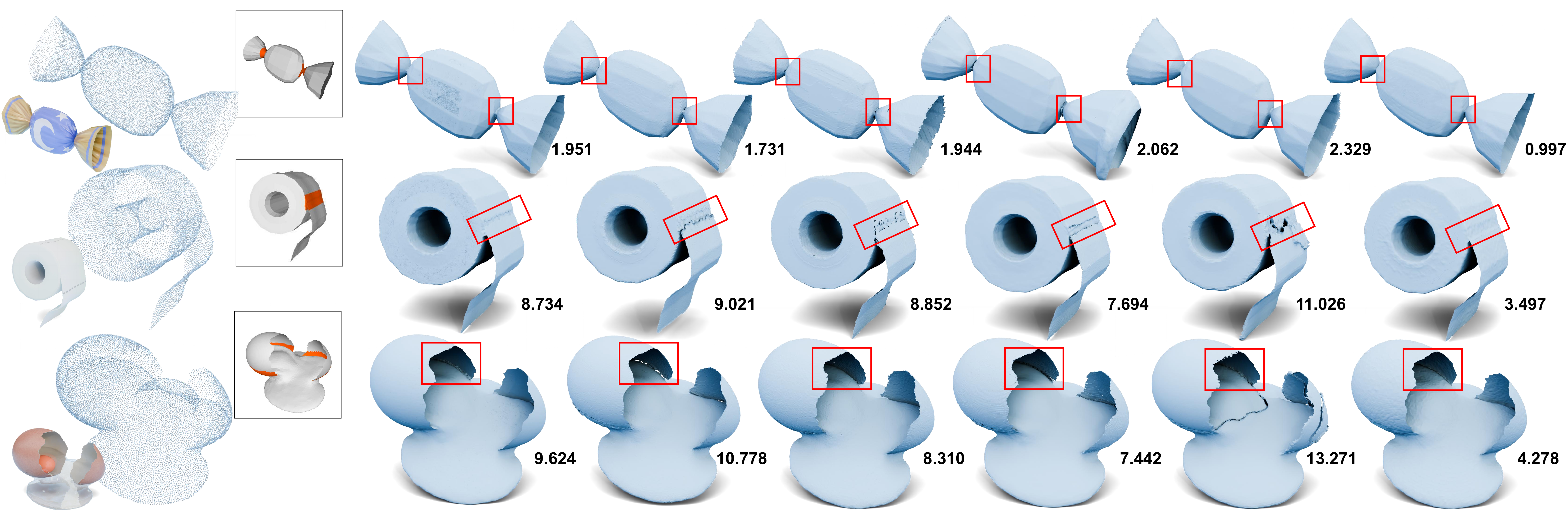}
        \makebox[0.18\linewidth]{Raw Input}
    \makebox[0.12\linewidth]{DEUDF}
    \makebox[0.12\linewidth]{DUDF}
    \makebox[0.12\linewidth]{GeoUDF}
    \makebox[0.12\linewidth]{CAPUDF}
    \makebox[0.12\linewidth]{DACPO}
\makebox[0.12\linewidth]{Ours}\\
\caption{\revise{Results on real-world models containing non-manifold structures. Regions highlighted by red rectangles indicate the locations of non-manifold structures. }\rednote{The regions highlighted in red in the GT are used to compute the local Chamfer Distance.}}
    \label{fig:nonmanifold}
\end{figure*}

\paragraph{Non-manifold Surfaces}
GWN-based methods cannot handle non-manifold structures because computing the generalized winding number field requires integration over the boundary surface. At non-manifold configurations, such as edges shared by more than two faces or self-intersections, the notion of a well-defined boundary orientation breaks down, leading to inconsistent winding numbers and invalid indicator fields. In contrast, our method remains robust in these cases by diffusing bi-directional normals instead of performing boundary integration.
As shown in Figure~\ref{fig:nonmanifold}, our method successfully captures the underlying structure and produces coherent UDF fields, demonstrating its ability to handle real-world models, particularly man-made objects that frequently contain non-manifold structures. 
\begin{figure}
    \centering
    \includegraphics[width=\linewidth]{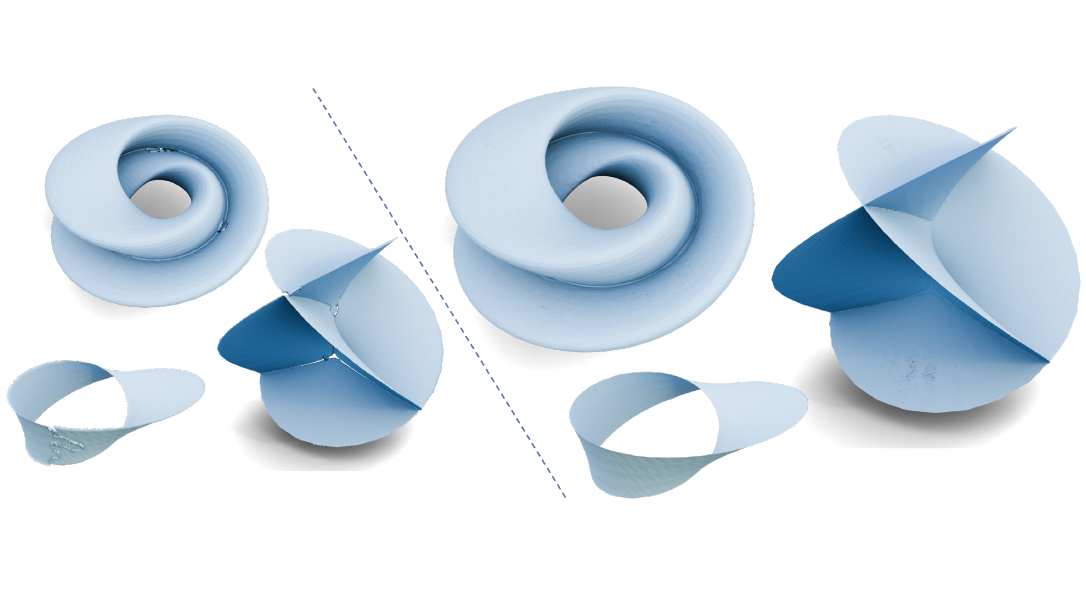}
      \makebox[0.45\linewidth]{DACPO}
    \makebox[0.5\linewidth]{Ours}
    \caption{\rednote{Qualitative comparison on non-orientable models. Since a globally consistent normal orientation cannot be defined on non-orientable surfaces, methods relying on global normal orientation (e.g., DACPO) often suffer from topological breakdowns. In contrast, our bi-directional normal formulation avoids this requirement and successfully recovers the target geometry.}}
    \label{fig:non-oriented}
\end{figure}

\paragraph{Non-orientable Geometries}\rednote{Our method can naturally handle non-orientable surfaces, such as the  M\"obius band, where a globally consistent normal orientation is mathematically undefined.}
\begin{wrapfigure}{r}{0.06\textwidth} 
\centering \vspace{-1.5em} \hspace{0.5em} 
\makebox[40pt][r]{\includegraphics[width=0.10\textwidth]{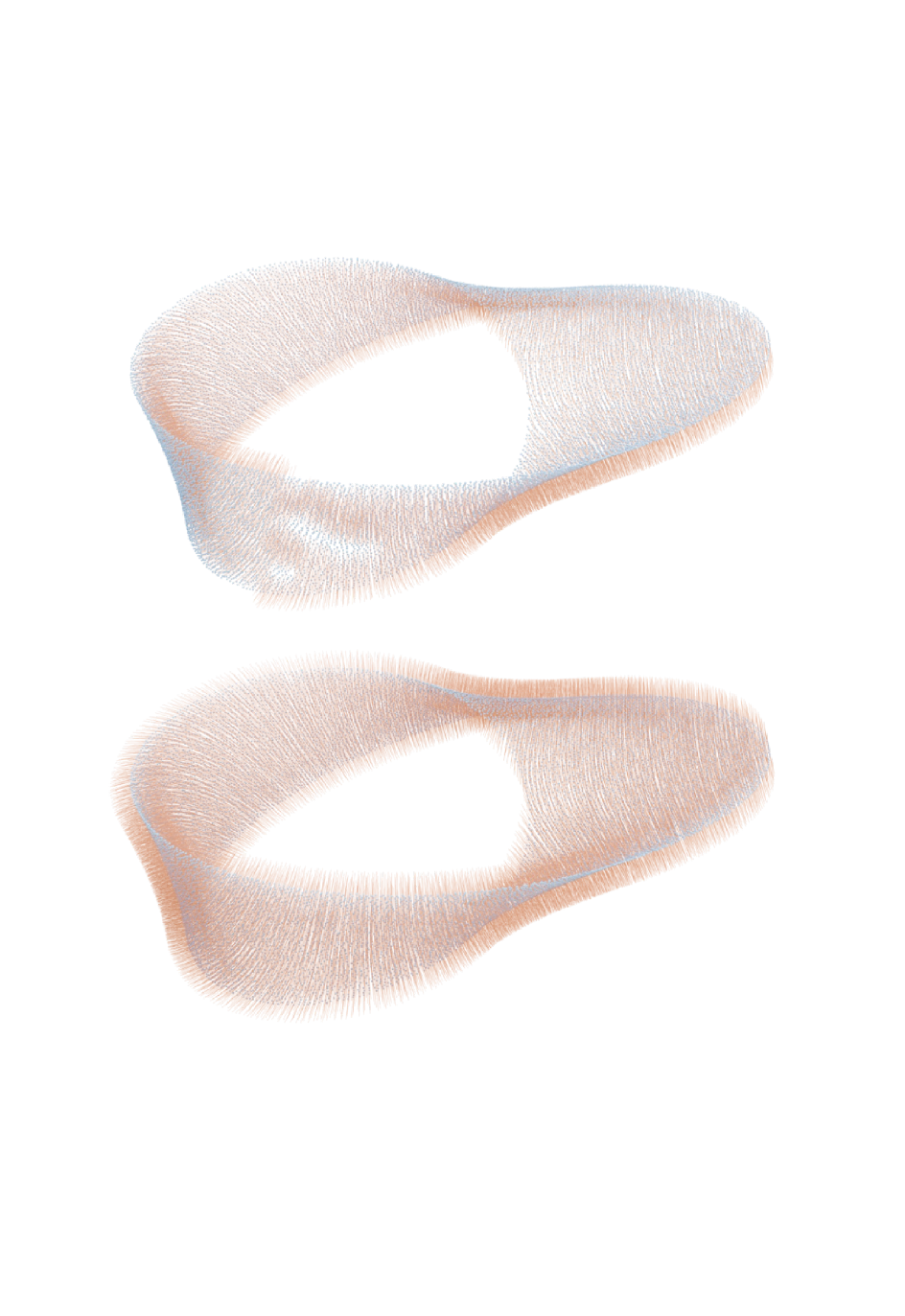}} 
\end{wrapfigure}
\rednote{As illustrated in the inset, methods that rely on globally consistent normal orientations (e.g., DACPO, shown in the upper part) fail to produce a valid orientation field on such geometries. In contrast, by representing normals as bi-directional vectors (shown in the lower part), our formulation removes the requirement of global orientation consistency and remains applicable to non-orientable surfaces. Additional reconstruction results on non-orientable surfaces are provided in Figure~\ref{fig:non-oriented}.}
\begin{figure}[htb]
    \centering
    \includegraphics[width=\linewidth]{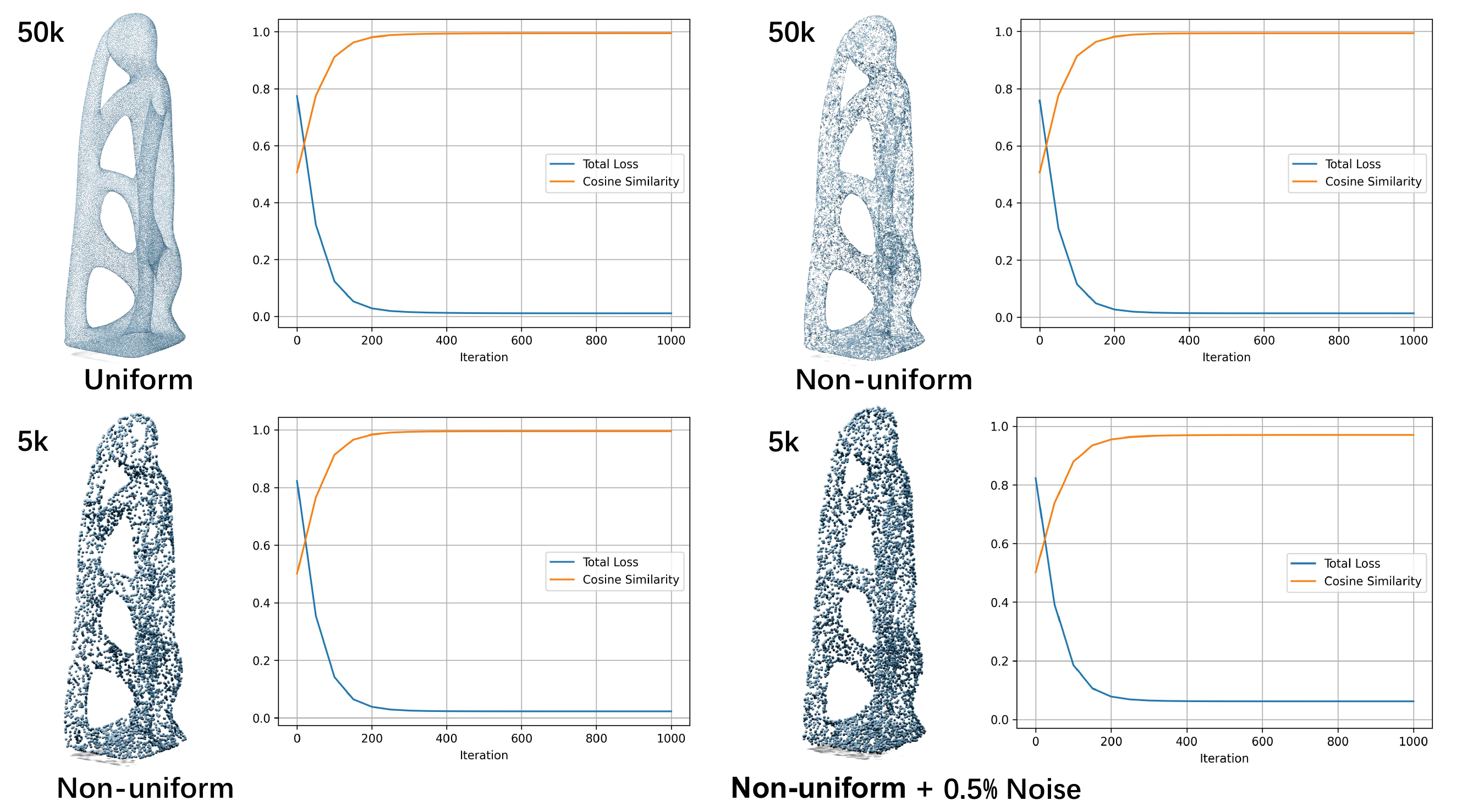}
    \caption{Convergence plots under different conditions. For dense, uniform, non-uniform, sparse, and noisy inputs, the normals consistently converge during optimization.}
    \label{fig:energy-analysis}
\end{figure}

\begin{table*}[!htbp]
\centering
\resizebox{0.98\textwidth}{!}{
\begin{tabular}{l|cc|cc|cc|cc|cc|cc|cc}
\hline
 & \multicolumn{2}{c|}{Fertility (1k)} & \multicolumn{2}{c|}{Knot (1k)} & \multicolumn{2}{c|}{Bunny (10k)} & \multicolumn{2}{c|}{4-Children (10k)} & \multicolumn{2}{c|}{Bears (10k)} & \multicolumn{2}{c|}{Buddha (100k)} & \multicolumn{2}{c}{Armadillo (100k)} \\ \hline
                 
Model         & CD        & HD            & CD      & HD         & CD       & HD        & CD        & HD              & CD       & HD        & CD       & HD           & CD        & HD            \\ \hline    
Hoppe+sPSR & 1.707     & 31.472         & \underline{0.629 }  & 30.691     & 0.843    & 8.382     & 1.144     & 18.193       & 0.949    & 9.107      & \underline{0.215}    &\underline{ 8.461 }     &\underline{ 0.263}     & 6.991        \\
GCNO+sPSR             & \underline{1.524}     & 21.772      & \textbf{0.468 }  & \underline{7.965}    & 0.447    & \textbf{7.260 }   & 0.593     & 17.311      & 0.575    & 9.095    & -        & -                & -         & -            \\
SNO+sPSR               & 1.956     & 25.181        & 1.093   & 7.983    &\underline{ 0.364 }   & 7.727   & 0.553     & 18.917      & 0.596    & \underline{8.543 }    & 0.426    & 11.286      & 0.324     & 9.071     \\
WNNC              & 2.589     & \underline{19.863 }        & 2.589   &20.751     & 0.406    & 8.030   & \underline{0.548}    &\underline{ 15.734  }       & \underline{0.547}    & 8.075      & 0.246    & 10.912      & 0.292     & \underline{6.714}      \\
DWG               & 2.449     & 49.701       & 1.495   & 36.447     & 0.402    & 22.582       & 0.649     & 21.605        & 0.600    & 22.843     & 0.291    & 16.771        & 0.296     & 17.464      \\ \hline
Ours              & \textbf{0.954 }    & \textbf{8.642}        & 0.812   & \textbf{7.427}     & \textbf{0.266 }   & \underline{7.523}   & \textbf{0.419 }    & \textbf{9.615   }     & \textbf{0.385 }   & \textbf{8.172 }      & \textbf{0.106 }   & \textbf{8.240   }    &\textbf{ 0.209 }    &\textbf{ 6.242 }  \\ \hline

\end{tabular}
}
\vspace{0.1in}  
\resizebox{0.98\textwidth}{!}{
\begin{tabular}{l|cc|cc|cc|cc|cc|cc}
\hline
        & \multicolumn{2}{c|}{Paperplane (1k)} & \multicolumn{2}{c|}{Strawberry (1k)} & \multicolumn{2}{c|}{Lion (10k)} & \multicolumn{2}{c|}{Cloth (10k)} & \multicolumn{2}{c|}{Candy (10k)} & \multicolumn{2}{c}{Toilet-Paper (10k)} \\ \hline
Model        & CD                & HD               & CD                & HD               & CD             & HD             & CD              & HD             & CD              & HD             & CD                 & HD                \\ \hline     
Hoppe et al. & 1.699             & 27.953            & 2.758             & 60.831            & 0.804          & \underline{10.707}          & 1.208           & 15.235          & 1.998           & 21.704          & 1.735              & 22.592             \\
CAPUDF       & 78.448            & 405.323           & 3.695             & 146.782           & 0.395          & 121.948         & 0.344           & \underline{10.815 }         & 0.145           & 80.892          & 0.177              & 19.965             \\
GeoUDF*      & \textbf{0.290 }            & 35.244            &\textbf{ 1.085  }           & 62.792            & \underline{0.354}          & 14.662          & 0.268           & 23.296          & 0.162           & 14.134          & 0.167              & 16.507             \\
DEUDF        & 5.386             & 64.944            & 12.323            & 477.193           & \textbf{0.201 }         & 15.221          &\underline{ 0.172}           & 10.872          & \textbf{0.039}           & \underline{7.201 }          &\underline{ 0.135 }            & \underline{12.929  }           \\
DUDF         & 1.789             & \underline{24.128    }        & 1.962             & \underline{50.656  }          & 0.777          & 19.082          & 0.870           & 11.945          & 0.423           & 10.517          & 1.205              & 14.383             \\
LoSF*        & -                 & -                & -                 & -                & 7.268          & 46.257          & 4.091           & 31.083          & 5.424           & 35.975         & 4.453              & 56.211             \\
DACPO        & 1.778             & 64.773           & 3.412             & 66.476            & 0.525          & 15.721          & 0.658           & 19.475          & 0.155           & 14.363          & 0.521              & 26.477             \\\hline
Ours         & \underline{0.647 }            & \textbf{21.521  }          &\underline{ 1.329    }         & \textbf{36.225  }          & 0.374          & \textbf{8.900 }          & \textbf{0.162 }          & \textbf{8.794 }          &\underline{0.086}           & \textbf{6.382 }          & \textbf{0.131 }             & \textbf{10.221 }            \\ \hline
\end{tabular}
}
\caption{Quantitative results on watertight models (top) and non-watertight models (bottom), evaluated using Chamfer Distance (CD, $\times 10^{3}$) and Hausdorff Distance (HD, $\times 10^{3}$). LoSF requires relatively dense sampling and therefore cannot handle sparse inputs with only 1k points.  All compared methods utilize an identical regular grid resolution of $128^3$ during the mesh extraction. The best results are highlighted in bold, while the second-best results are underlined. *~denotes methods that utilize pre-training.}
\label{tab:non-watertight}
\end{table*}

\revise{\subsection{Convergence \& Efficiency}
For most models, our optimization procedure converges rapidly: the energy decreases sharply toward zero and stabilizes within 500 iterations, requiring only a few seconds of runtime.
As the energy decreases, the bi-directional normals become increasingly aligned with the ground truth. Figure~\ref{fig:energy-analysis} illustrates this behavior using the Wood-Thinker model as an example.
Moreover, the convergence exhibits little dependence on sampling density or distribution (uniform vs. non-uniform) and is only marginally affected by noise.

Detailed runtime profiling of VAD is reported in Table~\ref{tab:timing}. The computation of bi-directional normals empirically scales linearly with input size. The diffusion of bi-directional normals to obtain $\mathbf{Y}_f$ field and the subsequent solution of the Poisson equation are more computationally expensive, depending on both point density and grid resolution. }

\begin{table}[!htbp]
\centering
\setlength\tabcolsep{2pt}
\resizebox{0.48\textwidth}{!}{
\begin{tabular}{l|cccc}
\hline
\multirow{2}{*}{\begin{tabular}[c]{@{}l@{}}\textbf{Bi-directional normal}\\ \textbf{optimization (s)}\end{tabular}} & \multicolumn{4}{c}{Input}                          \\ \cline{2-5} 
& 1k          & 5k         & 15k       & 100k        \\ \hline
Building Voronoi (s)                                                                                 & 0.014       & 0.026      & 0.17      & 2.40        \\
Sampling bisectors (s)                                                                              & 0.062 (10k) & 0.11 (20k) & 0.62 (1M) & 10.99 (10M) \\
Adam solver (s)                                                                                   & 1.83        & 2.07       & 9.06      & 42.81       \\ \hline
Total time (s)                                                                                    & 1.906       & 2.21       & 9.85      & 56.20       \\ \hline
\textbf{Diffusion \& integration (s)}                                                             &             &            &           &             \\ \hline
$32^3$                                                                                            & 0.46        & 1.95       & 3.64      & 32.38       \\
$64^3$                                                                                            & 5.32        & 11.56      & 28.85     & 226.92      \\
$128^3$                                                                                           & 40.10       & 119.26     & 234.70    & 1742.93     \\
$256^3$                                                                                           & 351.51      & 799.37     & 1731.03   & -           \\ \hline
\end{tabular}
}

\vspace{1em}


    \centering
    \includegraphics[width=0.47\textwidth]{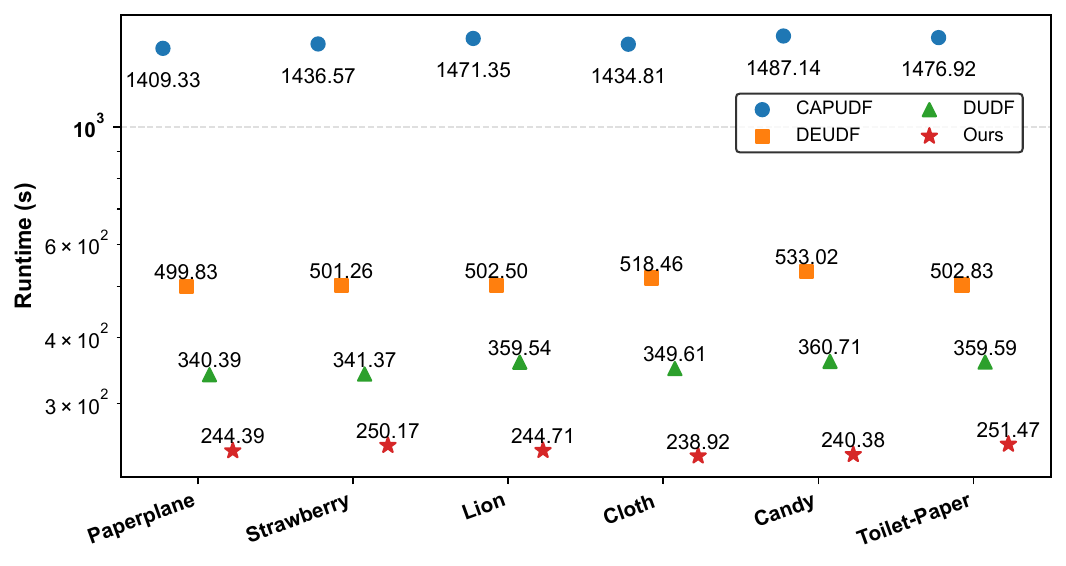}

\caption{\revise{Runtime performance. We provide detailed profiling of bi-directional normal optimization, including building Voronoi diagrams, uniformly sampling on Voronoi bisectors (with the total number of samples reported), and minimizing the energy using the Adam solver. For the UDF computation step, which consists of bi-directional normal diffusion and integration, we evaluate different grid resolutions from $32^3$ to $256^3$. We also compare the overall runtime with other UDF reconstruction methods on point clouds with 10k points, using a grid resolution of $128^3$. }}
\label{tab:timing}
\end{table}


\begin{figure}[!htbp]
    \centering
    \includegraphics[width=0.95\linewidth]{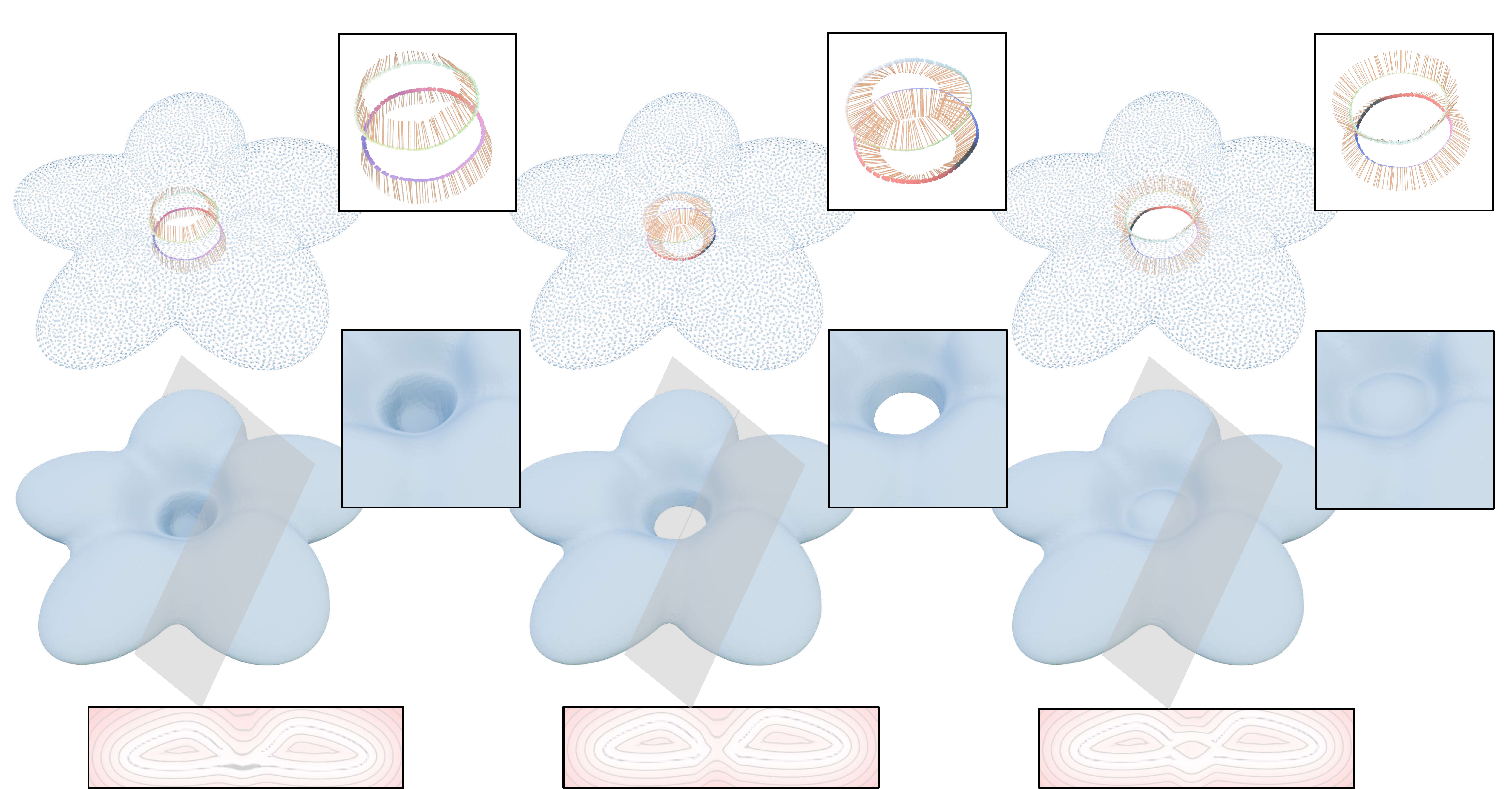}
\caption{\revise{Controllability. For the given petal-like model, we allow users to specify different boundary orientations (highlighted in close-up views). VAD proceeds with these user-specified orientations as hard constraints and then optimize bi-directional normals for other points. Each boundary condition yields a meaningful reconstruction result.
}}
    \label{fig:control}
\end{figure}

\begin{figure*}[!htbp]
    \centering
    \includegraphics[width=0.94\textwidth]{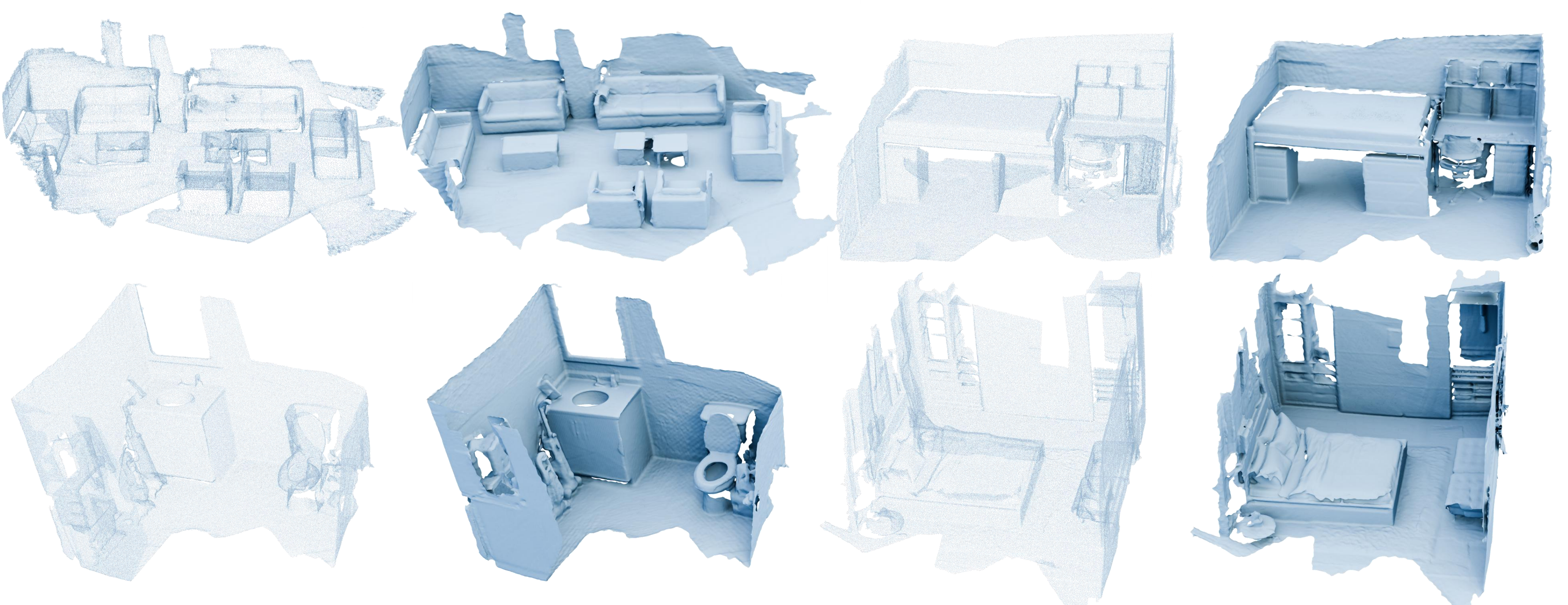}
    \caption{Results on indoor scenes with around 150k input points, showing that the capability of handling scene-level data. \rednote{The grid resolution is set to $256^3$}.}
    \label{fig:scene}
\end{figure*}

\begin{figure*}
    \centering    \includegraphics[width=0.95\linewidth]{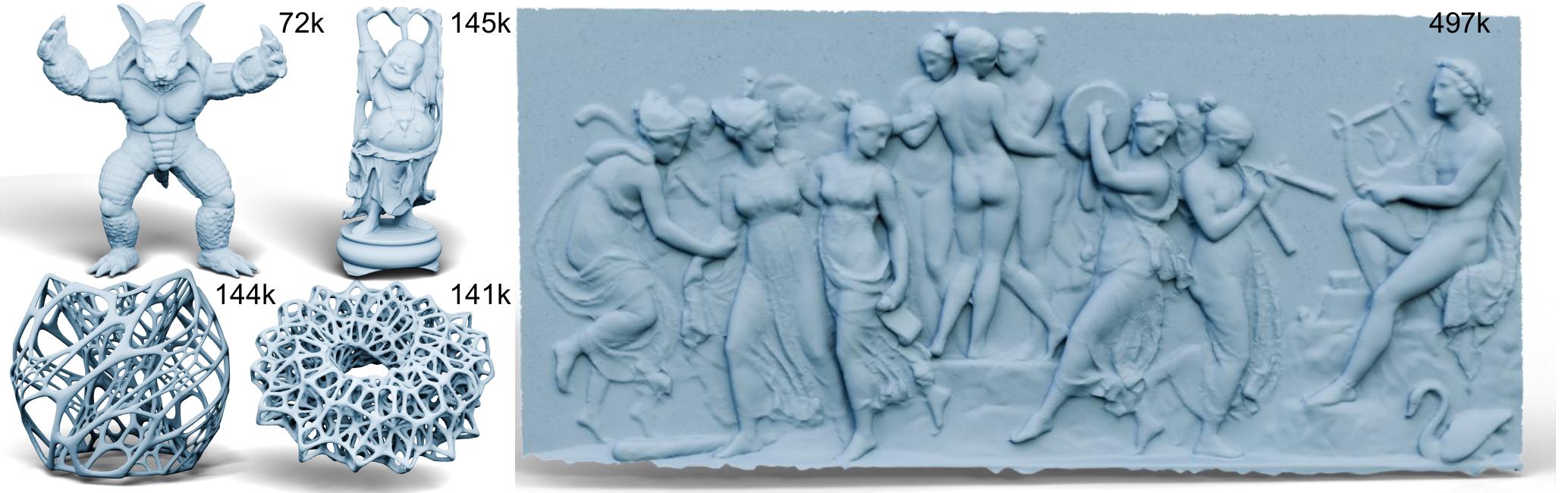}
    \caption{Results on models with complex topology and fine details. Our method effectively handles intricate geometry, complex topological structures, and non-manifold features. \rednote{The grid resolution is set to $256^3$}.}

    \label{fig:complex}
\end{figure*}
\begin{figure*}
    \centering    \includegraphics[width=1.0\textwidth]{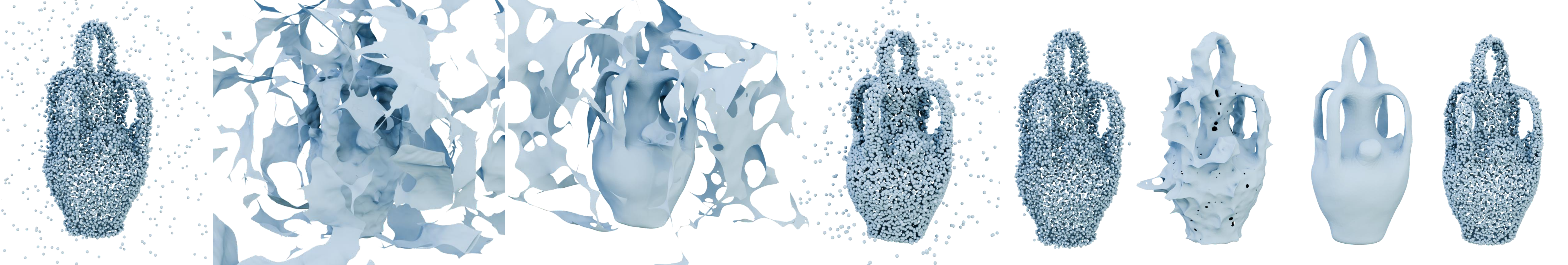}\\
    \makebox[0.7in][c]{(a) Input}%
\makebox[1.3in][c]{\parbox[t]{1.3in}{\centering(b) VAD w/o\\ denoising}}%
\makebox[1.4in][c]{\parbox[t]{1.4in}{\centering(c) VAD w/\\ denoising}}
\makebox[0.8in][c]{\parbox[t]{0.8in}{\centering(d) Optimized\\ points}}
\makebox[0.6in][c]{\parbox[t]{0.6in}{\centering(e) Outlier\\ removal}}
\makebox[0.6in][c]{\parbox[t]{0.6in}{\centering(f) VAD w/o\\ denoising}}
\makebox[0.7in][c]{\parbox[t]{0.7in}{\centering(g) VAD w/\\ denoising}}
\makebox[0.7in][c]{\parbox[t]{0.7in}{\centering(h) Optimized\\ points}}\caption{\revise{Limitations. Our method struggles with noisy inputs containing outliers. (a-d) Directly applying VAD with denoising still results in fragmented surfaces and disconnected components, since point position optimization cannot correct outliers far from the target surface. (e-h) Preprocessing to remove outliers effectively improves VAD’s performance on such inputs.} }

    \label{fig:outlier}
\end{figure*}

\paragraph{Controllability}  
As discussed in Section~\ref{discussions}, surface reconstruction under sparse settings is inherently ill-posed without normal information, as multiple feasible solutions may exist. Our method addresses this by allowing user control within the reconstruction pipeline. 
Specifically, we provide an interactive interface that enables users to specify orientations for key regions, such as boundaries, which are then enforced as hard constraints during bi-directional normal optimization. Two examples are shown in Figure~\ref{fig:control}.

\subsection{Robustness}
\label{subsec:robustness}
\begin{figure}
    \centering
    \includegraphics[width=0.95\linewidth]{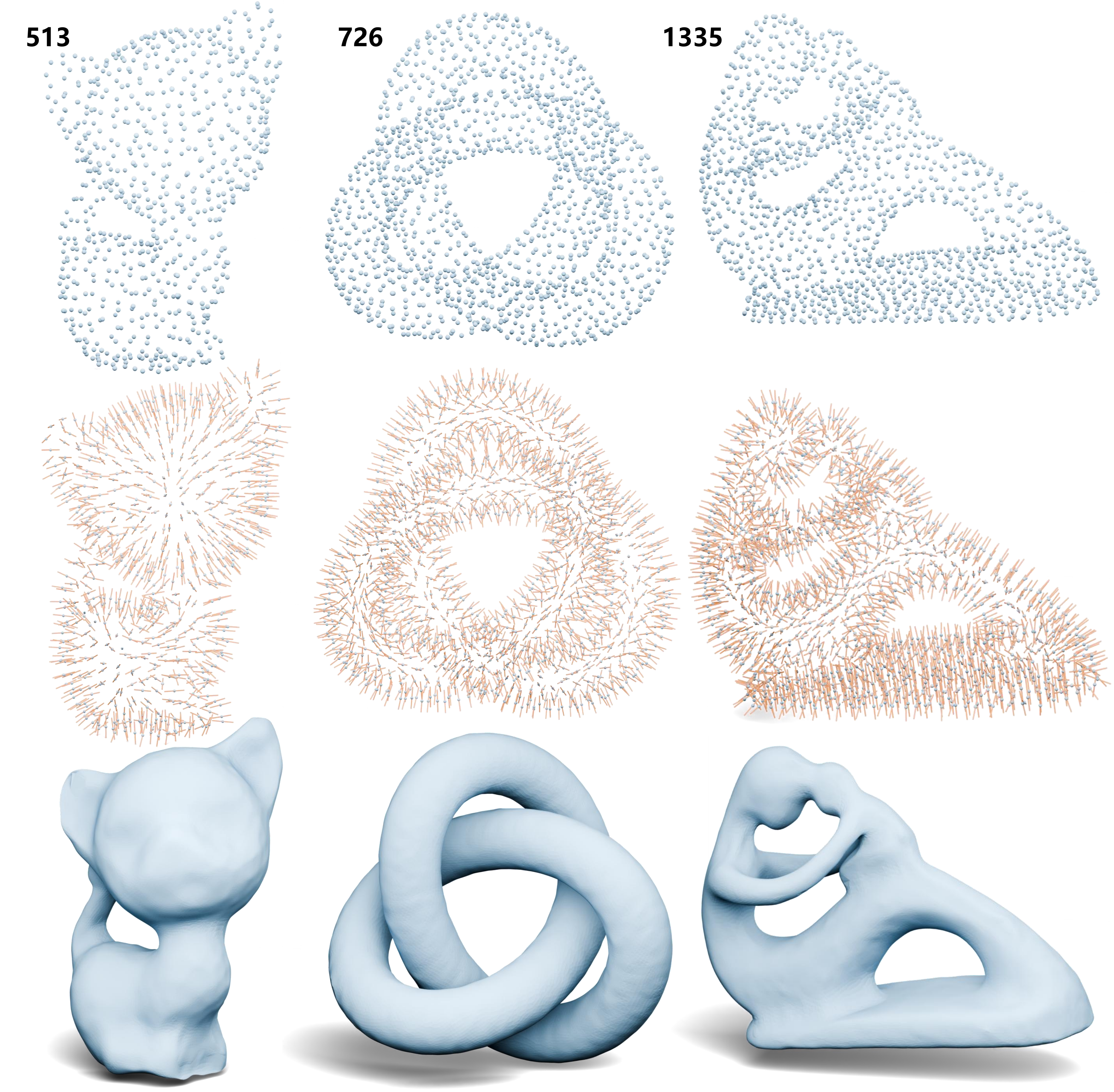}
\caption{Surface reconstruction from sparse inputs. 
We show the optimized bi-directional normals and the corresponding reconstructed surfaces.}

    \label{fig:sparse3}
\end{figure}

\begin{figure}
    \centering
    \includegraphics[width=\linewidth]{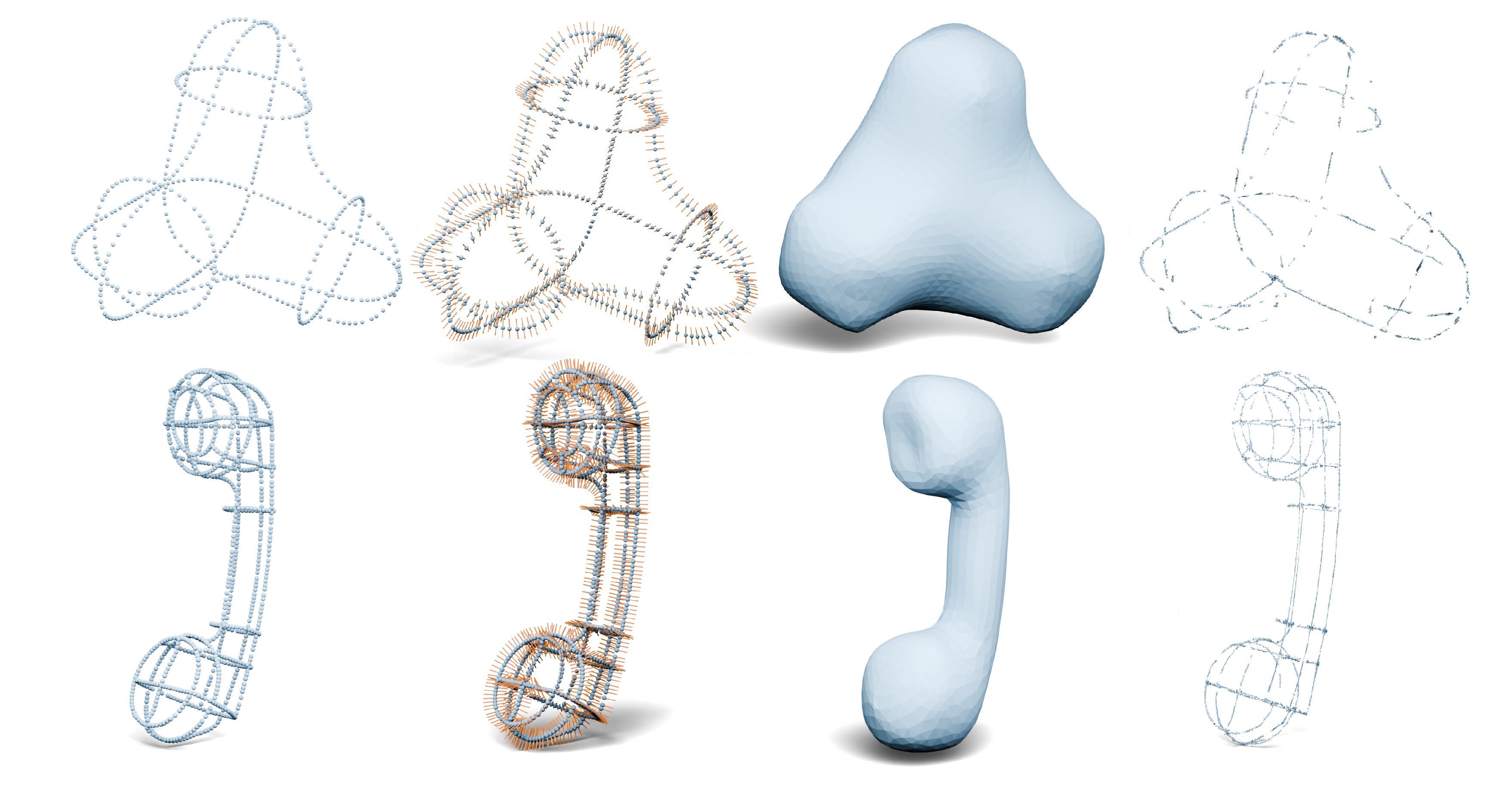}
    \makebox[0.1\textwidth]{(a)}%
        \makebox[0.12\textwidth]{(b)}%
        \makebox[0.12\textwidth]{(c)}%
        \makebox[0.12\textwidth]{(d)}
    \caption{Normals and reconstructed surfaces from wireframe inputs. (a) Input point clouds. (b) Bi-directional normals estimated by our method. (c) Surface reconstructed with our method. (d) Surface reconstructed with CAPUDF.}
    \label{fig:wireframe}
\end{figure}

\begin{figure*}
    \centering
    \includegraphics[width=0.9\textwidth]{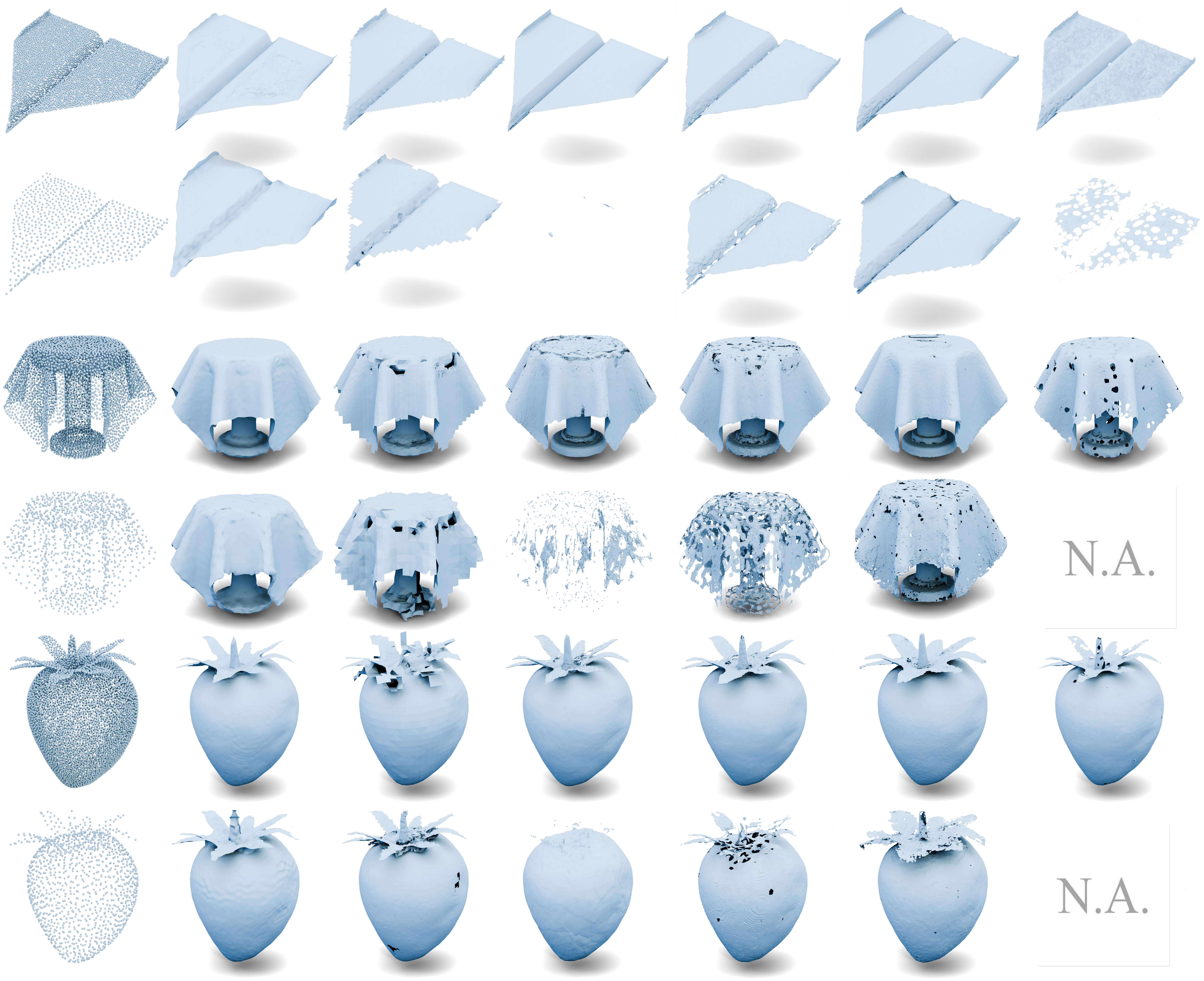}
    \includegraphics[width=0.9\textwidth]{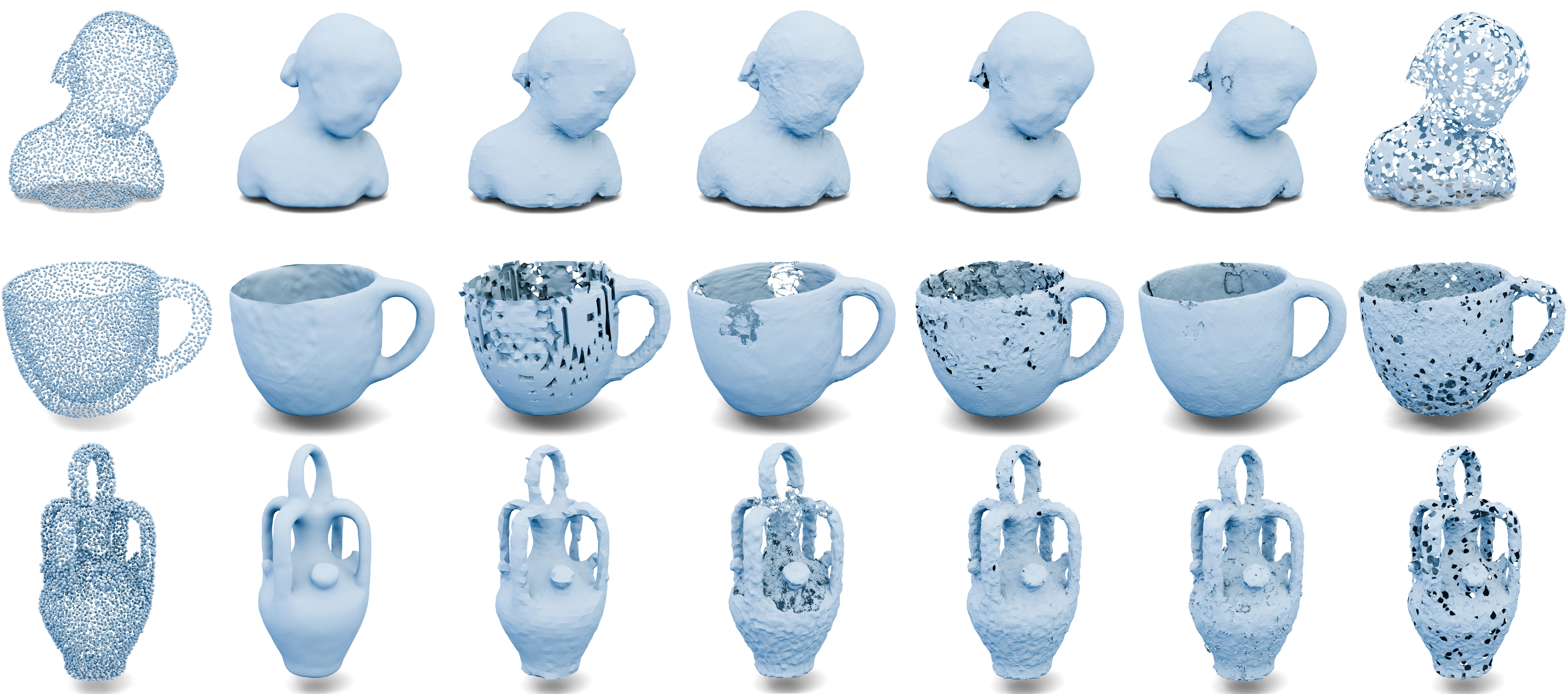}
    \makebox[0.15\textwidth]{Input}%
        \makebox[0.134\textwidth]{Ours}%
        \makebox[0.134\textwidth]{Hoppe}
        \makebox[0.134\textwidth]{CAPUDF}%
        \makebox[0.134\textwidth]{DUDF}
        \makebox[0.134\textwidth]{GeoUDF}%
        \makebox[0.13\textwidth]{DEUDF}
\caption{\revise{Robustness tests. The top six rows show reconstructions under varying sampling densities, including open surfaces and non-manifold structures. The bottom three rows show results on  noisy inputs (3K points) with 0.5\% Gaussian noise. Compared with baseline methods, our approach demonstrates stronger robustness, yielding smoother and more accurate surfaces.}}
    \label{fig:sparse-open}
\end{figure*}

\paragraph{Sparse Inputs}
Our method offers controllability and remains effective even under sparse inputs. Figure~\ref{fig:sparse3} shows reconstruction results on watertight models with 500–1500 points, while Figure~\ref{fig:sparse-open} presents results on open surfaces with non-manifold structures. Our method also performs well on wireframe data, where points are extremely unevenly distributed (Figure~\ref{fig:wireframe}). In contrast, deep learning-based methods such as CAPUDF tend to overfit the input points, reproducing the wireframe structure rather than generating coherent surfaces that span the frames.


\paragraph{Complex Geometries and Topologies} Our method can handle models with fine details and complex topologies, as shown in Figure~\ref{fig:complex}. Beyond object-level reconstruction, VAD can also be applied to indoor scenes, as shown in Figure~\ref{fig:scene}. These results highlight the versatility of our method in producing consistent and reliable reconstructions across diverse and complex geometries.

\begin{figure}[htb]
    \centering
    \includegraphics[width=\linewidth]{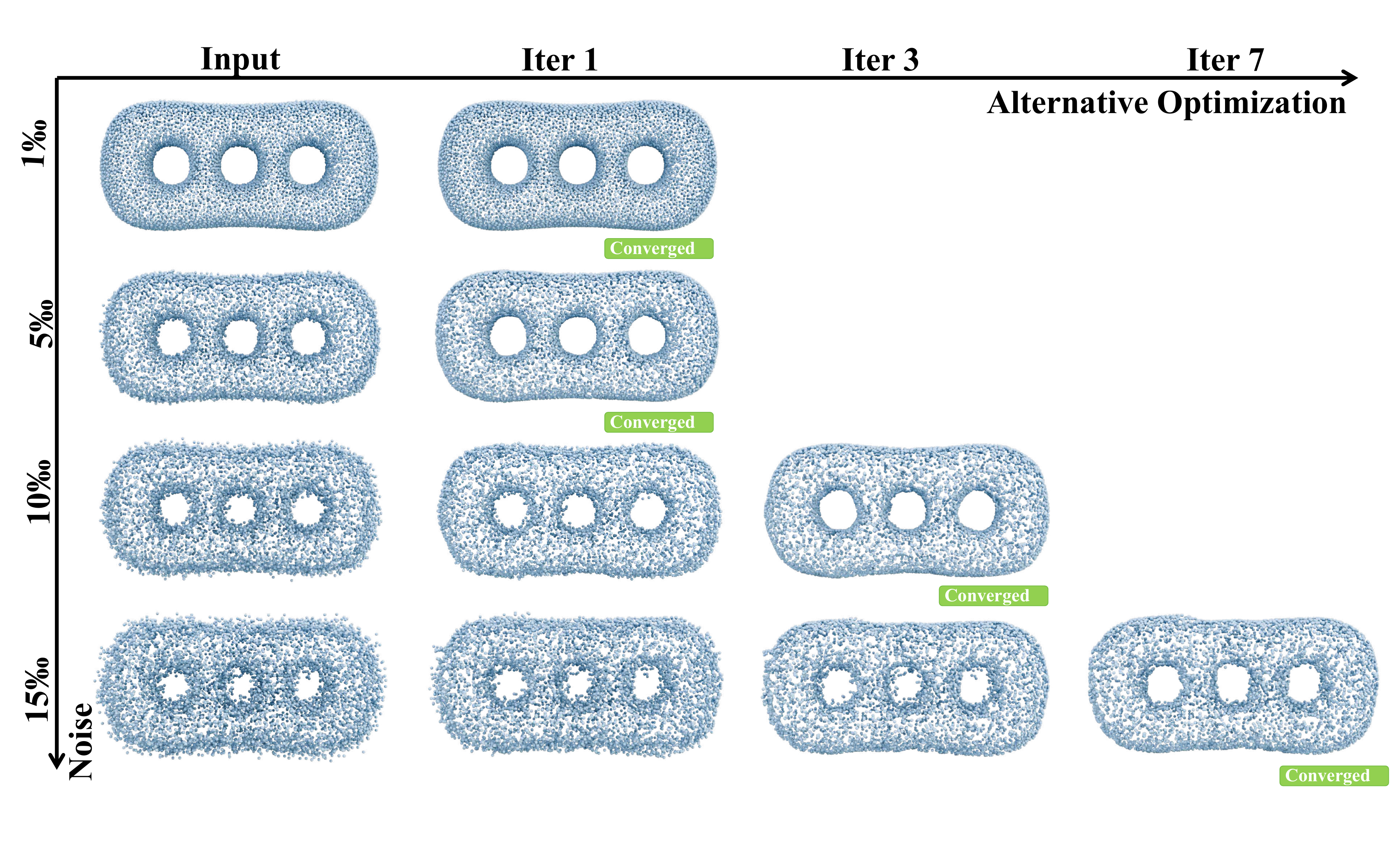}
    \caption{\rednote{Convergence behavior under different noise levels. Rows correspond to inputs corrupted with different amounts of noise, while columns show the point cloud positions after successive iterations of alternative optimization. Each iteration (Iter) consists of one normal optimization step and one position optimization step. As the noise level increases, more iterations are required before the point positions converge to a stable geometry.}}
    \label{fig:noise_analysis}
\end{figure}

\rednote{\paragraph{Noisy Points}}
\rednote{Our method is robust to low and moderate levels of noise, benefiting from the denoising effect of the proposed alternating optimization. As the noise level increases, additional optimization iterations may be required to reach convergence. Figure~\ref{fig:noise_analysis} illustrates this behavior: inputs with 0.1\% and 0.5\% Gaussian noise converge after a single position optimization, while noisier inputs (1\% and 1.5\%) require several additional alternating updates. Nevertheless, for most practical noise levels, a single position optimization step is already sufficient to produce a stable and accurate reconstruction.}

Figure~\ref{fig:sparse-open} presents reconstruction results on sparse and noisy inputs (3k points with 0.5\% Gaussian noise). VAD consistently produces high-quality surfaces, whereas existing learning-based approaches often exhibit broken regions, holes, and non-smooth artifacts.

\begin{figure}[!htbp]
    \centering        \includegraphics[width=0.5\textwidth]{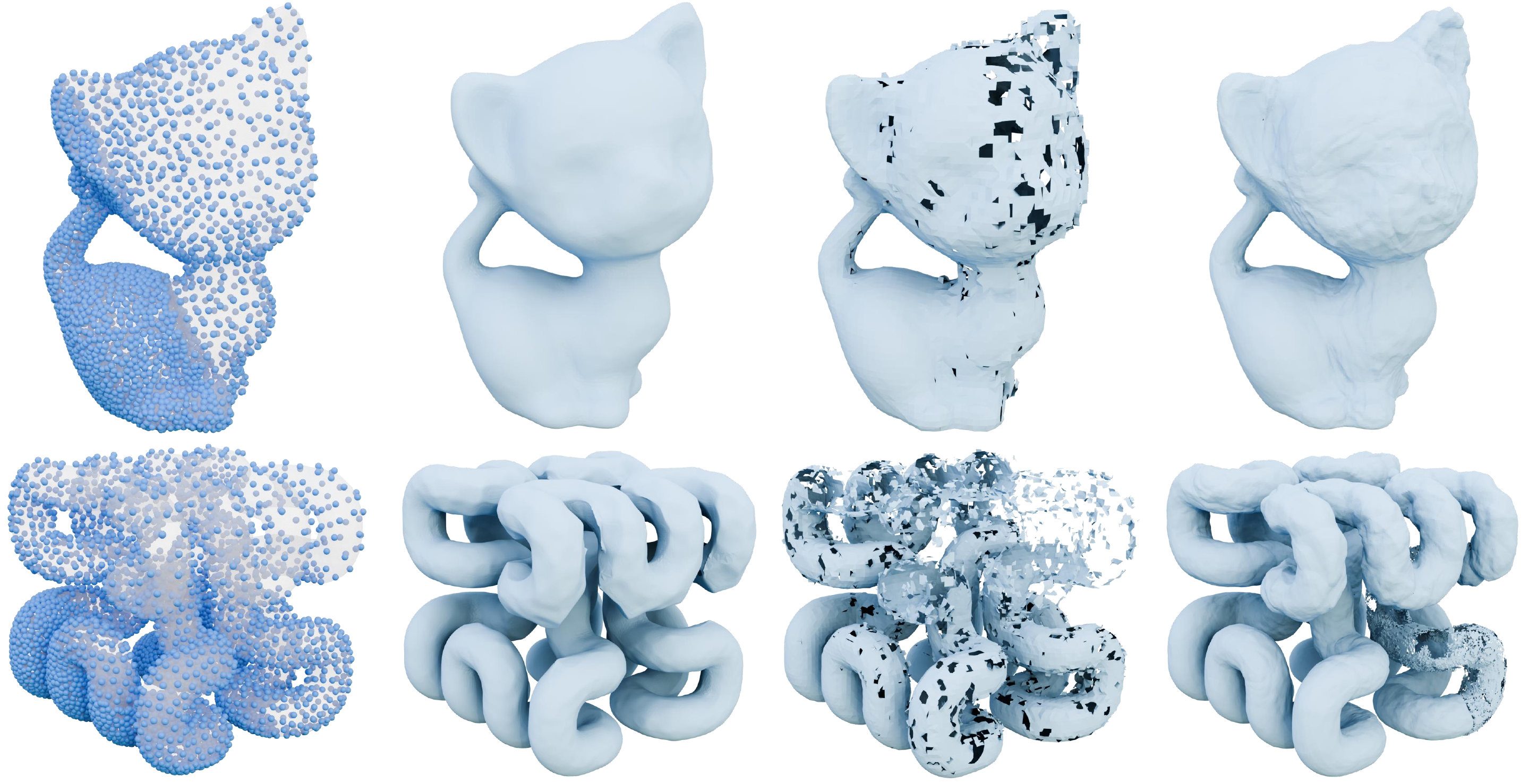}
        \makebox[0.12\textwidth]{Input}%
        \makebox[0.12\textwidth]{Ours}%
        \makebox[0.12\textwidth]{GeoUDF}%
        \makebox[0.12\textwidth]{CAPUDF}
    \caption{\revise{Surface reconstruction under varying point cloud densities. 
Each model is designed with a spatially varying density, ranging from dense to sparse regions. 
Our method produces stable reconstructions, demonstrating robustness to changes in sampling rate.}}

    \label{fig:varydensity}
\end{figure}

\paragraph{Varying Sampling Rates}  
To evaluate robustness under different sampling rates, we construct test models with spatially varying point densities, ranging from low (200 points), medium (1k points), to relatively high (5k points). 
\revise{Experimental results show that our method remains stable across all densities and is more robust than deep learning-based methods such as CAPUDF and GeoUDF in handling unevenly distributed points (see Figure~\ref{fig:varydensity}).}

\begin{figure}
    \centering    \includegraphics[width=0.48\textwidth]{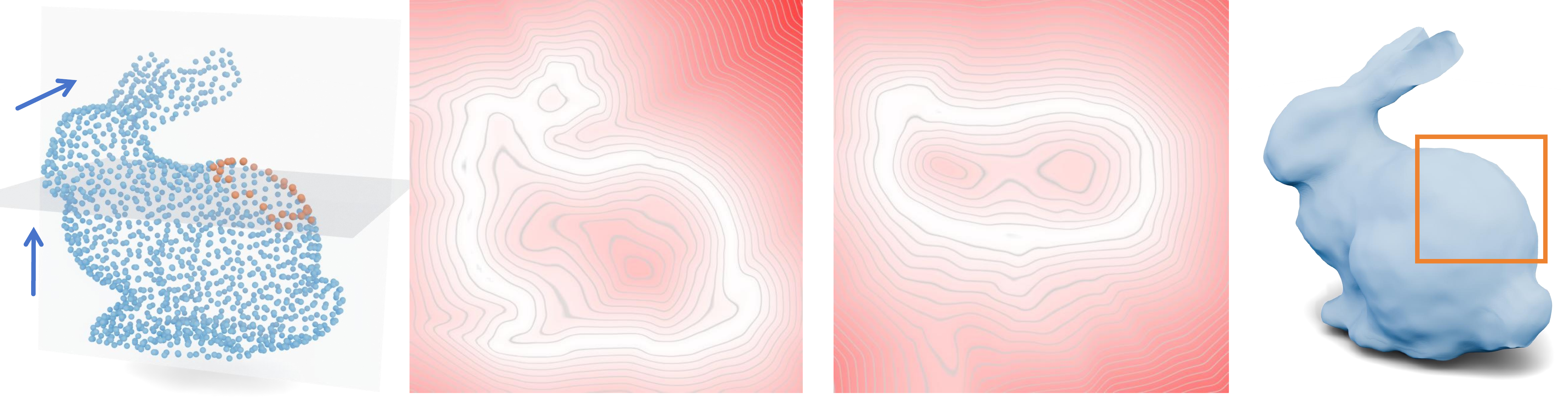}      \includegraphics[width=0.48\textwidth]{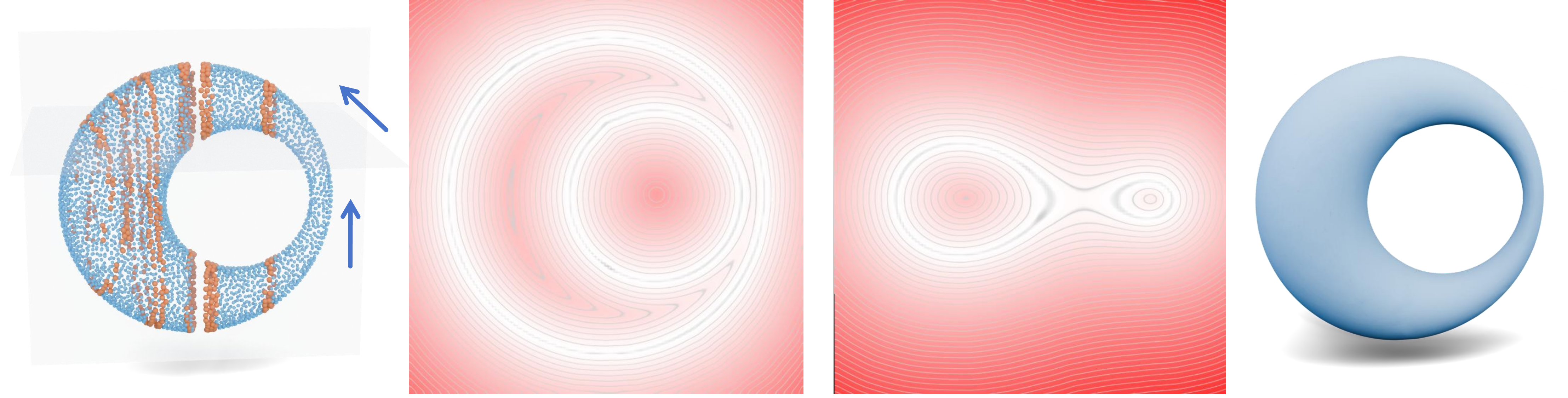}
     \begin{subfigure}[b]{0.115\textwidth}
        \centering
        \caption*{(a)}
    \end{subfigure}
    \begin{subfigure}[b]{0.118\textwidth}
        \centering
        \caption*{(b)}
    \end{subfigure}
    \begin{subfigure}[b]{0.118\textwidth}
        \centering
        \caption*{(c)}
    \end{subfigure}
    \begin{subfigure}[b]{0.113\textwidth}
        \centering
        \caption*{(d)}
    \end{subfigure}
    \caption{Surface reconstruction from inputs with missing regions. Our method successfully completes the UDF in (b) and (c) and reconstructs plausible surfaces in (d), demonstrating strong robustness to incomplete data.}
    \label{fig:incomplete}
\end{figure}

\paragraph{Incomplete Models}
We evaluate VAD on surfaces with incomplete geometry, including holes, missing regions, and disconnected fragments, as shown in Figure~\ref{fig:incomplete}. 
Benefiting from the smoothness of the underlying UDF, VAD effectively propagates structural information across small gaps and sparse regions, enabling plausible  reconstructions.
\begin{figure}[htb]
    \centering
    \includegraphics[width=\linewidth]{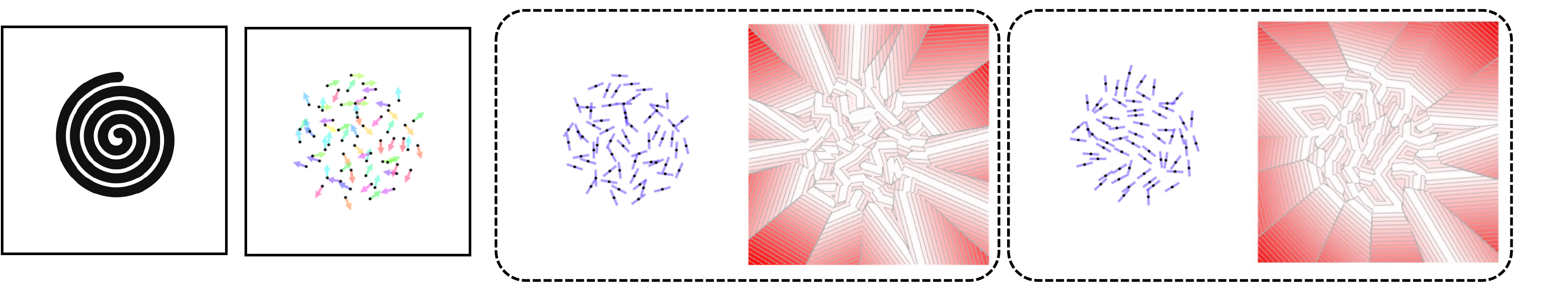}
    {\small
     \makebox[0.1\linewidth]{Original shape}
    \makebox[0.19\linewidth]{Input}
    \makebox[0.3\linewidth]{Iteration 0}
    \makebox[0.3\linewidth]{Final}}
    \caption{\rednote{Failure case under extreme sparsity on complex topology. For the spiral, when the sampling interval significantly exceeds the Local Feature Size (LFS), shape inference becomes inherently ambiguous. Lacking sufficient local geometric clues, our framework cannot reliably resolve the topology, resulting in chaotic normal orientations and an erroneous distance field.}}
    \label{fig:LFS_failure}
\end{figure}

\subsection{Limitations}
Voronoi diagrams play a central role in our method. For clean inputs, the Voronoi diagram is accurate, and the minimization step effectively resolves discrepancies of the projection distance field along bisectors. For inputs with moderate noise, our denoising strategy dynamically updates point positions during optimization, improving the quality of the Voronoi diagram. However, when the input point cloud is heavily contaminated with large amount of outliers, the resulting Voronoi diagram becomes unstable, and alternating optimization of normals and positions is insufficient to handle outliers. This often leads to incorrect orientation estimation and degraded reconstruction quality. To mitigate this issue, we recommend a preprocessing step to filter out outliers and reduce noise before Voronoi construction. As illustrated in Figure~\ref{fig:outlier}, such preprocessing improves the stability of iterative updates and yields more reliable surface reconstructions.

\rednote{Beyond noise and outliers, our method also faces inherent geometric limitations under extreme sparsity. As a geometry-based optimization approach, our framework lacks global data-driven shape priors. When the input exhibits severe ambiguity---such as extremely sparse sampling on topologically complex models where the sampling interval significantly exceeds the Local Feature Size (LFS)---inferring the underlying shape becomes fundamentally ill-posed. In such extreme scenarios, the available geometric clues are insufficient to reliably resolve the topology, resulting in chaotic normal orientations and an erroneous distance field. We illustrate this limitation in Figure~\ref{fig:LFS_failure}, where the sparse sampling of a complex structure leads to orientation failure.}

\section{Conclusion and Future Directions}
\label{sec:conclusion}

In this work, we introduced VAD, a method for computing unsigned distance fields from unoriented point clouds. By formulating a projection distance field based on Voronoi partitioning, our approach treats bi-directional normals as optimization variables, minimizing discrepancies along Voronoi bisectors. Once the normals are aligned, the heat method is applied to construct the unsigned distance field. Unlike existing methods that rely on neural networks or assume watertightness, VAD requires no pre-oriented normals and robustly and efficiently handles open surfaces, non-manifold structures, and non-orientable geometries. Extensive experiments demonstrate that VAD produces high-quality UDFs across a wide range of challenging inputs, offering a practical and effective alternative to existing reconstruction techniques.

\revise{In our current implementation, both the diffusion of bi-directional normals and the integration from the diffused vector field are performed based on~\cite{feng2024heat}. 
Because this implementation relies on voxelization, \rednote{the current grid resolution is limited to $512^3$.} 
This restriction can be alleviated by adopting an octree-based Poisson solver, as in the popular Poisson surface reconstruction algorithm~\cite{Kazhdan2006PoissonSR,kazhdan2013screened}, which would significantly enhance the scalability of VAD for larger point clouds. 
In addition, employing parallel solvers, such as those proposed in~\cite{tao2019parallel}, could further improve the runtime performance of the UDF computation step.}

\bibliographystyle{ACM-Reference-Format}
\bibliography{bib}

\appendix
\section{SDF Extension}
Although our method is mainly designed for computing UDFs from point clouds of open surfaces with possible non-manifold structures, it can also be adapted to compute SDFs for watertight manifold surfaces. 

Inspired by Power Crust~\cite{amenta2001power}, which selects Voronoi vertices far from their generating sites and classifies them as inside or outside poles to distinguish interior from exterior regions, we similarly leverage Voronoi vertices with known inside/outside attributes to determine the correct orientation of point cloud normals.

The algorithm begins by scaling the bounding box of the point cloud by a factor of five and adding its eight corner points as auxiliary sites in the Voronoi diagram. One of these corner points is easily chosen and labeled as an outside point, and its neighboring Voronoi vertices are subsequently marked as outside. 
\begin{figure}
    \centering
    \includegraphics[width=0.5\textwidth]{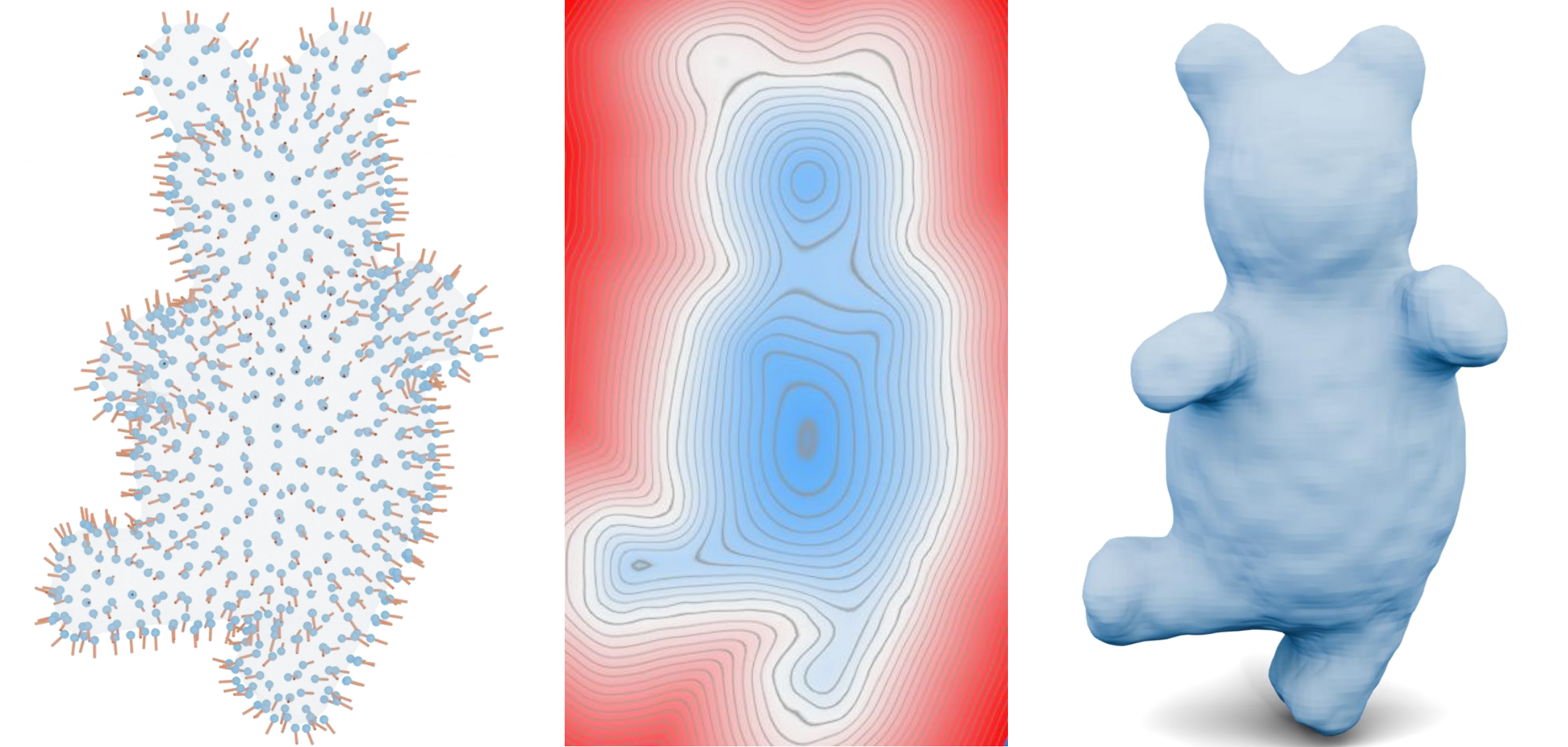}
\caption{Extending VAD for SDF computation on a toy bear model with 500 sample points. The method recovers signed distances, enabling surface extraction via Marching Cubes.}
    \label{fig:sdf}
\end{figure}

These labeled vertices then guide the outward orientation of adjacent surface points. Once a surface point is oriented, it can in turn help classify nearby unvisited Voronoi vertices as inside or outside. This alternating procedure continues until all surface points have been oriented.

After all normals are consistently oriented, the SDF can be computed directly using the heat method~\cite{feng2024heat}. Figure~\ref{fig:sdf} illustrates this process: starting from sparse surface points of a toy bear model, we first estimate bi-directional normals, orient them outward, and then compute the SDF. The reconstructed surface is extracted as the zero level set using the standard Marching Cubes algorithm~\cite{Lorensen1987}.

\begin{figure}
    \centering
\begin{minipage}[c]{0.03\linewidth} 
\centering
\begin{tabular}{c}
\rotatebox{90}{\small Ours}\\ [0.9cm] 
\rotatebox{90}{\small PCA}\\  [0.9cm] 
\rotatebox{90}{\small Ours}\\ [1.0cm] 
\rotatebox{90}{\small PCA}
\end{tabular}
\end{minipage}
\hfill 
\begin{minipage}[c]{0.95\linewidth} 
    \includegraphics[width=\linewidth]{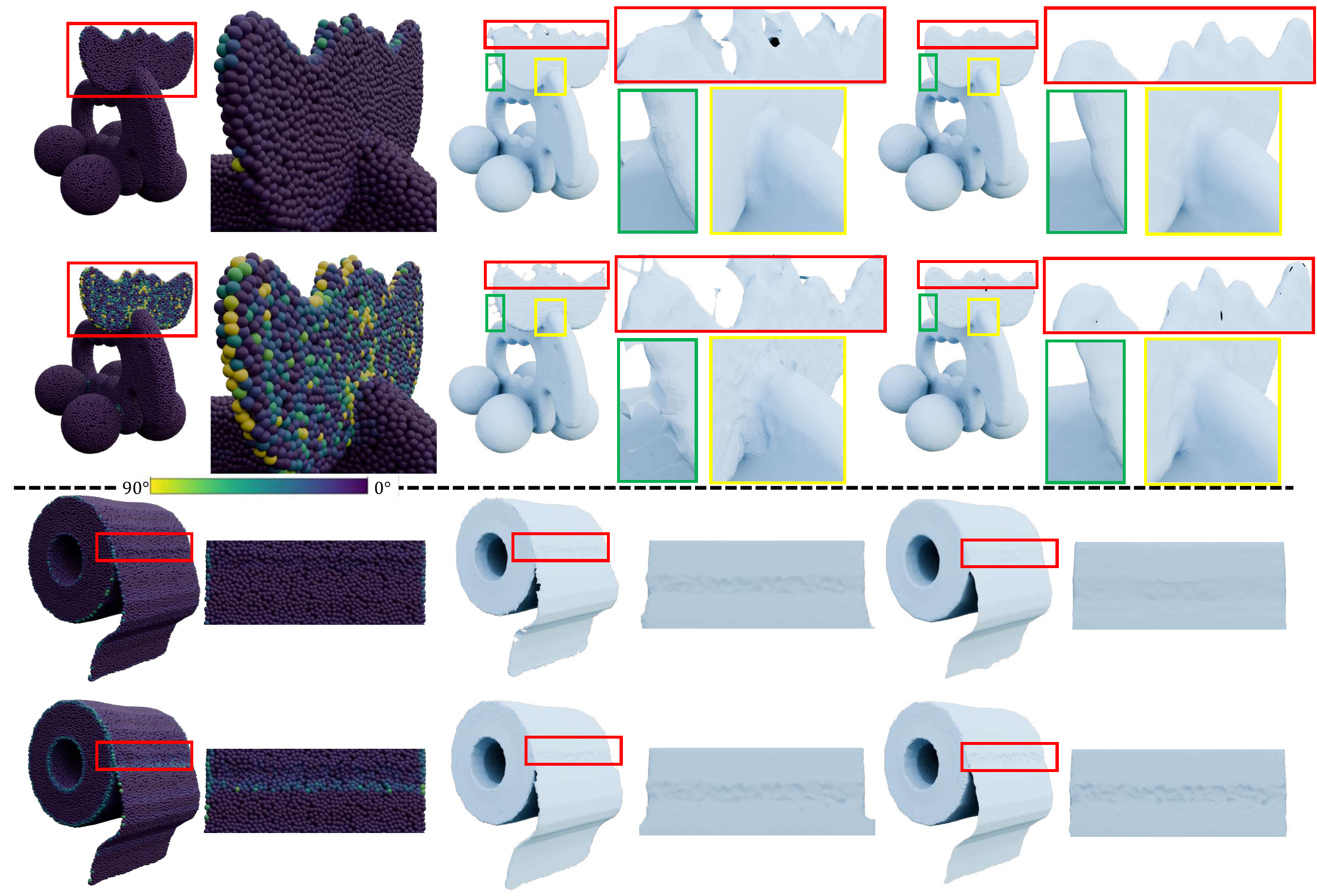}
\end{minipage}
    \makebox[0.34\linewidth][c]{\small (a) Normal Error}%
    \makebox[0.31\linewidth][c]{\small (b) w/o Diff}%
    \makebox[0.31\linewidth][c]{\small (c) w/ Diff}%
  \caption{\rednote{Ablation study on normal estimation and its impact on surface reconstruction. 
  (a) Normal Error: PCA exhibits unstable normal orientations, particularly around thin plates and non-manifold structures. (b) Direct Projection (w/o Diff): Extracting the zero-level set from the same projection field, inaccurate PCA normals cause erroneous surface fluctuations, while our optimized normals produce smoother results in flat regions but remain limited in high-curvature areas. (c) Heat Diffusion (w/ Diff):  Compared to direct projection, the diffusion-based formulation stabilizes both reconstructions. Nevertheless, the geometric fluctuations caused by inaccurate PCA normals remain visible, whereas our optimized normals yield a more faithful surface.}}\vspace{-0.2in}
    \label{fig:normal_diffusion_ablation}
\end{figure}

\begin{figure}
    \centering
    \includegraphics[width=\linewidth]{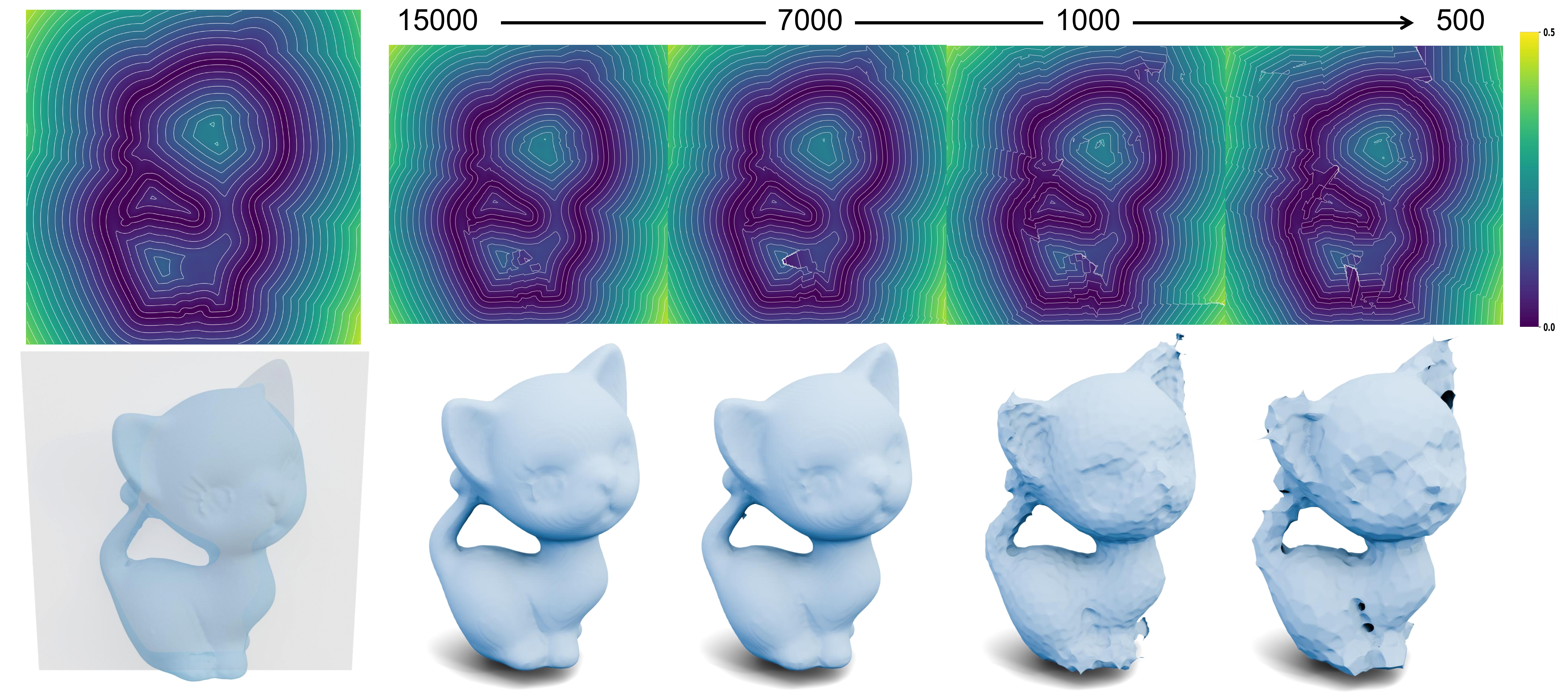}
     \makebox[0.24\linewidth][c]{\small  GT }%
    \makebox[0.7\linewidth][c]{\small Projection}%
    \caption{\rednote{Direct projection-based UDF fields and reconstruction results at different point densities. Without diffusion, the projection field degrades significantly under sparse sampling, showing noticeable discontinuities and artifacts near Voronoi boundaries.}}
    \label{fig:projection-fields}\vspace{-0.2in}
\end{figure}

\begin{figure}
    \centering
    \includegraphics[width=\linewidth]{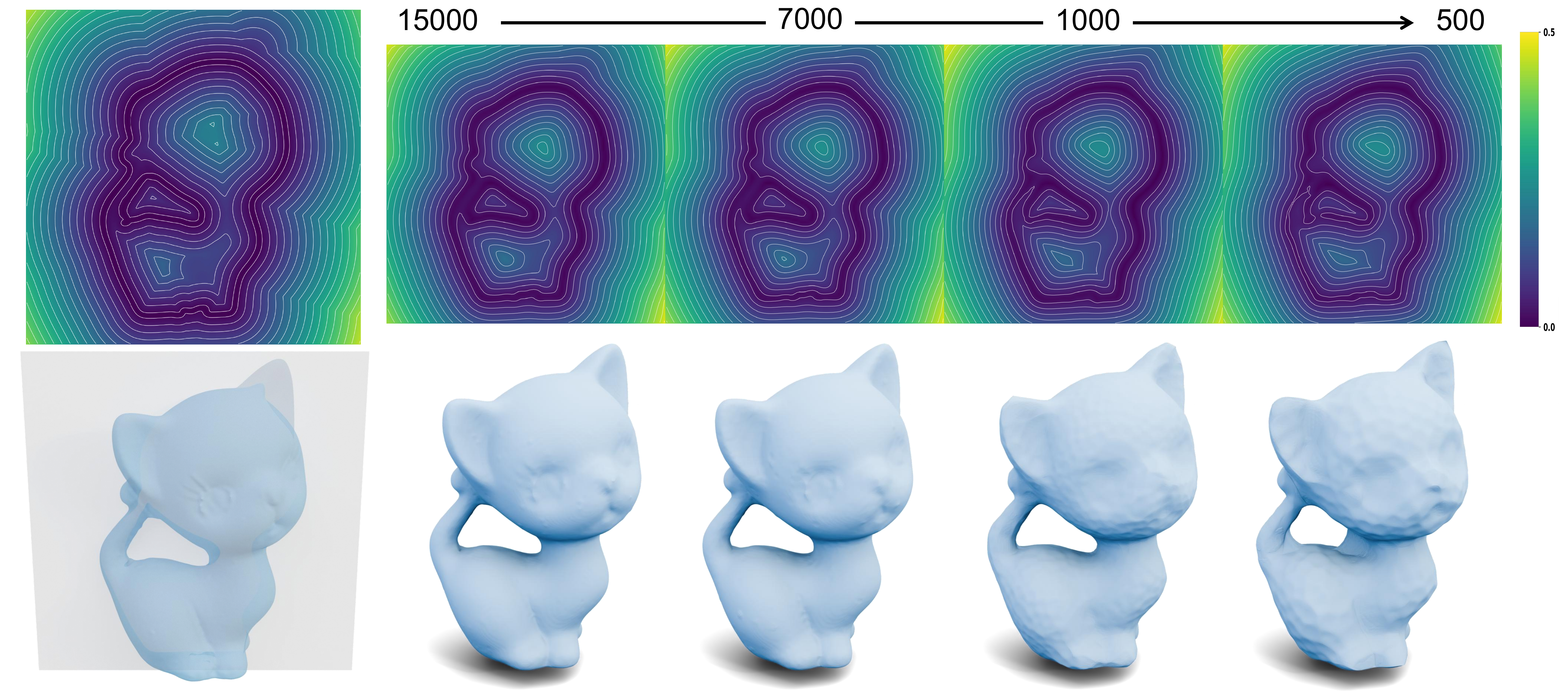}
     \makebox[0.24\linewidth][c]{\small  GT }%
    \makebox[0.7\linewidth][c]{\small Heat diffusion }%
    \caption{\rednote{Heat-diffusion-based UDF fields and reconstruction results at different point densities. The diffusion step effectively smooths the field and produces consistent gradients even under sparse sampling.}}
    \label{fig:heat_fields}
\end{figure}

\rednote{
\section{Further Ablation Studies}
To comprehensively evaluate the individual contributions of our proposed pipeline, we present additional ablation studies isolating the effects of our bi-directional normal optimization, the global heat diffusion module, as well as the impact of key weighting parameters.

\subsection{Effectiveness of Bi-directional Normal Optimization}
We first isolate the impact of our normal estimation strategy by comparing it against traditional PCA-based normals. As illustrated in Figure~\ref{fig:normal_diffusion_ablation}, directly computing a projection-based UDF from unoptimized PCA normals often results in topological tearing and geometric fragmentation. This degradation stems from the local nature of PCA, whose Euclidean KNN neighborhoods fail to provide reliable planar approximations in geometrically challenging regions, such as thin structures and non-manifold junctions. In contrast, our Voronoi-assisted neighborhood construction establishes a geometrically meaningful local support, producing a mathematically consistent bi-directional normal field that remains robust in these ambiguous configurations.


\subsection{Robustness Gained from Heat Diffusion}
We then evaluate the effect of the heat diffusion module by comparing the unsigned distance field (UDF) computed directly from optimized normals with the final UDF obtained after diffusion and Poisson integration.  Figures~\ref{fig:projection-fields} and~\ref{fig:heat_fields} provide slice-based visualizations of the resulting scalar fields under varying input densities.

The impact of diffusion becomes increasingly significant under sparse sampling. As shown in Figures~\ref{fig:projection-fields} and~\ref{fig:heat_fields}, reconstructions without diffusion degrade rapidly as the point density decreases, leading to fragmented surfaces, missing structures, and unstable topology. In contrast, the diffusion-based formulation produces stable reconstructions that preserve global shape consistency even under severe sparsity. These results suggest that while our normal optimization provides an accurate localized field, the global diffusion stage significantly improves robustness against sparse and incomplete observations, enabling highly reliable UDF reconstruction.}

\subsection{Ablation Studies on Weighting Parameters}
\begin{table}[t]
\setlength\tabcolsep{2pt}
\centering
\resizebox{0.48\textwidth}{!}{
\begin{tabular}{ccccccc}
\hline
 $(10^{2},1,10^{-3})$ &
$(10^{2},1,10^{-2})$ &
$(10^{2},1,10^{-4})$ &
$(10^{2},10,10^{-3})$ &
$(10^{2},10^{-1},10^{-3})$ &
$(10^{3},1,10^{-3})$ &
$(10^{1},1,10^{-3})$ \\ \hline
 0.9617 & 0.9438 & 0.9529 & 0.9103 & 0.9497 & 0.9478 & 0.9201 \\
 0.9812 & 0.9747 & 0.9811 & 0.9801 & 0.9762 & 0.9735 & 0.9844 \\ \hline
\end{tabular}
}
\caption{\revise{Ablation study on the weighting parameters $(\lambda_a,\lambda_d,\lambda_g)$. We report the average NC under two point densities, $n=3{,}000$ (top) and $n=10{,}000$ (bottom). The results show that VAD is less sensitive to parameter variations at higher point densities.}}
\label{tab:ablation}
\end{table}
We conduct ablation studies to examine the influence of the weighting parameters $\lambda_a$, $\lambda_d$, and $\lambda_g$ on bi-directional normal optimization. Specifically, we evaluate seven representative parameter combinations under two different point densities, and report the average NC in Table~\ref{tab:ablation}. The results show that when more points are available, the method is relatively insensitive to parameter variations, whereas with fewer points, the performance becomes more affected. Overall, the configuration $\lambda_a=10^2$, $\lambda_d=1$, and $\lambda_g=10^{-3}$ provides stable and reasonable performance across different conditions, which justifies our choice of fixing these values in the experiments.

\begin{figure}
    \centering
    \includegraphics[width=\linewidth]{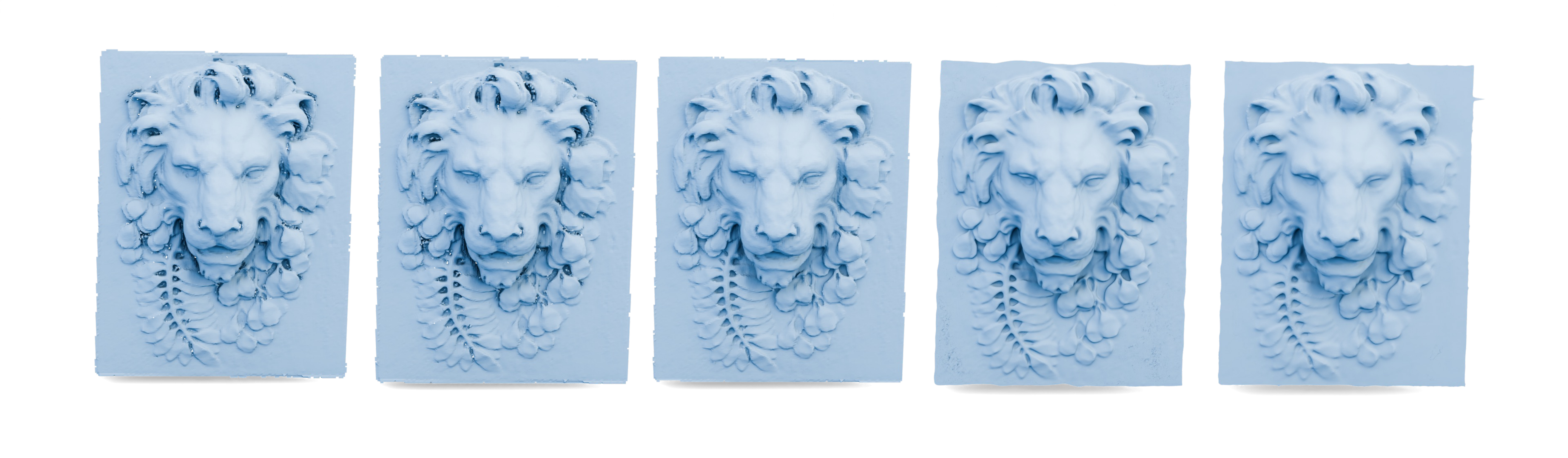}
      \makebox[0.17\linewidth]{CAPUDF}
    \makebox[0.17\linewidth]{GeoUDF}
    \makebox[0.17\linewidth]{MeshUDF}
    \makebox[0.17\linewidth]{DCUDF}
    \makebox[0.17\linewidth]{DCUDF2}
    \caption{\rednote{Flexibility of mesh extraction. We apply different mesh extraction methods to the UDF generated by our approach. The resulting reconstructions demonstrate that our method is decoupled from the downstream meshing stage and can be seamlessly integrated with different extraction algorithms. Depending on the target geometry and application requirements, users may choose different mesh extraction methods to achieve the desired balance between geometric fidelity, topological behavior, and reconstruction quality.}}
    \label{fig:flexible1}
\end{figure}

\rednote{
\section{Additional Results on Mesh Extraction Flexibility}
To further demonstrate the flexibility and extractor-agnostic nature of our method, we apply multiple representative mesh extraction algorithms directly to the continuous UDF field generated by VAD (see Figure~\ref{fig:flexible1}. These include single-layered approaches like CAP-UDF~\cite{Zhou2022CAP-UDF}, GeoUDF~\cite{DBLP:conf/iccv/RenHCHW23}, and MeshUDF~\cite{guillard2022udf}, as well as double-covered such as DCUDF~\cite{hou2023dcudf} and DCUDF2~\cite{chen2025dcudf2}.

Despite their distinct underlying design principles and varying strategies for resolving topological ambiguities (e.g., handling non-manifold junctions or localized connectivity traits), each method successfully extracts a coherent surface mesh from the exact same underlying UDF representation. 

This experiment highlights that VAD does not rigidly rely on a specific meshing mechanism. Instead, it recovers a robust, high-fidelity UDF field that seamlessly serves as a plug-and-play input for diverse downstream pipelines. Consequently, practitioners retain full flexibility to select the most suitable extraction method tailored to their specific application requirements---whether prioritizing lightweight single-sheet representations or maximizing feature-preserving topological wrapping.}
\begin{figure*}
    \centering
    \includegraphics[width=0.95\linewidth]{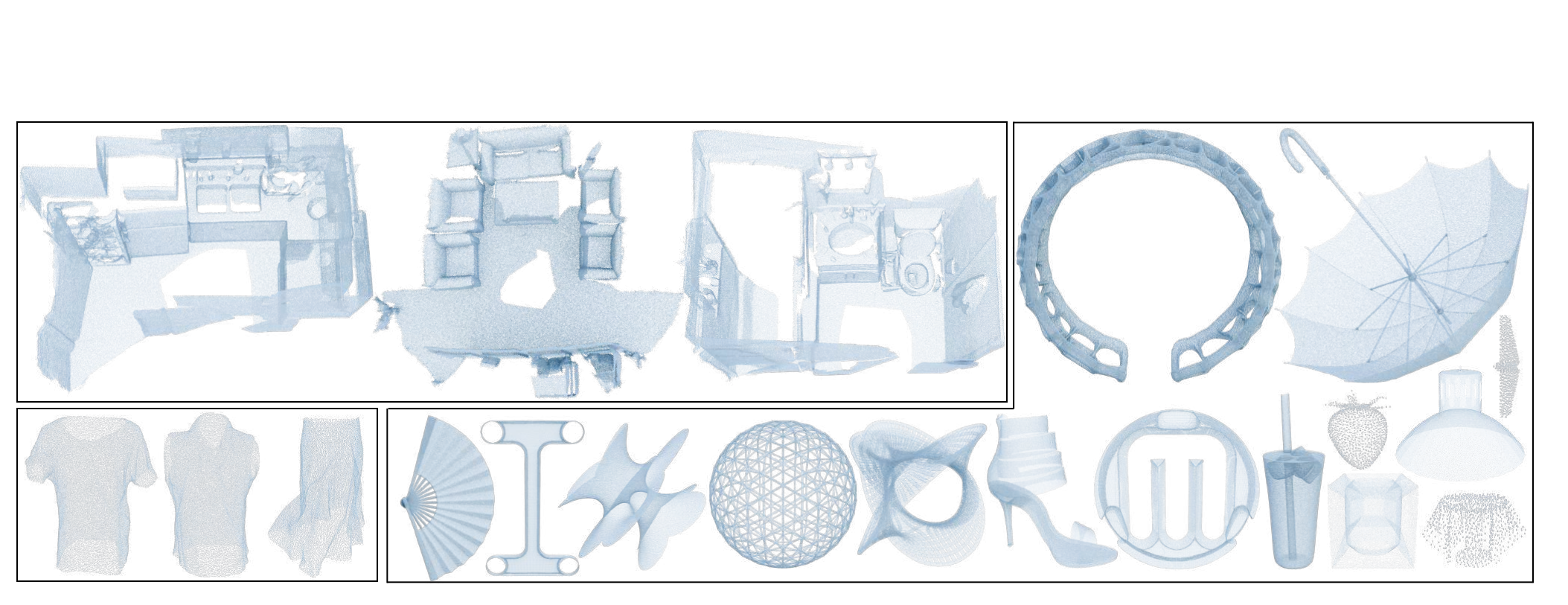}
    \caption{\rednote{Representative examples from the proposed benchmark. Three models are rendered for each category, including indoor scenes, clothes, medial-axis structures, non-manifold geometries, daily objects, open-boundary shapes, high-genus models, and sparse inputs. The benchmark spans a broad range of geometric characteristics, topological complexity encountered in real-world reconstruction tasks.}}
    \label{fig:benchmark}
\end{figure*}

\rednote{\section{Extended Benchmark and Evaluation Protocol}

\begin{table*}[htb]
\centering
\resizebox{0.95\textwidth}{!}{
\begin{tabular}{lcccc}
\toprule
Category & \#Models & Input Points & GT Type & Characteristics \\
\midrule

Indoor Scenes 
& 20
& $\sim$150K 
& Mesh 
& Noisy RGB-D reconstructions with open boundaries \\

Clothes
& 20
& $\sim$10K 
& Mesh 
& Thin deformable surfaces \\

Medial Axis 
& 20 
& $\sim$100K 
& Point 
& Skeletal / thin / Non-manifold structures \\

Non-Manifold Daily
& 23 
& $\sim$100K 
& Mesh 
& Non-manifold geometry \\

Open-Boundary Daily
& 22 
& $\sim$100K 
& Mesh 
& Partial / open surfaces \\

High-Genus 
& 5
& $\sim$200K 
& Mesh 
& Complex topology \\

Sparse 
& 5
& $\sim$100-1000
& Point 
& Sparse input supervision \\

\bottomrule
\end{tabular}
}\caption{\rednote{Statistics of the benchmark dataset. We report the number of models, average input point count, ground-truth representation, and geometric characteristics. Medial-axis structures are evaluated using point-based supervision, while all other categories use mesh-based ground truth.}}
\label{tab:dataset_stats}
\end{table*}
\begin{table*}[htb]
\centering

\resizebox{\textwidth}{!}{
\begin{tabular}{l|cc|cc|cc|cc|cc|cc|cc}
\toprule

& \multicolumn{2}{c|}{Scenes}
& \multicolumn{2}{c|}{Clothes}
& \multicolumn{2}{c|}{MA}
& \multicolumn{2}{c|}{Non-Manifold}
& \multicolumn{2}{c|}{Open}
& \multicolumn{2}{c|}{High-Genus}
& \multicolumn{2}{c}{Sparse} \\

\cline{2-15}

Method
& CD $\times 10^3$ & HD
& CD $\times 10^3$ & HD
& CD $\times 10^3$  & HD
& CD $\times 10^3$ & HD
& CD $\times 10^3$ & HD
& CD $\times 10^3$ & HD
& CD $\times 10^3$ & HD
\\

\midrule

Hoppe et al.
& 0.7917 & 0.03131
& 0.3537 & 0.01221
& -- & --
& -- & --
& 0.7087 & 0.01558
&4.4691 & 0.11025 
& 11.8754 & 0.13615
\\

DACPO
& 0.1185 & 0.00892
& 0.3765& 0.02065
& -- & --
& -- & --
& 0.1154 & 0.00703
& 0.3465 & 0.00548
& 13.4434 & 0.08898

\\

CAP-UDF
&  6.8245 & 0.12231  
&  0.6903 & 0.02553  
&  1.9865 & 0.05901  
&  6.5640 & 0.09577  
& 11.0669 & 0.13806  
&  1.5725 & 0.02262  
& 11.6721 & 0.10109  
\\

CAP-UDF (DCUDF)
&  6.6218 & 0.11975  
&  0.6682 & 0.03349 
&  1.9701 & 0.06159 
&  6.2984 & 0.09673  
& 10.2424 & 0.13899 
&  1.7527 & 0.02259 
& 15.1319 & 0.16720 
\\

GeoUDF
&  0.3912 & 0.00468      
&  0.3813 & 0.01577     
&  5.6150 & 0.02381     
&  1.8520 & 0.00962     
&  0.2483 & 0.00391      
&  0.2804 & 0.00406     
&  6.9560 & 0.05304     
\\

DEUDF
& 0.4658 & 0.041626
& 0.1149 & 0.01221
& 1.8884 & 0.04881
& 4.4817 & 0.11849
& 1.7935 & 0.06026
& 8.1864 & 0.14423
& failed & failed
\\

DUDF (CAP-UDF)
&  0.5806 & 0.02778     
&  0.3470 & 0.01488      
&  1.1510 & 0.01863     
&  1.4068 & 0.01908     
&  0.4363 & 0.01489      
&  2.2371 & 0.04122     
&  5.6620 & 0.03456     
\\
DUDF (MeshUDF)
&  0.6218 & 0.02855     
&  0.3669 & 0.01475     
&  1.1877 & 0.01839     
&  2.2667 & 0.03666        
&  0.4144 & 0.01404     
&  2.4925 & 0.03438      
&  7.2036 & 0.08780     
\\

\midrule

Ours
& 0.2158 & 0.01201
& 0.3668 & 0.01248
& 1.0071 & 0.01698
& 0.2085& 0.00815
& 0.1421 & 0.00839
& 0.4091 & 0.00794
& 1.0782 & 0.02127
\\

\bottomrule
\end{tabular}
}\caption{\rednote{
Quantitative comparison using Chamfer Distance (CD $\times 10^3$ $\downarrow$) and Hausdorff Distance (HD $\downarrow$) across different geometric categories.
The evaluated categories include indoor scenes (Scenes), clothes, medial-axis structures (MA), non-manifold daily objects (Non-Manifold), open-boundary daily objects (Open), high-genus models, and sparse-input settings. Medial-axis and Sparse results are evaluated using point-based ground truth, while all other categories are evaluated against mesh-based surfaces. For fair comparison, all methods are evaluated using meshes extracted at a marching cubes resolution of $512^3$. Our diffusion and UDF computation are performed on the same grid resolution. \citep{hoppe1992surface} and~\cite{li2025divideandconquerapproachglobalorientation} are omitted on Medial-axis and Non-Manifold categories due to their reliance on consistent normal orientations.
}}
\label{tab:benchmark_cd_hd}
\end{table*}

\begin{table*}[htb]
\centering
\resizebox{\textwidth}{!}{
\begin{tabular}{l|cc|cc|cc|cc|cc|cc|cc}
\toprule

& \multicolumn{2}{c|}{Scenes}
& \multicolumn{2}{c|}{Clothes}
& \multicolumn{2}{c|}{MA}
& \multicolumn{2}{c|}{Non-Manifold}
& \multicolumn{2}{c|}{Open}
& \multicolumn{2}{c|}{High-Genus}
& \multicolumn{2}{c}{Sparse} \\

\cline{2-15}

Method
& F@0.005 & F@0.010
& F@0.005 & F@0.010
& F@0.005 & F@0.010
& F@0.005 & F@0.010
& F@0.005 & F@0.010
& F@0.005 & F@0.010
& F@0.005 & F@0.010
\\

\midrule

Hoppe et al.
& 97.0313 & 99.1291
& 99.8867 & 99.9794
& -- & --
& -- & --
& 97.6858 & 99.3124
& 94.9003 & 95.6445
& 37.5484 & 69.9338
\\

DACPO
& 99.9584 & 99.9996
& 98.9237 & 99.6582
& -- & --
& -- & --
& 99.8756 & 99.9949
& 99.5525 & 99.9998
& 8.4111 & 29.4000
\\

CAP-UDF
& 88.3942 & 89.9443  
& 98.9831 & 99.1133  
& 94.9150 & 98.1091  
& 87.8754 & 89.8187 
& 84.3529 & 85.3077    
& 91.8638 & 96.7619  
& 12.2190 & 35.7651   
\\

CAP-UDF (DCUDF)
& 88.8213 & 90.4551  
& 98.3420 & 99.0647  
& 94.3230 & 98.0916  
& 88.3702 & 90.0542  
& 85.3809 & 86.2813  
& 89.7867 & 95.6173  
& 10.7046 & 32.3054  
\\

GeoUDF
& 99.9995 & 100.0000        
& 99.7839 & 99.9672     
& 69.7207 & 87.5602     
& 96.1523 & 96.6065     
& 99.9759 & 100.0000        
& 99.9982 & 100.0000        
& 21.7307 & 53.8358     
\\

DEUDF
& 98.9145 &  99.6065
&  99.9224 &  99.9751
&  95.0075 & 97.7712
& 92.2286 & 93.1475
& 97.9586 & 98.3772
&  88.6751 & 91.8140
& failed & failed
\\

DUDF (CAP-UDF)
& 98.6982 & 99.6814     
& 99.5225 & 99.9552     
& 96.8520 & 99.9414     
& 93.7906 & 96.3661     
& 98.8739 & 99.7845     
& 88.2003 & 95.8213     
& 31.1707 & 62.1202     
\\
DUDF (MeshUDF)
& 98.5542 & 99.6891     
& 99.4343 & 99.9559     
& 96.6314 & 99.9437     
& 93.7642 & 95.7648     
& 99.2027 & 99.8546     
& 87.2161 & 94.6011     
& 33.8671 & 66.6529     
\\

\midrule

Ours
& 99.7314 & 99.9961
& 99.4912& 99.7384
& 97.1116 & 99.9728
& 99.4327 & 99.9774
& 99.8611 & 99.9994
& 99.5687& 99.9874
& 87.7528 & 92.4855

\\

\bottomrule
\end{tabular}
}
\caption{\rednote{Quantitative comparison using F-score ($\uparrow$) under different thresholds across geometric categories. The evaluated categories include indoor scenes (Scenes), clothes, medial-axis structures (MA), non-manifold daily objects (Non-Manifold), open-boundary daily objects (Open), high-genus models, and sparse-input settings. Medial-axis and Sparse results are evaluated using point-based ground truth, while all other categories are evaluated against mesh-based surfaces. For fair comparison, all methods are evaluated using meshes extracted at a marching cubes resolution of $512^3$. Our diffusion and UDF computation are performed on the same grid resolution. \citep{hoppe1992surface} and~\cite{li2025divideandconquerapproachglobalorientation} are omitted on Medial-axis and Non-Manifold categories due to their reliance on consistent normal orientations.}}
\label{tab:benchmark_fscore}
\end{table*}

We construct a comprehensive benchmark consisting of more than 100 3D models spanning diverse geometric categories, including indoor scenes, garments, open surfaces, and complex non-manifold structures. The dataset is designed to cover a wide range of real-world point cloud scenarios with varying topology, sampling density, and geometric complexity. A detailed breakdown of the dataset composition is provided in Table~\ref{tab:dataset_stats}.

On this benchmark, we perform extensive quantitative evaluations using Chamfer Distance and Hausdorff Distance (Table~\ref{tab:benchmark_cd_hd}), as well as F-score (Table~\ref{tab:benchmark_fscore}), enabling a systematic assessment across different geometric settings and reconstruction difficulties.

To ensure fair comparisons across methods with different reconstruction pipelines, we evaluate several baselines under multiple surface extraction strategies when applicable. For instance, CAP-UDF and DUDF are additionally evaluated using DCUDF-based extraction in addition to their original pipelines, ensuring that performance differences reflect the quality of the underlying representation rather than differences in meshing heuristics. Furthermore, some models in the Medial and Non-Manifold categories contain non-orientable structures or orientation ambiguities. Since orientation-based methods such as PCA~\cite{hoppe1992surface} and DACPO~\cite{li2025divideandconquerapproachglobalorientation} require globally consistent normal orientations, they are not evaluated on these categories.

Overall, this benchmark provides a more comprehensive evaluation protocol for assessing the scalability and generalization ability of different reconstruction methods across diverse geometric scenarios.}
\end{document}